\pdfoutput=1

\documentclass[11pt]{article}

\usepackage[preprint]{acl}

\usepackage{array}
\usepackage{times}
\usepackage{latexsym}

\usepackage[T1]{fontenc}

\usepackage[utf8]{inputenc}

\usepackage{microtype}

\usepackage{inconsolata}
\usepackage{enumitem}
\usepackage[utf8]{inputenc} 
\usepackage{hyperref}       
\usepackage{url}            
\usepackage{tabularx, booktabs}       
\usepackage{amsfonts}       
\usepackage{nicefrac}       
\usepackage{microtype}      
\usepackage{xcolor}         
\usepackage{longtable}      
\usepackage{graphicx}       
\usepackage{lscape}         
\usepackage{wrapfig} 
\usepackage{array}
\usepackage{booktabs}
\usepackage{multirow}
\usepackage{gb4e}
\noautomath
\usepackage{caption}
\usepackage{subcaption}
\usepackage{tablefootnote}
\usepackage{CJKutf8}        
\usepackage{fontawesome}
\usepackage[inkscapearea=page]{svg}
\usepackage{adjustbox}
\usepackage{devanagari}     

\definecolor{lightred}{HTML}{e99090}
\newcommand{\name}{\textsc{\small CaLMQA}}
\newcommand{\ayanew}{\textsc{\small Aya-Expanse-32B}}
\newcommand{\claude}{\textsc{\small Claude-3-Opus}}

\newcommand{\gemini}{\textsc{\small Gemini-1.5-Pro}}

\newcommand{\gpt}{\textsc{\small GPT-4}}
\newcommand{\gptturbo}{\textsc{\small GPT-4-Turbo}}
\newcommand{\gpto}{\textsc{\small GPT-4o}}
\newcommand{\gptomini}{\textsc{\small GPT-4o-Mini}}
\newcommand{\llama}{\textsc{\small Llama-3-70B}}
\newcommand{\mixtral}{\textsc{\small Mixtral-8x22B}}

\title{\raisebox{-0.5ex}{\includegraphics[height=1.2em]{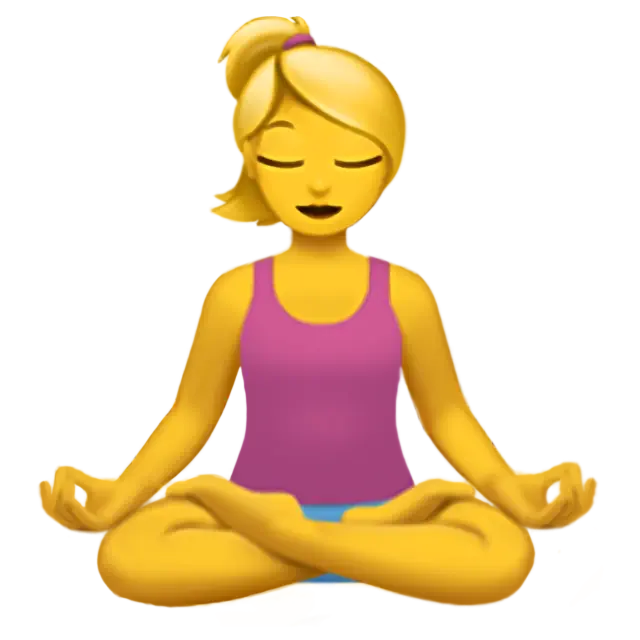}} CaLMQA: Exploring culturally specific long-form \\question answering across 23 languages}

\author{
    Shane Arora$^{\clubsuit*}$
    ~~ Marzena Karpinska$^{\heartsuit*}$
    ~~ Hung-Ting Chen$^{\diamondsuit}$ \\
    \textbf{Ipsita Bhattacharjee$^{\heartsuit}$
    ~~ Mohit Iyyer$^{\heartsuit \dag}$
    ~~ Eunsol Choi$^{\diamondsuit \dag}$} \\
    The University of Texas at Austin$^\clubsuit$\\
    New York University$^\diamondsuit$ \\
    University of Massachusetts Amherst$^\heartsuit$ \\    
    \hspace{0.5em} {\texttt{shane.arora@utexas.edu}}  {\texttt{\{hung-ting.chen, eunsol\}@nyu.edu}} \\
    \hspace{0.5em} {\texttt{\{mkarpinska, ibhattacharj, miyyer\}@umass.edu}} \\
}

\begin{document}
\maketitle

\def\thefootnote{*\textdagger}\footnotetext{These authors contributed equally to this work.}\def\thefootnote{\arabic{footnote}}

\begin{abstract}
Despite rising global usage of large language models (LLMs), their ability to generate \emph{long-form} answers to \textit{culturally specific} questions remains unexplored in many languages. To fill this gap, we perform the first study of textual multilingual long-form QA by creating \name{}, a dataset of \textbf{51.7K} culturally specific questions across \textbf{23} different languages. We define culturally specific questions as those that refer to concepts unique to one or a few cultures, or have different answers depending on the cultural or regional context. We obtain these questions by crawling naturally-occurring questions from community web forums in high-resource languages, and by hiring native speakers to write questions in under-resourced, rarely-studied languages such as Fijian and Kirundi. Our data collection methodologies are translation-free, enabling the collection of culturally unique questions like `Kuber iki umwami wa mbere w'uburundi yitwa Ntare?'' (Kirundi; English translation: ``Why was the first king of Burundi called Ntare (Lion)?''). We evaluate factuality, relevance and surface-level quality of LLM-generated long-form answers, finding that (1) for many languages, even the best models make critical surface-level errors (e.g., answering in the wrong language, repetition), especially for low-resource languages; and (2) answers to culturally specific questions contain more factual errors than answers to culturally agnostic questions -- questions that have consistent meaning and answer across many cultures.
We release \name{} to facilitate future research in cultural and multilingual long-form QA.
\end{abstract}

\begin{table}[h]
\centering
    \renewcommand{\arraystretch}{1.1}
    \begin{tabular}{c@{\hskip.2cm}l}
         \faGithub & \href{https://github.com/2015aroras/CaLMQA}{\path{github.com/2015aroras/CaLMQA}} \\
         \includeinkscape[width=0.9em]{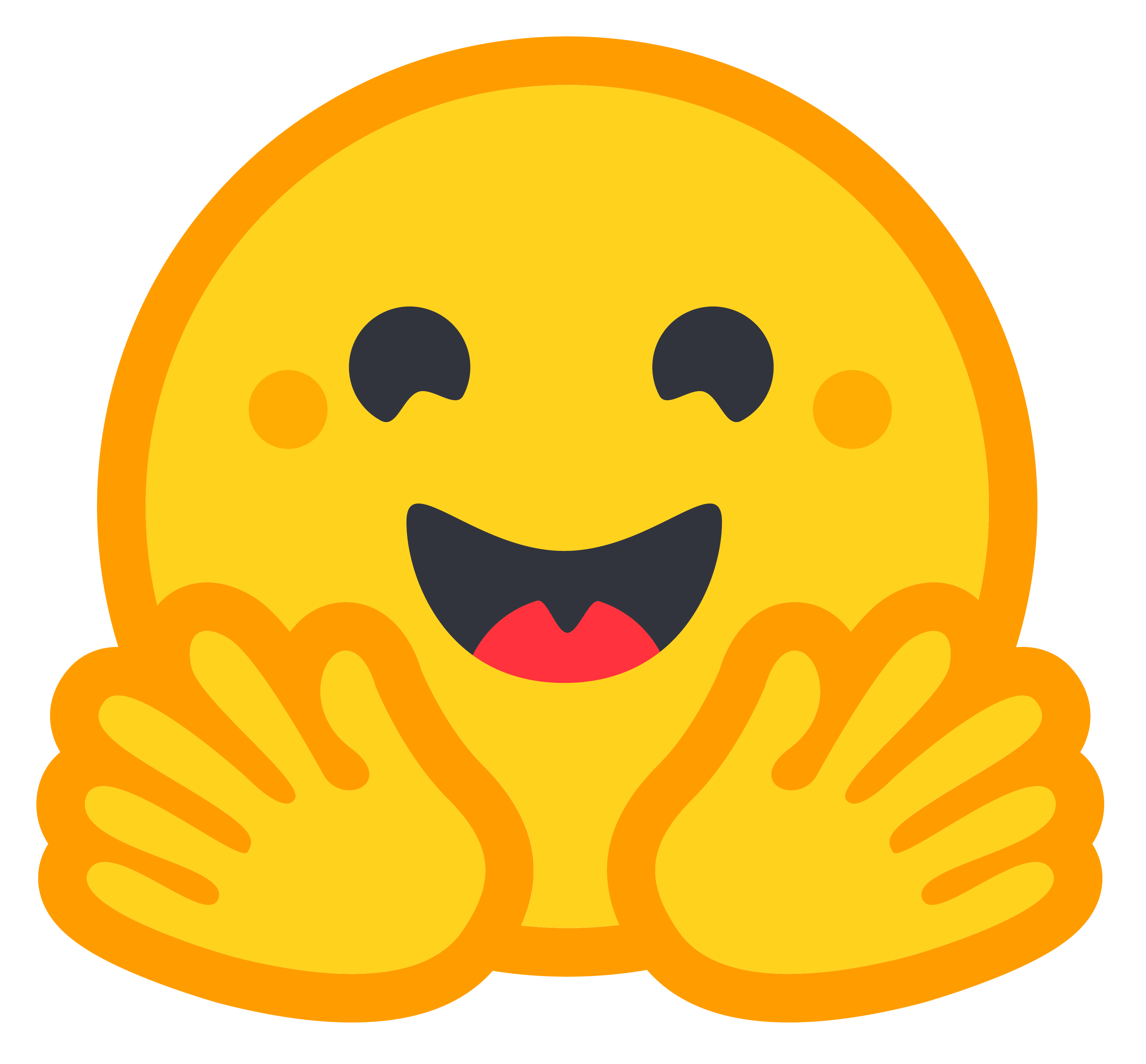} & \href{https://hf.co/datasets/shanearora/CaLMQA}{\path{hf.co/datasets/shanearora/CaLMQA}} \\
         \textcopyright{}  & \href{https://creativecommons.org/licenses/by/4.0/}{CC BY 4.0}
    \end{tabular}
\end{table}  
\section{Introduction}

\begin{figure*}[tbp]
  \includegraphics[width=0.95\linewidth]{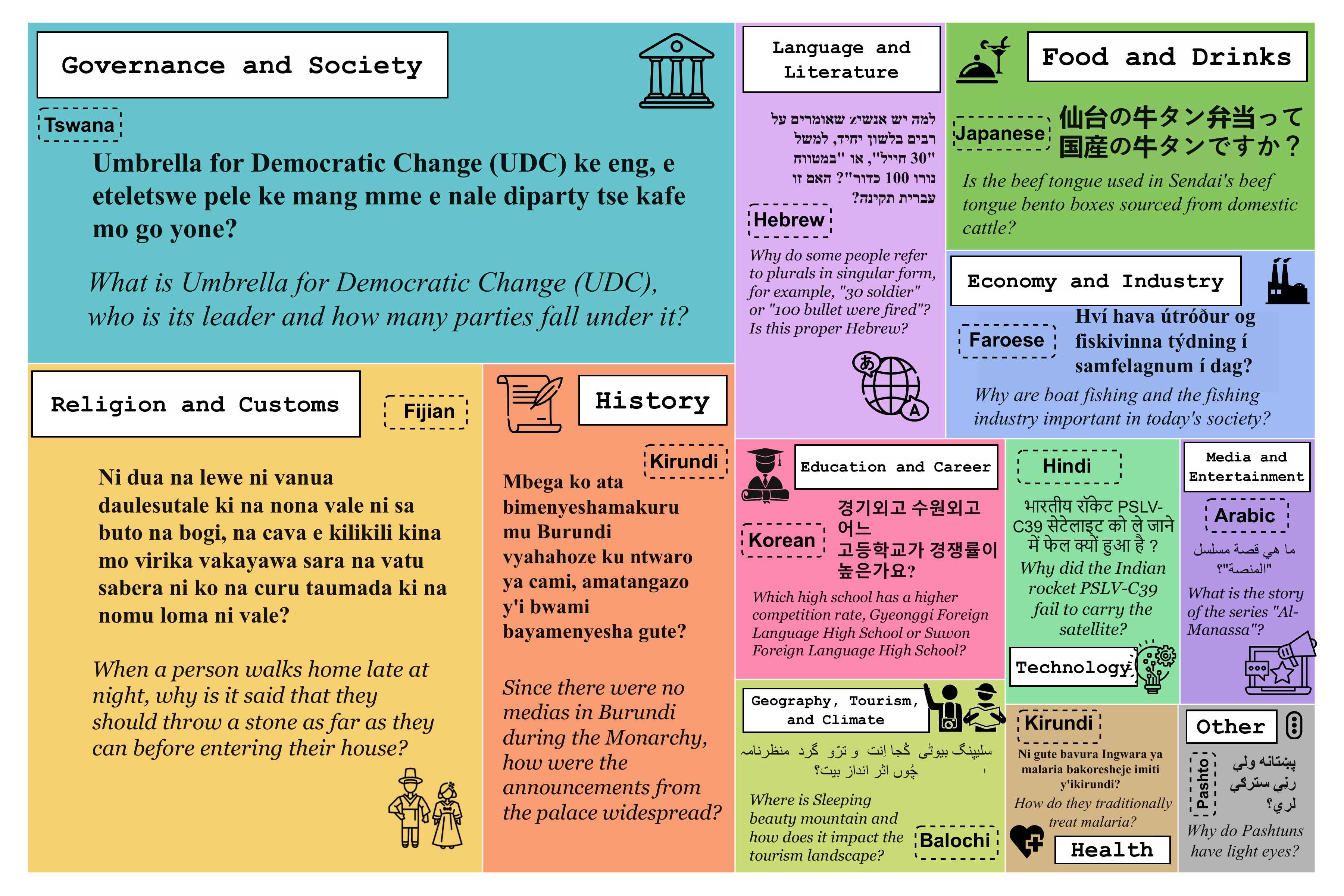} \vspace{-1em}
  \caption{Distribution of topics in \name{}, with box size indicating the frequency of each topic. Each topic is accompanied by an example and its English translation. \autoref{tab:categories_of_questions} contains descriptions of the topics, and \S\ref{sec:categorization} describes our topic classification method.}\vspace{-1em}
  \label{fig:categories}
\end{figure*}

While large language models (LLMs) are increasingly used by people across the world, most NLP efforts are focused on English and western cultures. Growing evidence reveals significant gaps in their performance across languages \citep{Qiu2023DetectingAM, Guerreiro2023HallucinationsIL} and their understanding of diverse cultures \citep{tao2024culturalbiasculturalalignment, li2024culturegen}, as well as a persistent bias toward Western-centric perspectives \citep{palta-rudinger-2023-fork,durmus2024measuringglobal, alkhamissi-etal-2024-investigating, naous-etal-2024-beer}. Existing research of multilingual QA largely focuses on assets derived from English resources~\citep{singh2024globalmmluunderstandingaddressing, Zhang2023xDialEvalAM, lai2023okapiinstructiontunedlargelanguage}, limiting their coverage of culturally unique concepts especially in low-resource languages. While some prior work collects short-answer and multiple-choice questions in non-English languages~\cite{myung2025blendbenchmarkllmseveryday, clark2020tydi, liu-etal-2019-xqa}, multilingual long-form QA, a task more aligned with real-world applications, remains unexplored.



In this work, we develop a translation-free multilingual QA dataset of long-form culturally specific questions: \textbf{C}ultur\textbf{a}l \textbf{L}ong-form \textbf{M}ultilingual \textbf{Q}uestion \textbf{A}nswering (\name). Questions are posed in the language of the target culture and demand nuanced, long-form responses. We only collect \textit{\textbf{culturally specific}} questions that (1) refer to concepts unique to one or a few cultures, such as ``Kuber iki umwami wa mbere w'uburundi yitwa Ntare?'' (Kirundi),\footnote{English translation: ``Why was the first king of Burundi called Ntare (Lion)?''} or (2) have different answers depending on the cultural or regional context, as in ``\ {\dn b\2\8{d}k kA lAis\?{\qva}s k\4s\? bntA h\4{\rs ?\re}}'' (Hindi).\footnote{English translation: ``How do you get a gun license?''}
We contrast the quality of an LLM's answers to these questions with its answers to \textit{\textbf{culturally agnostic}} questions that have consistent meaning and answer across many cultures (e.g., ``Why is smoking bad for the heart?''), which are prevalent in many translation-centric multilingual QA works.

Evaluation of multilingual long-form QA is challenging: lexical metrics for short-form QA do not correlate with human preferences in long-form QA \citep{krishna2021hurdles,xu2023critical} or transfer from English to other languages \citep{kang2024comparing, Koto2021EvaluatingTE, min2023factscore, song2024veriscore}. 
We thus adopt a multi-aspect evaluation protocol including (1) surface-level measures of language identification and repetition; (2) automatic factuality and relevance metrics run on translated answers; and (3) human evaluations from native speakers.
To distinguish the effects of culture and language on model performance, we use a baseline set of parallel culturally agnostic questions created by translating a seed set of 51 English questions into the 22 other languages, following common practice in prior work \citep{vayani2024languagesmatterevaluatinglmms, Artetxe_2020_xquad, lewis-etal-2020-mlqa, alonso2024medexpqa}.

We show that seven popular LLMs, including closed models such as \claude{}, \gemini{} and \gpto{}, suffer from basic surface-level issues, especially on low-resource languages (e.g., none of them reliably generate text in Afar). Also, open-weight models such as \mixtral{} and \llama{} often apologize instead of providing an answer or generate text in English when prompted with non-English questions. 
We observe that the factuality and relevance of LLM-generated culturally specific answers is significantly lower than that of culturally agnostic answers, underscoring the importance of studying culturally specific questions. Factuality and relevance drop considerably on low-resource languages, with \gptturbo{} and \gpto{} performing best.

We conduct a human evaluation on a subset of the data (spanning five languages) for the best-performing models. 
Native speakers rate and rank answers from different LLMs, and an analysis of their annotations reveals that omissions and factuality issues are strong predictors of answer quality ratings. This human evaluation also supports our automatic factuality and relevance evaluations in that culturally agnostic questions are twice as likely to receive higher ratings than culturally specific questions, regardless of the generation model.

Overall, our work establishes a foundation for studying multilingual long-form question answering by releasing \name{} -- the first textual multilingual long-form question answering (LFQA) dataset, with 51.7K questions across 23 languages derived from culturally specific sources.

\section{CaLMQA: Cultural Long-form Multilingual Question Answering}
Each of the 51.7K examples in \name{} consists of (1) a \textit{culturally specific} question written in one of 23 languages, (2) an optional human-written English translation (for low-resource languages), and (3) an optional human-written reference answer (for high- and mid-resource languages). We detail \name's collection process and statistics below.

\subsection{What questions are culturally specific?}

Culture is a multifaceted and abstract concept that eludes a simple definition \citep{adilazuarda2024measuringmodelingculturellms, Liu2024CulturallyAA}. We define culturally specific questions as questions that (1) refer to topics, concepts, objects, entities or events that are unique to one or a few cultures, or (2) have different answers depending on the cultural or regional context. Our notion of culturally specific questions is based on \citet{Liu2024CulturallyAA}: ``1) basic concepts that
are `configured' differently, reflecting the cultural-
specific way of thinking, and 2) concepts that are
unique to a culture''; our definition embeds the former by including questions with answers dependent on culture, and the latter by including questions that refer to concepts related to culture. \citet{Liu2024CulturallyAA} taxonomizes cultural NLP works into 10 categories including values, norms and morals, and knowledge; we collect that cultural knowledge in \name{}.

\subsection{Data Collection}   \label{sec:data_collection}
We collect our dataset through two processes. For high- and mid-resource languages, we follow prior work~\cite{fan2019eli5} and collect questions from community Q\&A forums. For low resource languages where such web content is scarce, we hire freelancers to write culturally specific questions.

\paragraph{Culturally specific questions for high- and mid-resource languages:}
Many countries have their own community forums where people can exchange information, similar to Quora, Reddit or StackExchange in English. We collect culturally specific questions from these websites via a crowdsourcing process that we scale with LLM assistance: first, we ask English-proficient Prolific\footnote{\url{https://www.prolific.com/}}  crowdworkers from different countries to provide a link to a community web forum in their language that contains many complex questions that cover a diverse range of topics. Next, we ask workers to collect culturally specific questions and real users' answers from the identified websites, for \$0.65-1.33 USD per question. We manually review all provided examples and websites.
Our workers yielded 923 questions across 11 languages with answers at a cost of \$1427 USD (\autoref{tab:dataset_statistics}, left). Refer to \S\ref{sec:high_resource_collection} for more details.

We scale our question collection process by automating the collection and verification of questions. We obtain around 10k questions for each language. For English, Chinese and Russian, we use existing Hugging Face datasets containing questions scraped from our chosen websites \citep{eli5-category, wangrui6ZhihuKOL, its5Qyandex}. For the remaining high-resource languages (except Hebrew, for which we were unsuccessful), we implement and utilize website-specific question extraction scripts. We do not collect answers due to the challenges of extracting them. We filter our questions using \gptomini{}, with two model passes that assess each question's cultural specificity and general quality, retaining 52\% of questions (prompts in \autoref{tab:cultural_filter_prompt} and \autoref{tab:general_filter_prompt}). We apply these filters on the worker-collected questions too, retaining ${>}90\%$ of questions. This procedure yielded 50,227 additional questions at a cost of \$34 USD.

\paragraph{Culturally specific questions for low-resource languages:}
Unlike existing LFQA datasets, \name{} also includes twelve \emph{low-resource languages}  (\autoref{tab:dataset_statistics}, right). We choose languages with scarce online resources that are not well-studied in prior work, but for which we can also find at least one annotator bilingual in English.
We hire 29 native speakers (1-3 annotators per language, depending on their availability) on Upwork,\footnote{\url{https://www.upwork.com/}} each of whom receives guidelines, takes a paid (\$7 USD) comprehension test, and then writes culturally specific questions with English translations for \$0.65-1.00 USD per question. As having them write answers for all of these languages is prohibitively expensive, we collect answers and their English translations only for Kirundi (\$2 USD per question, \$106 USD total). This process yielded a total of 548 questions with English translations at a cost of \$833 USD. The protocol was reviewed and deemed \textit{exempt} by an Institutional Review Board. Please refer to \S\ref{sec:low_resource_collection} for more details.

\paragraph{Quality control:} We screened crowdworkers through a qualification task to ensure understanding of culturally specific, long-form questions. Each question curated by workers was manually reviewed by at least one author; workers provided clarifications or replaced unsuitable questions as needed. In the case of low-resource languages, the questions were also checked by another annotator of that language. See \S\ref{sec:high_resource_collection} for detailed guidelines.


\begin{table*}
\scriptsize
\centering
\resizebox{1\textwidth}{!}{%
\begin{tabular}{@{}lrrrr@{}}
\toprule
\addlinespace
\textsc{Language}   & \textsc{\# Q}  & \textsc{\# A} & \textsc{Q. Bytes}  & \textsc{A. Bytes} \\ 
 & & & \textsc{(avg/std)}  & \textsc{(avg/std)} \\ 
\midrule
English & 2617 & 78 & 205.1 / 209.4 & 674.1 / 475.9
    \\
Arabic & 5300 & 85 & 127.0 / 77.2 & 2105.0 / 2378.6
    \\
Chinese & 5901 & 75 & 69.0 / 49.2 &  588.8 / 939.7
    \\
German & 4091 & 96 & 427.8 / 451.9 & 1169.0 / 744.7 
    \\
Hebrew & 96 & 96 & 142.5 / 84.2&  2043.6 / 1934.9 
    \\
Hindi & 6404 & 91 & 133.7 / 46.5 & 3618.8 / 1867.1
    \\
Hungarian & 3843 & 75 & 366.0 / 441.8  & 379.3 / 333.2
    \\
Japanese & 6466 & 75 & 814.2 / 696.7  & 920.6 / 637.1 
    \\
Korean & 5875 & 75 & 248.8 / 198.5 & 1008.6 / 936.3 
    \\
Russian & 5403 & 75 & 291.0 / 487.7 & 4546.7 / 5067.9 
    \\
Spanish & 5058 & 102 & 547.3 / 544.0 & 852.0 / 817.9 
    \\ \midrule
    Total & 51150 & 923 & 152.8 / 140.8 & 1640.8 / 2291.3 \\
\bottomrule
\\
\end{tabular}
\quad
\begin{tabular}{@{}lrrr@{}}
\toprule
\addlinespace
\textsc{Language}   & \textsc{\# Q} & \textsc{Q. Bytes} & \textsc{A. Bytes} \\  
 & & \textsc{(avg/std)} & \textsc{(avg/std)} \\
\midrule
Afar & 25 & 43.7 / 16.5 &  N/A
    \\
Balochi & 65 & 122.7 / 52.4 &  N/A
    \\
Faroese & 30 & 47.8 / 16.6 &  N/A
    \\
Fijian & 75 & 75.0 / 36.9 &  N/A
    \\
Hiligaynon & 65 & 93.4 / 39.1 &
   N/A \\
Kirundi & 53 & 64.6 / 21.2 & 557.2 / 160.9
    \\
Papiamento & 10 & 66.8 / 28.5 &  N/A
    \\
Pashto & 75 & 64.8 / 26.9 &  N/A
    \\
Samoan & 25 & 51.2 / 19.3 &  N/A
    \\
Tongan & 10 & 81.2 / 19.2 &  N/A
    \\
Tswana & 65 & 87.2 / 43.4 &  N/A
    \\
Wolof & 50 & 45.3 / 18.9 &  N/A\\ \midrule
    Total & 548 & 75.1 / 41.3 & 557.2 / 160.9
    \\
\bottomrule
\end{tabular}%
\vspace{-1em}
}
\caption{Data statistics of high- \& mid-resource language (left) and low-resource language (right) culturally specific questions. We report the number of bytes in the UTF-8 encoding as token counts will significantly vary between the languages. For high- \& mid-resource languages, answers were only obtained for a subset of questions collected by crowdworkers, due to challenges with extracting and ranking answers automatically. For low-resource languages, we collect answers for Kirundi only. See \autoref{tab:dataset_statistics} for culturally agnostic questions.}
\label{anal_tab:cultural_question_data}
\vspace{-0.4em}
\end{table*}

\subsection{Dataset Analysis}
\autoref{anal_tab:cultural_question_data} and \autoref{tab:dataset_statistics} summarize the statistics \name's 51.7K culturally specific questions. We measure the length of questions with bytes~\cite{clark2020tydi} as token count is not comparable across languages due to different compression rates~\cite{Ahia2023DoAL}. High- and mid-resource language questions are generally longer than low-resource language questions, except for Arabic and Balochi. This can be largely attributed to different collection method (gathered from community forums vs. manually written by crowdworkers); see \autoref{tab:examples} for examples.

Finally, we categorize \name's questions based on their topic by first manually curating a set of categories and developing \gptturbo-based pipeline. \autoref{fig:categories} shows a treemap of the question categories with examples. We find that the distribution of categories of culturally specific questions is similar between different languages. See \S\ref{sec:categorization} for details.

\section{Evaluating LLMs on \name{}}
We evaluate answers from seven state-of-the-art LLMs using automatic metrics for surface quality, relevance and factuality, combining these into a unified metric. We supplement this with human evaluation of LLM answers across five languages.

\paragraph{Models:}
We evaluate four closed-source LLMs (\claude{}, \gemini{}, \gptturbo{}, \gpto{} \citep{anthropic-claude, geminiteam2024gemini, openai2024gpt4, openai-gpt4o} and three open-weights LLMs (\ayanew{}, \llama{}, \mixtral{} \citep{dang2024ayaexpansecombiningresearch, llama3modelcard, mistral-mixtral8x22b}. Model details are in Appendix \autoref{tab:evaluated_models}.

\paragraph{Inference Setting: } Each model is prompted with a question from our dataset in a zero-shot setup without instructions. We use greedy decoding and limit outputs to 2048 tokens. The total cost of API calls is \$530 USD.\footnote{We note the total cost of calls for each model as follows: \gemini{} \$17 USD, \gpto{} \$40 USD, \gptturbo{} \$80, \llama{} and \mixtral{} \$4 USD, and \claude{} \$390 USD.}  

\paragraph{Data:}

For controlled comparison of LLM performance on questions with and without cultural knowledge requirements, we assemble an evaluation set of 3,644 questions from three sources: (1) all 1471 human-collected culturally specific questions, (2) 100 randomly sampled automatically collected questions per language, and (3) 51 culturally agnostic questions from \texttt{\href{https://www.reddit.com/r/explainlikeimfive/}{r/explainlikeimfive}} translated into 22 languages using \gptturbo{}, which has demonstrated superior translation performance~\citep{yan2024gpt4vshumantranslators, jiao2023chatgptgoodtranslatoryes}. For Balochi, Fijian, and Kirundi, where translation quality was poor, we hire native speakers. This subset allows comprehensive evaluation while managing computational costs compared to using our full dataset of 51.7K questions.

\subsection{Automatic Evaluation Metrics}    \label{sec:automatic_eval}

Since common QA metrics like BLEU \citep{papineni-etal-2002-bleu} and ROUGE \citep{lin-2004-rouge} do not correlate well with human judgement for long-form QA \citep{xu2023critical, krishna2021hurdles}, we (1) identify answers with surface level issues (e.g. incorrect language), (2) measure factuality and relevance of the remaining answers using the VeriScore pipeline of~\citet{song2024veriscore} and LLM-as-a-Judge~\citep{zheng2023judgingllmasajudgemtbenchchatbot} with \gpto{} respectively, and (3) combine our individual measures to produce a single metric of answer quality.

\paragraph{Identifying surface-level issues ($S_{surf} \in \{0,1\}$):} Useful answers must be in the correct language (i.e., the language of the question) and free from word or phrase repetition. We start by detecting answers in the wrong language using a pipeline that combines \texttt{polyglot}\footnote{\url{https://pypi.org/project/polyglot/}}  and \texttt{langid}\footnote{\url{https://pypi.org/project/py3langid/}}, which yields optimal results for most languages (see \autoref{app:acc_lang_detect_on_test_set} for accuracy). Balochi, Kirundi, Papiamento, and Hiligaynon are excluded due to low language identification accuracy. Then, we identify responses with repetitions by employing \texttt{tiktoken}\footnote{\url{https://github.com/openai/tiktoken}} with the \texttt{o200\_base} encoding and flagging any answers in which a sequence of 20 tokens is repeated four or more times.\footnote{\gemini{} often returned an API error for questions in low-resource languages; we mark such answer as invalid.} See \S\ref{app:auto_eval_method} for details. We assign a score of 1 if there is no surface issue, 0 otherwise. 

We only evaluate factuality and relevance for answers without surface-level issues. 

\paragraph{Evaluating factuality ($S_{fact} \in [0, 1]$):}
To evaluate factuality of long-form texts, \textsc{FactScore} \citep{min2023factscore} verifies automatically extracted claims against retrieved evidence, and recent work expands this to multilingual texts by translating the non-English responses into English \citep{shafayat2024multifactassessingfactualitymultilingual}. Following this, we translate our questions and answers into English using \textsc{GPT-4o}. Then, we apply the claim extraction and verification pipeline introduced in \citet{song2024veriscore}, which improves the robustness of \textsc{FactScore} by focusing exclusively on verifiable, non-trivial claims and using Google Search to obtain evidence.\footnote{We use Google Search via the Serper API at a total cost of \$510 USD.}\footnote{VeriScore's claim extraction and verification open-source models were run on 1xA40 GPU for 48h.} Finally, for every valid answer (i.e., answer without surface-level issues), we obtain a list of claims, the corresponding top 10 search results, and faithfulness labels (supported or unsupported); see \autoref{fig:claim_extract_pipe} for more details. The $S_{fact}$score will be the fraction of claims that are deemed supported, or 0\% if there are no verifiable claims.

\paragraph{Evaluating relevance ($S_{rel} \in \{0, 1\}$):}
LLM prompting has been shown to have reasonable agreement with human annotations in English and multilingual settings \citep{Hada2023AreLL, Hu2024ThemisAR}. Hence, to evaluate the relevance of long-form answers to their questions, we employ LLM-as-a-Judge \citep{zheng2023judgingllmasajudgemtbenchchatbot} using \gpto{}. That is, we prompt \gpto{} to decide whether each answer is relevant to its question, using the prompt in \autoref{tab:relevance_prompt} with the English translation of the question and answer from our factuality evaluation, at a total cost of \$120 USD.

\paragraph{Overall performance:} We combine three metrics to measure the overall quality of the general answer. We obtain the overall quality score at the instance level $S$ by multiplying the surface-level quality, factuality and relevance scores ($S = S_{surf}*S_{fact}*S_{rel}$).

\begin{table*}[t]
\footnotesize
\centering
\begin{tabularx}{\textwidth}{@{}l*{7}{>{\centering\arraybackslash}X}@{}}
\toprule
\multirow{2}{*}{\textsc{}} & \multicolumn{3}{c}{\textbf{Surface Level}} & \multicolumn{2}{c}{\textbf{Fine-Grained}} & \textbf{Overall} \\ 
\cmidrule(lr){2-4} \cmidrule(lr){5-6} \cmidrule(lr){7-7}
& Wrong Lang. (\%) $\downarrow$ & Repetitions (\%) $\downarrow$ & W/o Issues (\%) $\uparrow$ 
& Factual Pr. (\%) $\uparrow$& Relevance (\%) $\uparrow$ & Overall Score $S$ $\uparrow$ \\
\midrule
\gpto{}         & 2.7 / 1.2 & 7.5 / 1.4 & 90.4 / 97.4 & 69.6 / \textbf{52.2} & 88.2 / 95.7 & \textbf{56.9} / \textbf{49.2} \\
\gptturbo{}     & 3.6 / 1.6 & 3.0 / 0.5 & 93.4 / 97.9& 69.9 / 51.9 & 85.0 / 94.4 & \textbf{56.9} / 48.7 \\
\claude{}       & 4.3 / 1.2 & 0.6 / \textbf{0.1} & \textbf{95.0} / \textbf{98.7}& 63.6 / 45.5 & 84.4 / 93.6 & 52.9 / 42.6 \\
\ayanew{}       & 19.8 / 6.7& 7.9 / 1.8 & 73.4 / 91.7 & 63.8 / 45.6 & 84.7 / 91.9 & 43.4 / 39.5\\
\gemini{}       & \textbf{0.3}*/ \textbf{0.3}*& \textbf{0.1}*/ \textbf{0.1}* & 58.2 / 82.9 & \textbf{71.1} / 48.7 & \textbf{98.2} / 96.5& 40.9 / 46.6 \\
\mixtral{}      & 33.2 / 11.2 & 10.7 / 7.7& 57.2 / 81.3 & 64.0 / 46.2 & 95.5 / 93.0 & 35.6 / 35.7\\
\llama{}        & 76.0 / 70.0 & 0.5 / 0.6  & 23.5 / 29.5 & 66.6 / 46.7 & 97.4  / \textbf{97.5} & 15.3 / 13.5 \\
\bottomrule
\end{tabularx}
\caption{Model performance aggregated across languages. Each cell reports values for culturally agnostic / culturally specific portions. Due to language identification errors, we exclude Balochi, Kirundi, Papiamento, and Hiligaynon from the aggregation. Fine-grained metrics are only computed over answers that lack surface-level issues. *\gemini{} returned API errors for 41.4\% (agnostic) /  16.7\% (specific) of answers, which likely obscures surface-level errors that it makes.}
\vspace{-1em}
\label{tab:main_automatic}
\end{table*}

\subsection{Results of automatic evaluation}

 \autoref{tab:main_automatic} reports micro-averaged automatic metrics of each model on culturally agnostic and culturally specific sets, respectively. 

 \paragraph{Answers to culturally agnostic questions are more factual:} Generated answers to culturally agnostic questions tend to be more factual (64\%--71\%) than answers to culturally specific questions (45\%--52\%).\footnote{Models generate a similar number of factual claims on average for both culturally specific and culturally agnostic questions, with the former yielding slightly lower mean claim counts (see \autoref{fig:mean_claim_count}).} By contrast, surface issues and relevance are relatively consistent between culturally specific and culturally agnostic questions.

\begin{figure*}[tbp]
\centering
  \adjustbox{trim=1.2cm 1.3cm 2cm 1cm}{%
    \includeinkscape[width=1.2\linewidth]{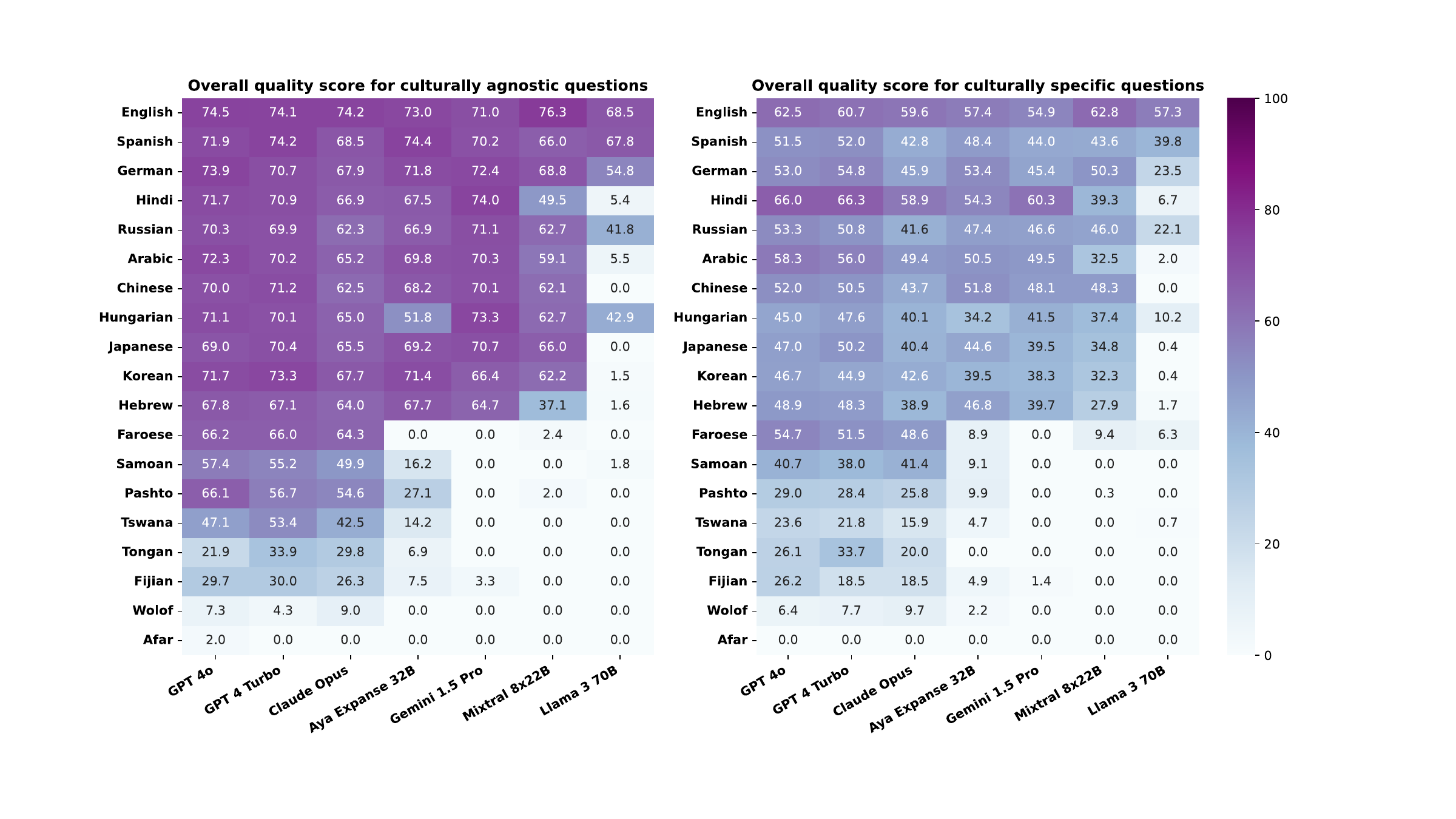}%
  }
  \caption{Answer scores $S$ based on our quality criteria: surface issues, factuality and relevance. The left heatmap shows the results for culturally agnostic questions while the right heatmap shows the results for culturally specific questions. Closed- and open-weight models perform comparably on high- to mid- resource languages, while open-weight models are much worse on low-resource languages. Scores degrade on culturally specific questions due to factual imprecision (see \autoref{fig:factuality_by_lang}).}
  \vspace{-1em}
  \label{fig:overall_by_lang}
\end{figure*}

\paragraph{Open-weight models perform much worse than closed-weight models in low-resource languages:} \autoref{fig:overall_by_lang} shows the overall scores for each model by language. Open-weight models are comparable to their closed counterparts on high- and mid-resource languages, with \ayanew{} outperforming \claude{} in 8 of these languages on culturally agnostic questions. The closed models significantly outperform the open models on the low-resource languages, scoring mostly 22 -- 66 while the open models mostly score below 10. This gap is attributed to surface-level issues, which are present in as high as 70\% for \llama{} (see \autoref{tab:main_automatic}). The exception is \gemini{}, which throwing API errors when prompted in low-resource languages.

We show specific categories of answer deficiencies detected by our surface-level issue metrics in \autoref{tab:faulty_generations_examples}, and provide further analysis in Appendix~\ref{app:auto_eval_analysis}.



\section{Human Evaluation}    \label{sec:human_eval}
Given the limitations of automatic metrics, we supplement our evaluation with native speaker judgments across five languages: Kirundi, Fijian, Hindi, German, and English.

\paragraph{Evaluation setup:} We evaluate \claude, \gptturbo, and \mixtral{}. For each language we sampled 10 culturally specific and 10 culturally agnostic questions.\footnote{For culturally specific questions, annotators selected 10 questions they were confident in answering accurately. For culturally agnostic questions, we supplied annotators with bullet-point answers in English.}

\begin{figure*}[t]
\centering
\includegraphics[width=0.95\textwidth]{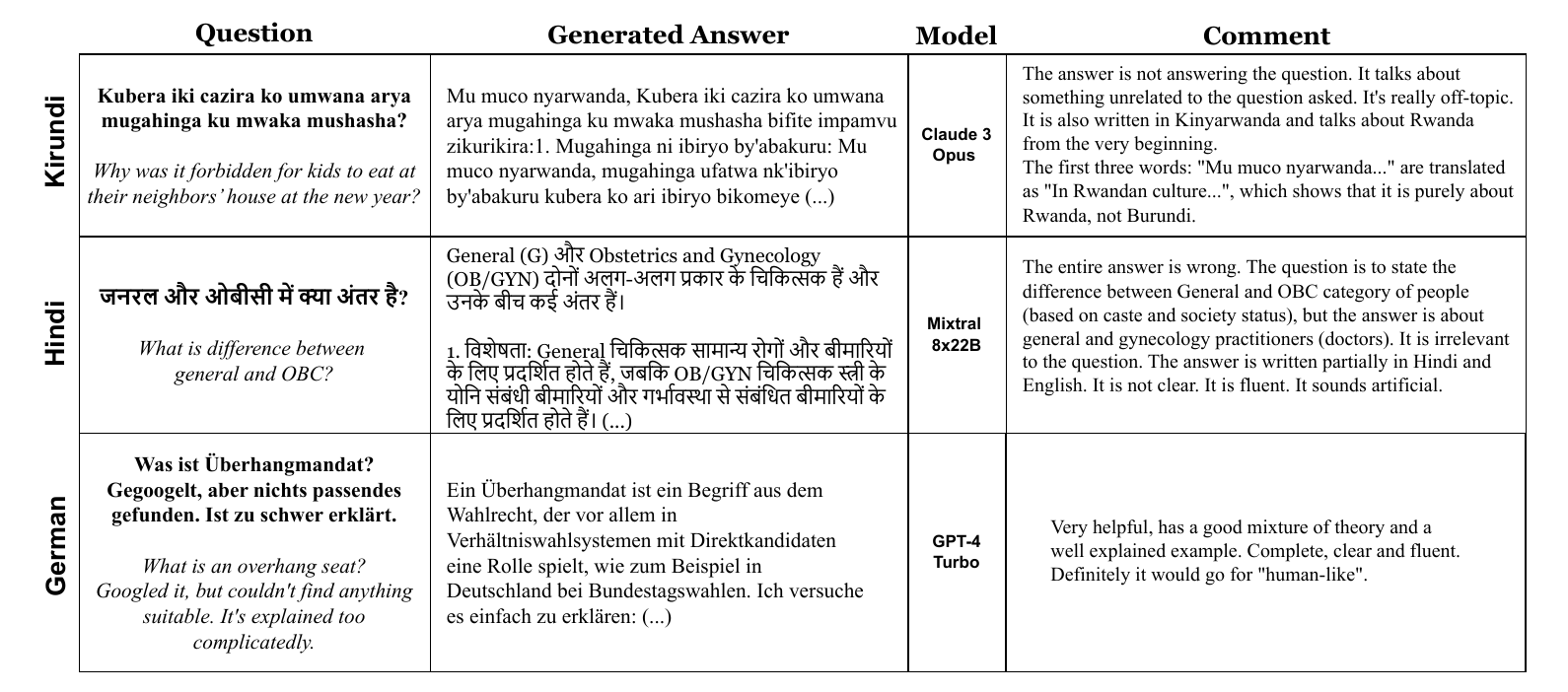} \vspace{-1em}
\caption{Examples of comments on LLM-generated answers written by human annotators.}\vspace{-0.5em}
\label{fig:model_gen_ans_with_comments}
\vspace{-0.2em}
\end{figure*}

We recruit native speakers via Prolific and Upwork, all of whom participated in the question collection process, paying \$7.50 USD per question and an additional \$8.00 USD for reviewing the guidelines, totaling \$720 USD. 
Annotators are presented with a question, reference answer (if available), and answers generated by the three models in random order. For each candidate answer, they are asked to: (1) identify whether it is in the correct language, (2) mark minor and major errors,\footnote{This step was included to help the annotators visualize any issues with the answer and encourage them to read the entire answer. Hence, we did not require annotators to classify errors beyond a simple minor vs major distinction.} (3) evaluate factuality, (4) note significant omissions, (5) comment on the answer's overall quality (\autoref{fig:model_gen_ans_with_comments}), and (6) rate it on a 5-point scale (excellent, good, average, poor, unusable).
Finally, annotators rank the three answers from best to worst and provide a free-form explanation for their ranking. We provide details of the workflow in \autoref{fig:humeval} and \S\ref{app:hum_eval_details}.  
The study was reviewed by the Institutional Review Board and received a \textit{non-human subject} determination.

\subsection{Results of human evaluation} Looking at the overall answer ratings, human annotators prefer \gptturbo{}'s answers, followed by \claude{}'s  and then lastly \mixtral's (\autoref{fig:ratings_by_model_by_qtype}). To confirm, we fit a cumulative link mixed model (\texttt{clmm()}) for predicting ratings from models (\autoref{tab:clmm_model}), with annotators nested within language included as a random effect.\footnote{We use \texttt{clmm} from the \texttt{ordinal} package \cite{clmm2023} because of the ordinal nature of our response variable (ratings) and repeated measures, with annotators rating each model multiple times for different questions. 
} We find that a \mixtral{} answer has an $88\%$ chance of having a lower rating than a \claude{} answer (\textit{p}<.001) and a $94\%$ chance of having a lower rating than a \gptturbo{} answer (\textit{p}<.001). Also, a \claude{} answer has a $30\%$ chance of having a lower rating than a \gptturbo{} answer (\textit{p}<.001).

\paragraph{Answer ratings are lower for culturally specific questions:} \autoref{fig:ratings_by_model_by_qtype} suggests that LLMs generate worse answers for culturally specific questions than for culturally agnostic questions. To check this, we fit a cumulative link mixed model for predicting ratings from question type (Table \ref{app:stats_agnos_sign_better_score}), with annotators nested within language included as a random effect. We see that an answer to a culturally agnostic question has a $67\%$ chance of having a higher rating than an answer to a culturally specific question (\textit{p}<.001). \claude's performance drop on culturally specific questions is notable: its answer to a culturally specific question has an 80\% chance of receiving a lower rating compared to a culturally agnostic question (\textit{p}<.001).

\paragraph{Factuality and omission issues are strong predictors of answer rating:} To determine which variables of this experiment (e.g., model, question type, factuality issues, omissions) correlate with answer rating, we fit cumulative link mixed models for predicting the rating, with each variable being used as the sole predictor of a separate model. \autoref{app:r_squared_diff_predictors} shows R\textsuperscript{2} values of these models. We observe high marginal R\textsuperscript{2} for the factuality issues model (R\textsuperscript{2}=0.560) and the omissions model (R\textsuperscript{2}=0.740), indicating that these factors are strong predictors of answer rating.\footnote{In case of mixed effects models, \textit{marginal} R\textsuperscript{2} refers to the proportion of variance explained by the fixed effects (predictors) alone. 
}

\begin{figure*}
\centering
\includegraphics[width=0.95\textwidth]{figures/calmqaplot_v2.pdf} \vspace{-1.3em}
\caption{Distribution of human ratings of answers by model and question type. Each model generates 50 answers per question type. Humans give higher ratings for culturally agnostic answers, especially for \claude.
}\vspace{-1em}
\label{fig:ratings_by_model_by_qtype}
\end{figure*}

\subsection{Analyzing annotator comments} 
We analyze annotators' comments to gain insights into answer quality. For each comment field, we iteratively develop and apply an annotation schema, linking the results to the corresponding ratings and scores. (See Appendix \S\ref{app:hum_eval_details} for details.)


\paragraph{Factuality errors are more frequent for culturally specific answers:} All 12 issues regarding incorrect dates, entity and events (e.g., ``It is mentioned that Nifty was launched in 1995 but it was actually launched in 1996.'') occur in culturally specific answers, likely due to a greater prevalence of dates, entities and events in culturally specific questions about topics like \textsc{History} than culturally agnostic topics like \textsc{Health and Wellness}.

\paragraph{\gptturbo{} answers rank first due to content.} We analyze the reasons mentioned for ranking each model's answers as best. Having good \textsc{Content} (e.g. due to being complete; see \autoref{tab:categories_of_rank_annotations} for a description) is a reason for \gptturbo{} being chosen as best answer $51\%$ of the time (e.g. ``\textit{Answer 1 (\gptturbo) is the perfect answer and and explains all the points needed to understand how to play the game `Teen Patti'.''}). In the culturally agnostic setting, where \claude{} and \gptturbo{} perform comparably, more \gptturbo{} wins ($48\%$) are attributed to \textsc{Content} than \claude{} wins ($32\%$). The full result can be found in \S\ref{app:hum_eval_details} (\autoref{tab:model_winrate_reason}).

\section{Related Work}

\paragraph{Cultural \& Multilingual NLP:} 

Cultural knowledge has been explored through the creation of knowledge bases \citep{fung2024massively, Nguyen2022ExtractingCC} as well as datasets for tasks like probing \citep{keleg2023dlama, yin-etal-2022-geomlama, Zhou2024DoesMT}, short-form QA and visual QA. Short-form QA work for multilingual cultural knowledge includes MMLU \citep{Hendrycks2020MeasuringMM} translations or MMLU-style datasets 
\citep{singh2024globalmmluunderstandingaddressing, lai2023okapiinstructiontunedlargelanguage, Kim2024CLIcKAB, Koto2024ArabicMMLUAM},
common sense datasets \citep{myung2025blendbenchmarkllmseveryday, Wibowo2023COPALIDIL, Koto2024IndoCultureEG}, and evaluations \citep{shen-etal-2024-understanding}. Visual long-form QA (LVQA) is less explored and mostly monolingual \cite{yu2024mmvetevaluatinglargemultimodal, alwajih-etal-2024-peacock}, but the contemporaneous work \citet{vayani2024languagesmatterevaluatinglmms} looks at LVQA in 100 languages. We are not aware of any textual LFQA datasets of cultural knowledge.

Some multilingual cultural works rely on translation for their multilinguality ~\citep{singh2024globalmmluunderstandingaddressing}, potentially limiting their coverage of cultural concepts. Surveys \citep{adilazuarda2024measuringmodelingculturellms, Liu2024CulturallyAA} call out a lack of multilingual datasets that cover a diverse set of cultural concepts. Our work seeks to make progress in this gap of the literature.


\paragraph{Evaluation of Long-Form QA:}

Evaluating long-form QA (LFQA) remains challenging. Lexical metrics of text generation like ROUGE~\citep{Lin2004ROUGEAP} and some neural-based metrics like BERTScore~\citep{Zhang2019BERTScoreET} and BLEURT \citep{Sellam2020BLEURTLR} show poor correlation with human ratings~\citep{krishna2021hurdles, xu2023critical, Cambazoglu2021QuantifyingHA, Chen2019EvaluatingQA}. For most other model-based evaluations \citep{zheng2023judgingllmasajudgemtbenchchatbot, Fu2023GPTScoreEA, Zhong2022TowardsAU}, correlation with human annotations is measured for tasks like instruction-following, summarization and machine translation but mostly not LFQA. 
\citet{Jiang2023TIGERScoreTB} assess effectiveness of metrics for LFQA, however this is done only on \gpt-created data.



\section{Conclusion \& Future Work}

We introduce \name{}, the first textual multilingual long-form QA dataset, which contains 51.7K culturally specific questions across 23 high- and low-resource languages. Our evaluation of seven state-of-the-art LLMs reveals that culturally specific questions are more difficult for models than culturally agnostic ones, evidenced by lower factuality and human ratings. Furthermore, we observe critical surface-level issues (wrong language, repetition) in all models, especially for low-resource languages. Our results stress the importance of diversifying pre- and post-training datasets to emphasize cultural knowledge acquisition, which can help improve culturally specific QA. Also, improving cross-lingual transfer to address data scarcity may help for underrepresented languages like Afar.

\section*{Limitations}

While we strive to cover as many aspects of the cultures represented in \name{} as possible, we acknowledge that it is not feasible to encompass every cultural nuance. Additionally, for low-resource languages, we employed workers to manually write questions, which impacts scalability. Finally, our culturally agnostic questions are translations from English performed by \gptturbo{}, and thus may not match the quality of human translations.

It would be ideal to have identical distributions of topics across language and type (culturally specific vs culturally agnostic). However, topics like religion, food \& drinks, history and literature, among many others, are naturally bound to the culture, making it impossible to have similar distributions for culturally specific and culturally agnostic questions. Moreover, such topics may have different relative significance for different cultures. Consequently, collecting questions representative of the topics important to people conflicts with having identical distributions between languages. Nevertheless, we found that the topic distribution is similar between languages.

Our automatic evaluation relies on surface-level measures such as language detection and token repetitions. While this approach allows us to determine that current LLMs still struggle with producing outputs in the correct language and without repetitions, it does not assess the fluency or completeness of outputs that lack these surface-level issues. This underscores the need for comprehensive metrics to evaluate overall answer quality in multilingual LFQA, which we leave to future work.

We assess factuality of model generated answers by translating them into English and extracting verifiable claims and validating them against evidence retrieved through web searches. However, this evaluation is influenced by three factors: (1) the quality of translation, (2) the quantity of extracted claims and (3) the availability of relevant online evidence. Our relevance evaluation also depends on the quality of translation. While we do not observe any evident issues with our pipelines during data inspection, it is possible then these factors influenced the results.

Our human evaluation uses 100 questions across 5 languages to demonstrate that models struggle to generate well-written, factual, and complete answers in non-English languages. Large-scale human evaluation is time-consuming and prohibitively expensive, and finding workers proficient in low-resource languages presented a significant challenge, constraining our evaluation efforts. However, we have shown that we can statistically justify various insights about LLM multilingual capabilities with our scale of data.

\section*{Ethical consideration}

The protocols for data collection and human evaluation described in this paper were reviewed and deemed \textit{exempt} by the Institutional Review Board. All annotators provided informed consent for the use and publication of their annotations and collected questions. They were compensated fairly for their work, with their preferred rates respected for both the question collection and evaluation tasks.
\section*{Acknowledgments}
We extend our gratitude to the annotators from Prolific and Upwork for their hard work and for sharing their expertise about their culture. This project was supported by NSF grant RI-2312948.

\bibliography{custom}

\begin{thebibliography}{69}
\providecommand{\natexlab}[1]{#1}

\bibitem[{Adilazuarda et~al.(2024)Adilazuarda, Mukherjee, Lavania, Singh, Aji, O'Neill, Modi, and Choudhury}]{adilazuarda2024measuringmodelingculturellms}
Muhammad~Farid Adilazuarda, Sagnik Mukherjee, Pradhyumna Lavania, Siddhant Singh, Alham~Fikri Aji, Jacki O'Neill, Ashutosh Modi, and Monojit Choudhury. 2024.
\newblock \href {https://arxiv.org/abs/2403.15412} {Towards measuring and modeling "culture" in llms: A survey}.
\newblock \emph{Preprint}, arXiv:2403.15412.

\bibitem[{Ahia et~al.(2023)Ahia, Kumar, Gonen, Kasai, Mortensen, Smith, and Tsvetkov}]{Ahia2023DoAL}
Orevaoghene Ahia, Sachin Kumar, Hila Gonen, Jungo Kasai, David~R. Mortensen, Noah~A. Smith, and Yulia Tsvetkov. 2023.
\newblock \href {https://api.semanticscholar.org/CorpusID:258841465} {Do all languages cost the same? tokenization in the era of commercial language models}.
\newblock \emph{ArXiv}, abs/2305.13707.

\bibitem[{AI@Meta(2024)}]{llama3modelcard}
AI@Meta. 2024.
\newblock \href {https://github.com/meta-llama/llama3/blob/main/MODEL_CARD.md} {Llama 3 model card}.

\bibitem[{AlKhamissi et~al.(2024)AlKhamissi, ElNokrashy, Alkhamissi, and Diab}]{alkhamissi-etal-2024-investigating}
Badr AlKhamissi, Muhammad ElNokrashy, Mai Alkhamissi, and Mona Diab. 2024.
\newblock \href {https://doi.org/10.18653/v1/2024.acl-long.671} {Investigating cultural alignment of large language models}.
\newblock In \emph{Proceedings of the 62nd Annual Meeting of the Association for Computational Linguistics (Volume 1: Long Papers)}, pages 12404--12422, Bangkok, Thailand. Association for Computational Linguistics.

\bibitem[{Alonso et~al.(2024)Alonso, Oronoz, and Agerri}]{alonso2024medexpqa}
Iñigo Alonso, Maite Oronoz, and Rodrigo Agerri. 2024.
\newblock \href {https://arxiv.org/abs/2404.05590} {Medexpqa: Multilingual benchmarking of large language models for medical question answering}.
\newblock \emph{Preprint}, arXiv:2404.05590.

\bibitem[{Alwajih et~al.(2024)Alwajih, Nagoudi, Bhatia, Mohamed, and Abdul-Mageed}]{alwajih-etal-2024-peacock}
Fakhraddin Alwajih, El~Moatez~Billah Nagoudi, Gagan Bhatia, Abdelrahman Mohamed, and Muhammad Abdul-Mageed. 2024.
\newblock \href {https://doi.org/10.18653/v1/2024.acl-long.689} {Peacock: A family of {A}rabic multimodal large language models and benchmarks}.
\newblock In \emph{Proceedings of the 62nd Annual Meeting of the Association for Computational Linguistics (Volume 1: Long Papers)}, pages 12753--12776, Bangkok, Thailand. Association for Computational Linguistics.

\bibitem[{{Anthropic}(2024)}]{anthropic-claude}
{Anthropic}. 2024.
\newblock \href {https://www-cdn.anthropic.com/de8ba9b01c9ab7cbabf5c33b80b7bbc618857627/Model_Card_Claude_3.pdf} {{The Claude 3 Model Family: Opus, Sonnet, Haiku}}.
\newblock Technical report, Anthropic.
\newblock Accessed: 2024-05-23.

\bibitem[{Artetxe et~al.(2020)Artetxe, Ruder, and Yogatama}]{Artetxe_2020_xquad}
Mikel Artetxe, Sebastian Ruder, and Dani Yogatama. 2020.
\newblock \href {https://doi.org/10.18653/v1/2020.acl-main.421} {On the cross-lingual transferability of monolingual representations}.
\newblock In \emph{Proceedings of the 58th Annual Meeting of the Association for Computational Linguistics}. Association for Computational Linguistics.

\bibitem[{Cambazoglu et~al.(2021)Cambazoglu, Bolotova-Baranova, Scholer, Sanderson, Tavakoli, and Croft}]{Cambazoglu2021QuantifyingHA}
Berkant~Barla Cambazoglu, Valeriia Bolotova-Baranova, Falk Scholer, Mark Sanderson, Leila Tavakoli, and W.~Bruce Croft. 2021.
\newblock \href {https://api.semanticscholar.org/CorpusID:232066842} {Quantifying human-perceived answer utility in non-factoid question answering}.
\newblock \emph{Proceedings of the 2021 Conference on Human Information Interaction and Retrieval}.

\bibitem[{Chen et~al.(2019)Chen, Stanovsky, Singh, and Gardner}]{Chen2019EvaluatingQA}
Anthony Chen, Gabriel Stanovsky, Sameer Singh, and Matt Gardner. 2019.
\newblock \href {https://api.semanticscholar.org/CorpusID:207901226} {Evaluating question answering evaluation}.
\newblock In \emph{Conference on Empirical Methods in Natural Language Processing}.

\bibitem[{Christensen(2023)}]{clmm2023}
Rune H.~B. Christensen. 2023.
\newblock \href {https://CRAN.R-project.org/package=ordinal} {\emph{ordinal---Regression Models for Ordinal Data}}.
\newblock R package version 2023.12-4.

\bibitem[{Clark et~al.(2020)Clark, Choi, Collins, Garrette, Kwiatkowski, Nikolaev, and Palomaki}]{clark2020tydi}
Jonathan~H. Clark, Eunsol Choi, Michael Collins, Dan Garrette, Tom Kwiatkowski, Vitaly Nikolaev, and Jennimaria Palomaki. 2020.
\newblock \href {https://arxiv.org/abs/2003.05002} {Tydi qa: A benchmark for information-seeking question answering in typologically diverse languages}.
\newblock \emph{Preprint}, arXiv:2003.05002.

\bibitem[{Dang et~al.(2024)Dang, Singh, D'souza, Ahmadian, Salamanca, Smith, Peppin, Hong, Govindassamy, Zhao, Kublik, Amer, Aryabumi, Campos, Tan, Kocmi, Strub, Grinsztajn, Flet-Berliac, Locatelli, Lin, Talupuru, Venkitesh, Cairuz, Yang, Chung, Ko, Shi, Shukayev, Bae, Piktus, Castagné, Cruz-Salinas, Kim, Crawhall-Stein, Morisot, Roy, Blunsom, Zhang, Gomez, Frosst, Fadaee, Ermis, Üstün, and Hooker}]{dang2024ayaexpansecombiningresearch}
John Dang, Shivalika Singh, Daniel D'souza, Arash Ahmadian, Alejandro Salamanca, Madeline Smith, Aidan Peppin, Sungjin Hong, Manoj Govindassamy, Terrence Zhao, Sandra Kublik, Meor Amer, Viraat Aryabumi, Jon~Ander Campos, Yi-Chern Tan, Tom Kocmi, Florian Strub, Nathan Grinsztajn, Yannis Flet-Berliac, Acyr Locatelli, Hangyu Lin, Dwarak Talupuru, Bharat Venkitesh, David Cairuz, Bowen Yang, Tim Chung, Wei-Yin Ko, Sylvie~Shang Shi, Amir Shukayev, Sammie Bae, Aleksandra Piktus, Roman Castagné, Felipe Cruz-Salinas, Eddie Kim, Lucas Crawhall-Stein, Adrien Morisot, Sudip Roy, Phil Blunsom, Ivan Zhang, Aidan Gomez, Nick Frosst, Marzieh Fadaee, Beyza Ermis, Ahmet Üstün, and Sara Hooker. 2024.
\newblock \href {https://arxiv.org/abs/2412.04261} {Aya expanse: Combining research breakthroughs for a new multilingual frontier}.
\newblock \emph{Preprint}, arXiv:2412.04261.

\bibitem[{Durmus et~al.(2024)Durmus, Nguyen, Liao, Schiefer, Askell, Bakhtin, Chen, Hatfield-Dodds, Hernandez, Joseph, Lovitt, McCandlish, Sikder, Tamkin, Thamkul, Kaplan, Clark, and Ganguli}]{durmus2024measuringglobal}
Esin Durmus, Karina Nguyen, Thomas~I. Liao, Nicholas Schiefer, Amanda Askell, Anton Bakhtin, Carol Chen, Zac Hatfield-Dodds, Danny Hernandez, Nicholas Joseph, Liane Lovitt, Sam McCandlish, Orowa Sikder, Alex Tamkin, Janel Thamkul, Jared Kaplan, Jack Clark, and Deep Ganguli. 2024.
\newblock \href {https://arxiv.org/abs/2306.16388} {Towards measuring the representation of subjective global opinions in language models}.
\newblock \emph{Preprint}, arXiv:2306.16388.

\bibitem[{Fan et~al.(2019)Fan, Jernite, Perez, Grangier, Weston, and Auli}]{fan2019eli5}
Angela Fan, Yacine Jernite, Ethan Perez, David Grangier, Jason Weston, and Michael Auli. 2019.
\newblock \href {https://arxiv.org/abs/1907.09190} {Eli5: Long form question answering}.
\newblock \emph{Preprint}, arXiv:1907.09190.

\bibitem[{Fu et~al.(2023)Fu, Ng, Jiang, and Liu}]{Fu2023GPTScoreEA}
Jinlan Fu, See-Kiong Ng, Zhengbao Jiang, and Pengfei Liu. 2023.
\newblock \href {https://api.semanticscholar.org/CorpusID:256662188} {Gptscore: Evaluate as you desire}.
\newblock In \emph{North American Chapter of the Association for Computational Linguistics}.

\bibitem[{Fung et~al.(2024)Fung, Zhao, Doo, Sun, and Ji}]{fung2024massively}
Yi~Fung, Ruining Zhao, Jae Doo, Chenkai Sun, and Heng Ji. 2024.
\newblock \href {https://arxiv.org/abs/2402.09369} {Massively multi-cultural knowledge acquisition \& lm benchmarking}.
\newblock \emph{Preprint}, arXiv:2402.09369.

\bibitem[{Gao et~al.(2021)Gao, Zhou, and Qiu}]{eli5-category}
Jingsong Gao, Qingren Zhou, and Rui Qiu. 2021.
\newblock {ELI5-Category:} a categorized open-domain qa dataset.

\bibitem[{{Gemini Team}(2024)}]{geminiteam2024gemini}
{Gemini Team}. 2024.
\newblock \href {https://arxiv.org/abs/2403.05530} {Gemini 1.5: Unlocking multimodal understanding across millions of tokens of context}.
\newblock \emph{Preprint}, arXiv:2403.05530.

\bibitem[{Guerreiro et~al.(2023)Guerreiro, Alves, Waldendorf, Haddow, Birch, Colombo, and Martins}]{Guerreiro2023HallucinationsIL}
Nuno~M. Guerreiro, Duarte~M. Alves, Jonas Waldendorf, Barry Haddow, Alexandra Birch, Pierre Colombo, and Andr{\'e} Martins. 2023.
\newblock \href {https://api.semanticscholar.org/CorpusID:257771892} {Hallucinations in large multilingual translation models}.
\newblock \emph{Transactions of the Association for Computational Linguistics}, 11:1500--1517.

\bibitem[{Hada et~al.(2023)Hada, Gumma, de~Wynter, Diddee, Ahmed, Choudhury, Bali, and Sitaram}]{Hada2023AreLL}
Rishav Hada, Varun Gumma, Adrian de~Wynter, Harshita Diddee, Mohamed Ahmed, Monojit Choudhury, Kalika Bali, and Sunayana Sitaram. 2023.
\newblock \href {https://api.semanticscholar.org/CorpusID:261822638} {Are large language model-based evaluators the solution to scaling up multilingual evaluation?}
\newblock In \emph{Findings}.

\bibitem[{Hendrycks et~al.(2020)Hendrycks, Burns, Basart, Zou, Mazeika, Song, and Steinhardt}]{Hendrycks2020MeasuringMM}
Dan Hendrycks, Collin Burns, Steven Basart, Andy Zou, Mantas Mazeika, Dawn~Xiaodong Song, and Jacob Steinhardt. 2020.
\newblock \href {https://api.semanticscholar.org/CorpusID:221516475} {Measuring massive multitask language understanding}.
\newblock \emph{ArXiv}, abs/2009.03300.

\bibitem[{Hu et~al.(2024)Hu, Lin, Gao, Yin, and Wan}]{Hu2024ThemisAR}
Xinyu Hu, Li~Lin, Mingqi Gao, Xunjian Yin, and Xiaojun Wan. 2024.
\newblock \href {https://api.semanticscholar.org/CorpusID:270737769} {Themis: A reference-free nlg evaluation language model with flexibility and interpretability}.
\newblock In \emph{Conference on Empirical Methods in Natural Language Processing}.

\bibitem[{its5Q(2022)}]{its5Qyandex}
its5Q. 2022.
\newblock \href {https://huggingface.co/datasets/its5Q/yandex-q} {its5q/yandex-q - datasets at hugging face}.

\bibitem[{Jiang et~al.(2023)Jiang, Li, Zhang, Huang, Lin, and Chen}]{Jiang2023TIGERScoreTB}
Dongfu Jiang, Yishan Li, Ge~Zhang, Wenhao Huang, Bill~Yuchen Lin, and Wenhu Chen. 2023.
\newblock \href {https://api.semanticscholar.org/CorpusID:263334281} {Tigerscore: Towards building explainable metric for all text generation tasks}.
\newblock \emph{Trans. Mach. Learn. Res.}, 2024.

\bibitem[{Jiao et~al.(2023)Jiao, Wang, tse Huang, Wang, Shi, and Tu}]{jiao2023chatgptgoodtranslatoryes}
Wenxiang Jiao, Wenxuan Wang, Jen tse Huang, Xing Wang, Shuming Shi, and Zhaopeng Tu. 2023.
\newblock \href {https://arxiv.org/abs/2301.08745} {Is chatgpt a good translator? yes with gpt-4 as the engine}.
\newblock \emph{Preprint}, arXiv:2301.08745.

\bibitem[{Kang et~al.(2024)Kang, Blevins, and Zettlemoyer}]{kang2024comparing}
Haoqiang Kang, Terra Blevins, and Luke Zettlemoyer. 2024.
\newblock \href {https://arxiv.org/abs/2402.10496} {Comparing hallucination detection metrics for multilingual generation}.
\newblock \emph{Preprint}, arXiv:2402.10496.

\bibitem[{Keleg and Magdy(2023)}]{keleg2023dlama}
Amr Keleg and Walid Magdy. 2023.
\newblock \href {https://arxiv.org/abs/2306.05076} {Dlama: A framework for curating culturally diverse facts for probing the knowledge of pretrained language models}.
\newblock \emph{Preprint}, arXiv:2306.05076.

\bibitem[{Kim et~al.(2024)Kim, Suk, Oh, Yoo, Thorne, and Oh}]{Kim2024CLIcKAB}
Eunsu Kim, Juyoung Suk, Philhoon Oh, Haneul Yoo, James Thorne, and Alice Oh. 2024.
\newblock \href {https://api.semanticscholar.org/CorpusID:268357672} {Click: A benchmark dataset of cultural and linguistic intelligence in korean}.
\newblock \emph{ArXiv}, abs/2403.06412.

\bibitem[{Koto et~al.(2021)Koto, Lau, and Baldwin}]{Koto2021EvaluatingTE}
Fajri Koto, Jey~Han Lau, and Timothy Baldwin. 2021.
\newblock \href {https://api.semanticscholar.org/CorpusID:235313819} {Evaluating the efficacy of summarization evaluation across languages}.
\newblock \emph{ArXiv}, abs/2106.01478.

\bibitem[{Koto et~al.(2024{\natexlab{a}})Koto, Li, Shatnawi, Doughman, Sadallah, Alraeesi, Almubarak, Alyafeai, Sengupta, Shehata, Habash, Nakov, and Baldwin}]{Koto2024ArabicMMLUAM}
Fajri Koto, Haonan Li, Sara Shatnawi, Jad Doughman, Abdelrahman~Boda Sadallah, Aisha Alraeesi, Khalid Almubarak, Zaid Alyafeai, Neha Sengupta, Shady Shehata, Nizar Habash, Preslav Nakov, and Timothy Baldwin. 2024{\natexlab{a}}.
\newblock \href {https://api.semanticscholar.org/CorpusID:267760288} {Arabicmmlu: Assessing massive multitask language understanding in arabic}.
\newblock \emph{ArXiv}, abs/2402.12840.

\bibitem[{Koto et~al.(2024{\natexlab{b}})Koto, Mahendra, Aisyah, and Baldwin}]{Koto2024IndoCultureEG}
Fajri Koto, Rahmad Mahendra, Nurul Aisyah, and Timothy Baldwin. 2024{\natexlab{b}}.
\newblock \href {https://api.semanticscholar.org/CorpusID:268856961} {Indoculture: Exploring geographically-influenced cultural commonsense reasoning across eleven indonesian provinces}.
\newblock \emph{ArXiv}, abs/2404.01854.

\bibitem[{Krishna et~al.(2021)Krishna, Roy, and Iyyer}]{krishna2021hurdles}
Kalpesh Krishna, Aurko Roy, and Mohit Iyyer. 2021.
\newblock \href {https://arxiv.org/abs/2103.06332} {Hurdles to progress in long-form question answering}.
\newblock \emph{Preprint}, arXiv:2103.06332.

\bibitem[{Lai et~al.(2023)Lai, Nguyen, Ngo, Nguyen, Dernoncourt, Rossi, and Nguyen}]{lai2023okapiinstructiontunedlargelanguage}
Viet~Dac Lai, Chien~Van Nguyen, Nghia~Trung Ngo, Thuat Nguyen, Franck Dernoncourt, Ryan~A. Rossi, and Thien~Huu Nguyen. 2023.
\newblock \href {https://arxiv.org/abs/2307.16039} {Okapi: Instruction-tuned large language models in multiple languages with reinforcement learning from human feedback}.
\newblock \emph{Preprint}, arXiv:2307.16039.

\bibitem[{Lewis et~al.(2020)Lewis, Oguz, Rinott, Riedel, and Schwenk}]{lewis-etal-2020-mlqa}
Patrick Lewis, Barlas Oguz, Ruty Rinott, Sebastian Riedel, and Holger Schwenk. 2020.
\newblock \href {https://doi.org/10.18653/v1/2020.acl-main.653} {{MLQA}: Evaluating cross-lingual extractive question answering}.
\newblock In \emph{Proceedings of the 58th Annual Meeting of the Association for Computational Linguistics}, pages 7315--7330, Online. Association for Computational Linguistics.

\bibitem[{Li et~al.(2024)Li, Jiang, Dziri, Ren, and Choi}]{li2024culturegen}
Huihan Li, Liwei Jiang, Nouha Dziri, Xiang Ren, and Yejin Choi. 2024.
\newblock \href {https://arxiv.org/abs/2404.10199} {Culture-gen: Revealing global cultural perception in language models through natural language prompting}.
\newblock \emph{Preprint}, arXiv:2404.10199.

\bibitem[{Lin(2004{\natexlab{a}})}]{lin-2004-rouge}
Chin-Yew Lin. 2004{\natexlab{a}}.
\newblock \href {https://aclanthology.org/W04-1013} {{ROUGE}: A package for automatic evaluation of summaries}.
\newblock In \emph{Text Summarization Branches Out}, pages 74--81, Barcelona, Spain. Association for Computational Linguistics.

\bibitem[{Lin(2004{\natexlab{b}})}]{Lin2004ROUGEAP}
Chin-Yew Lin. 2004{\natexlab{b}}.
\newblock \href {https://api.semanticscholar.org/CorpusID:964287} {Rouge: A package for automatic evaluation of summaries}.
\newblock In \emph{Annual Meeting of the Association for Computational Linguistics}.

\bibitem[{Liu et~al.(2024)Liu, Gurevych, and Korhonen}]{Liu2024CulturallyAA}
Chen~Cecilia Liu, Iryna Gurevych, and Anna Korhonen. 2024.
\newblock \href {https://api.semanticscholar.org/CorpusID:270285720} {Culturally aware and adapted nlp: A taxonomy and a survey of the state of the art}.

\bibitem[{Liu et~al.(2019)Liu, Lin, Liu, and Sun}]{liu-etal-2019-xqa}
Jiahua Liu, Yankai Lin, Zhiyuan Liu, and Maosong Sun. 2019.
\newblock \href {https://doi.org/10.18653/v1/P19-1227} {{XQA}: A cross-lingual open-domain question answering dataset}.
\newblock In \emph{Proceedings of the 57th Annual Meeting of the Association for Computational Linguistics}, pages 2358--2368, Florence, Italy. Association for Computational Linguistics.

\bibitem[{Min et~al.(2023)Min, Krishna, Lyu, Lewis, tau Yih, Koh, Iyyer, Zettlemoyer, and Hajishirzi}]{min2023factscore}
Sewon Min, Kalpesh Krishna, Xinxi Lyu, Mike Lewis, Wen tau Yih, Pang~Wei Koh, Mohit Iyyer, Luke Zettlemoyer, and Hannaneh Hajishirzi. 2023.
\newblock \href {https://arxiv.org/abs/2305.14251} {Factscore: Fine-grained atomic evaluation of factual precision in long form text generation}.
\newblock \emph{Preprint}, arXiv:2305.14251.

\bibitem[{{Mistral AI}(2024)}]{mistral-mixtral8x22b}
{Mistral AI}. 2024.
\newblock \href {https://mistral.ai/news/mixtral-8x22b/} {{Cheaper, Better, Faster, Stronger}}.
\newblock Technical report, Mistral AI.
\newblock Accessed: 2024-06-19.

\bibitem[{Myung et~al.(2025)Myung, Lee, Zhou, Jin, Putri, Antypas, Borkakoty, Kim, Perez-Almendros, Ayele, Gutiérrez-Basulto, Ibáñez-García, Lee, Muhammad, Park, Rzayev, White, Yimam, Pilehvar, Ousidhoum, Camacho-Collados, and Oh}]{myung2025blendbenchmarkllmseveryday}
Junho Myung, Nayeon Lee, Yi~Zhou, Jiho Jin, Rifki~Afina Putri, Dimosthenis Antypas, Hsuvas Borkakoty, Eunsu Kim, Carla Perez-Almendros, Abinew~Ali Ayele, Víctor Gutiérrez-Basulto, Yazmín Ibáñez-García, Hwaran Lee, Shamsuddeen~Hassan Muhammad, Kiwoong Park, Anar~Sabuhi Rzayev, Nina White, Seid~Muhie Yimam, Mohammad~Taher Pilehvar, Nedjma Ousidhoum, Jose Camacho-Collados, and Alice Oh. 2025.
\newblock \href {https://arxiv.org/abs/2406.09948} {Blend: A benchmark for llms on everyday knowledge in diverse cultures and languages}.
\newblock \emph{Preprint}, arXiv:2406.09948.

\bibitem[{Naous et~al.(2024)Naous, Ryan, Ritter, and Xu}]{naous-etal-2024-beer}
Tarek Naous, Michael Ryan, Alan Ritter, and Wei Xu. 2024.
\newblock \href {https://doi.org/10.18653/v1/2024.acl-long.862} {Having beer after prayer? measuring cultural bias in large language models}.
\newblock In \emph{Proceedings of the 62nd Annual Meeting of the Association for Computational Linguistics (Volume 1: Long Papers)}, pages 16366--16393, Bangkok, Thailand. Association for Computational Linguistics.

\bibitem[{Nguyen et~al.(2022)Nguyen, Razniewski, Varde, and Weikum}]{Nguyen2022ExtractingCC}
Tuan-Phong Nguyen, Simon Razniewski, Aparna~S. Varde, and Gerhard Weikum. 2022.
\newblock \href {https://api.semanticscholar.org/CorpusID:252907608} {Extracting cultural commonsense knowledge at scale}.
\newblock \emph{Proceedings of the ACM Web Conference 2023}.

\bibitem[{{OpenAI}(2024{\natexlab{a}})}]{openai2024gpt4}
{OpenAI}. 2024{\natexlab{a}}.
\newblock \href {https://arxiv.org/abs/2303.08774} {Gpt-4 technical report}.
\newblock \emph{Preprint}, arXiv:2303.08774.

\bibitem[{{OpenAI}(2024{\natexlab{b}})}]{openai-gpt4o}
{OpenAI}. 2024{\natexlab{b}}.
\newblock \href {https://openai.com/index/hello-gpt-4o/} {{Model release blog: GPT-4o}}.
\newblock Technical report, OpenAI.
\newblock Accessed: 2024-05-23.

\bibitem[{Palta and Rudinger(2023)}]{palta-rudinger-2023-fork}
Shramay Palta and Rachel Rudinger. 2023.
\newblock \href {https://doi.org/10.18653/v1/2023.findings-acl.631} {{FORK}: A bite-sized test set for probing culinary cultural biases in commonsense reasoning models}.
\newblock In \emph{Findings of the Association for Computational Linguistics: ACL 2023}, pages 9952--9962, Toronto, Canada. Association for Computational Linguistics.

\bibitem[{Papineni et~al.(2002)Papineni, Roukos, Ward, and Zhu}]{papineni-etal-2002-bleu}
Kishore Papineni, Salim Roukos, Todd Ward, and Wei-Jing Zhu. 2002.
\newblock \href {https://doi.org/10.3115/1073083.1073135} {{B}leu: a method for automatic evaluation of machine translation}.
\newblock In \emph{Proceedings of the 40th Annual Meeting of the Association for Computational Linguistics}, pages 311--318, Philadelphia, Pennsylvania, USA. Association for Computational Linguistics.

\bibitem[{Qiu et~al.(2023)Qiu, Ziser, Korhonen, Ponti, and Cohen}]{Qiu2023DetectingAM}
Yifu Qiu, Yftah Ziser, Anna Korhonen, E.~Ponti, and Shay~B. Cohen. 2023.
\newblock \href {https://api.semanticscholar.org/CorpusID:258841008} {Detecting and mitigating hallucinations in multilingual summarisation}.
\newblock \emph{ArXiv}, abs/2305.13632.

\bibitem[{Sellam et~al.(2020)Sellam, Das, and Parikh}]{Sellam2020BLEURTLR}
Thibault Sellam, Dipanjan Das, and Ankur~P. Parikh. 2020.
\newblock \href {https://api.semanticscholar.org/CorpusID:215548699} {Bleurt: Learning robust metrics for text generation}.
\newblock In \emph{Annual Meeting of the Association for Computational Linguistics}.

\bibitem[{Shafayat et~al.(2024)Shafayat, Kim, Oh, and Oh}]{shafayat2024multifactassessingfactualitymultilingual}
Sheikh Shafayat, Eunsu Kim, Juhyun Oh, and Alice Oh. 2024.
\newblock \href {https://arxiv.org/abs/2402.18045} {Multi-fact: Assessing factuality of multilingual llms using factscore}.
\newblock \emph{Preprint}, arXiv:2402.18045.

\bibitem[{Shen et~al.(2024)Shen, Logeswaran, Lee, Lee, Poria, and Mihalcea}]{shen-etal-2024-understanding}
Siqi Shen, Lajanugen Logeswaran, Moontae Lee, Honglak Lee, Soujanya Poria, and Rada Mihalcea. 2024.
\newblock \href {https://aclanthology.org/2024.naacl-long.316} {Understanding the capabilities and limitations of large language models for cultural commonsense}.
\newblock In \emph{Proceedings of the 2024 Conference of the North American Chapter of the Association for Computational Linguistics: Human Language Technologies (Volume 1: Long Papers)}, pages 5668--5680, Mexico City, Mexico. Association for Computational Linguistics.

\bibitem[{Singh et~al.(2024)Singh, Romanou, Fourrier, Adelani, Ngui, Vila-Suero, Limkonchotiwat, Marchisio, Leong, Susanto, Ng, Longpre, Ko, Smith, Bosselut, Oh, Martins, Choshen, Ippolito, Ferrante, Fadaee, Ermis, and Hooker}]{singh2024globalmmluunderstandingaddressing}
Shivalika Singh, Angelika Romanou, Clémentine Fourrier, David~I. Adelani, Jian~Gang Ngui, Daniel Vila-Suero, Peerat Limkonchotiwat, Kelly Marchisio, Wei~Qi Leong, Yosephine Susanto, Raymond Ng, Shayne Longpre, Wei-Yin Ko, Madeline Smith, Antoine Bosselut, Alice Oh, Andre F.~T. Martins, Leshem Choshen, Daphne Ippolito, Enzo Ferrante, Marzieh Fadaee, Beyza Ermis, and Sara Hooker. 2024.
\newblock \href {https://arxiv.org/abs/2412.03304} {Global mmlu: Understanding and addressing cultural and linguistic biases in multilingual evaluation}.
\newblock \emph{Preprint}, arXiv:2412.03304.

\bibitem[{Song et~al.(2024)Song, Kim, and Iyyer}]{song2024veriscore}
Yixiao Song, Yekyung Kim, and Mohit Iyyer. 2024.
\newblock \href {https://arxiv.org/abs/2406.19276} {Veriscore: Evaluating the factuality of verifiable claims in long-form text generation}.
\newblock \emph{Preprint}, arXiv:2406.19276.

\bibitem[{Tao et~al.(2024)Tao, Viberg, Baker, and Kizilcec}]{tao2024culturalbiasculturalalignment}
Yan Tao, Olga Viberg, Ryan~S. Baker, and Rene~F. Kizilcec. 2024.
\newblock \href {https://arxiv.org/abs/2311.14096} {Cultural bias and cultural alignment of large language models}.
\newblock \emph{Preprint}, arXiv:2311.14096.

\bibitem[{Tkachenko et~al.(2020-2022)Tkachenko, Malyuk, Holmanyuk, and Liubimov}]{LabelStudio}
Maxim Tkachenko, Mikhail Malyuk, Andrey Holmanyuk, and Nikolai Liubimov. 2020-2022.
\newblock \href {https://github.com/heartexlabs/label-studio} {{Label Studio}: Data labeling software}.
\newblock Open source software available from https://github.com/heartexlabs/label-studio.

\bibitem[{Vayani et~al.(2024)Vayani, Dissanayake, Watawana, Ahsan, Sasikumar, Thawakar, Ademtew, Hmaiti, Kumar, Kuckreja, Maslych, Ghallabi, Mihaylov, Qin, Shaker, Zhang, Ihsani, Esplana, Gokani, Mirkin, Singh, Srivastava, Hamerlik, Izzati, Maani, Cavada, Chim, Gupta, Manjunath, Zhumakhanova, Rabevohitra, Amirudin, Ridzuan, Kareem, More, Li, Shakya, Saad, Ghasemaghaei, Djanibekov, Azizov, Jankovic, Bhatia, Cabrera, Obando-Ceron, Otieno, Farestam, Rabbani, Baliah, Sanjeev, Shtanchaev, Fatima, Nguyen, Kareem, Aremu, Xavier, Bhatkal, Toyin, Chadha, Cholakkal, Anwer, Felsberg, Laaksonen, Solorio, Choudhury, Laptev, Shah, Khan, and Khan}]{vayani2024languagesmatterevaluatinglmms}
Ashmal Vayani, Dinura Dissanayake, Hasindri Watawana, Noor Ahsan, Nevasini Sasikumar, Omkar Thawakar, Henok~Biadglign Ademtew, Yahya Hmaiti, Amandeep Kumar, Kartik Kuckreja, Mykola Maslych, Wafa~Al Ghallabi, Mihail Mihaylov, Chao Qin, Abdelrahman~M Shaker, Mike Zhang, Mahardika~Krisna Ihsani, Amiel Esplana, Monil Gokani, Shachar Mirkin, Harsh Singh, Ashay Srivastava, Endre Hamerlik, Fathinah~Asma Izzati, Fadillah~Adamsyah Maani, Sebastian Cavada, Jenny Chim, Rohit Gupta, Sanjay Manjunath, Kamila Zhumakhanova, Feno~Heriniaina Rabevohitra, Azril Amirudin, Muhammad Ridzuan, Daniya Kareem, Ketan More, Kunyang Li, Pramesh Shakya, Muhammad Saad, Amirpouya Ghasemaghaei, Amirbek Djanibekov, Dilshod Azizov, Branislava Jankovic, Naman Bhatia, Alvaro Cabrera, Johan Obando-Ceron, Olympiah Otieno, Fabian Farestam, Muztoba Rabbani, Sanoojan Baliah, Santosh Sanjeev, Abduragim Shtanchaev, Maheen Fatima, Thao Nguyen, Amrin Kareem, Toluwani Aremu, Nathan Xavier, Amit Bhatkal, Hawau Toyin, Aman Chadha, Hisham Cholakkal,
  Rao~Muhammad Anwer, Michael Felsberg, Jorma Laaksonen, Thamar Solorio, Monojit Choudhury, Ivan Laptev, Mubarak Shah, Salman Khan, and Fahad Khan. 2024.
\newblock \href {https://arxiv.org/abs/2411.16508} {All languages matter: Evaluating lmms on culturally diverse 100 languages}.
\newblock \emph{Preprint}, arXiv:2411.16508.

\bibitem[{Wang(2023)}]{wangrui6ZhihuKOL}
Ray Wang. 2023.
\newblock \href {https://huggingface.co/datasets/wangrui6/Zhihu-KOL} {wangrui6/zhihu-kol - datasets at hugging face}.

\bibitem[{Wibowo et~al.(2023)Wibowo, Fuadi, Nityasya, Prasojo, and Aji}]{Wibowo2023COPALIDIL}
Haryo~Akbarianto Wibowo, Erland~Hilman Fuadi, Made~Nindyatama Nityasya, Radityo~Eko Prasojo, and Alham~Fikri Aji. 2023.
\newblock \href {https://api.semanticscholar.org/CorpusID:264935209} {Copal-id: Indonesian language reasoning with local culture and nuances}.
\newblock \emph{ArXiv}, abs/2311.01012.

\bibitem[{Xu et~al.(2023)Xu, Song, Iyyer, and Choi}]{xu2023critical}
Fangyuan Xu, Yixiao Song, Mohit Iyyer, and Eunsol Choi. 2023.
\newblock \href {https://arxiv.org/abs/2305.18201} {A critical evaluation of evaluations for long-form question answering}.
\newblock \emph{Preprint}, arXiv:2305.18201.

\bibitem[{Yan et~al.(2024)Yan, Yan, Chen, Li, Zhu, and Zhang}]{yan2024gpt4vshumantranslators}
Jianhao Yan, Pingchuan Yan, Yulong Chen, Judy Li, Xianchao Zhu, and Yue Zhang. 2024.
\newblock \href {https://arxiv.org/abs/2407.03658} {Gpt-4 vs. human translators: A comprehensive evaluation of translation quality across languages, domains, and expertise levels}.
\newblock \emph{Preprint}, arXiv:2407.03658.

\bibitem[{Yin et~al.(2022)Yin, Bansal, Monajatipoor, Li, and Chang}]{yin-etal-2022-geomlama}
Da~Yin, Hritik Bansal, Masoud Monajatipoor, Liunian~Harold Li, and Kai-Wei Chang. 2022.
\newblock \href {https://doi.org/10.18653/v1/2022.emnlp-main.132} {{G}eo{MLAMA}: Geo-diverse commonsense probing on multilingual pre-trained language models}.
\newblock In \emph{Proceedings of the 2022 Conference on Empirical Methods in Natural Language Processing}, pages 2039--2055, Abu Dhabi, United Arab Emirates. Association for Computational Linguistics.

\bibitem[{Yu et~al.(2024)Yu, Yang, Li, Wang, Lin, Liu, Wang, and Wang}]{yu2024mmvetevaluatinglargemultimodal}
Weihao Yu, Zhengyuan Yang, Linjie Li, Jianfeng Wang, Kevin Lin, Zicheng Liu, Xinchao Wang, and Lijuan Wang. 2024.
\newblock \href {https://arxiv.org/abs/2308.02490} {Mm-vet: Evaluating large multimodal models for integrated capabilities}.
\newblock \emph{Preprint}, arXiv:2308.02490.

\bibitem[{Zhang et~al.(2023)Zhang, D’Haro, Tang, Shi, Tang, and Li}]{Zhang2023xDialEvalAM}
Chen Zhang, L.~F. D’Haro, Chengguang Tang, Ke~Shi, Guohua Tang, and Haizhou Li. 2023.
\newblock \href {https://api.semanticscholar.org/CorpusID:264128252} {xdial-eval: A multilingual open-domain dialogue evaluation benchmark}.
\newblock In \emph{Conference on Empirical Methods in Natural Language Processing}.

\bibitem[{Zhang et~al.(2019)Zhang, Kishore, Wu, Weinberger, and Artzi}]{Zhang2019BERTScoreET}
Tianyi Zhang, Varsha Kishore, Felix Wu, Kilian~Q. Weinberger, and Yoav Artzi. 2019.
\newblock \href {https://api.semanticscholar.org/CorpusID:127986044} {Bertscore: Evaluating text generation with bert}.
\newblock \emph{ArXiv}, abs/1904.09675.

\bibitem[{Zheng et~al.(2023)Zheng, Chiang, Sheng, Zhuang, Wu, Zhuang, Lin, Li, Li, Xing, Zhang, Gonzalez, and Stoica}]{zheng2023judgingllmasajudgemtbenchchatbot}
Lianmin Zheng, Wei-Lin Chiang, Ying Sheng, Siyuan Zhuang, Zhanghao Wu, Yonghao Zhuang, Zi~Lin, Zhuohan Li, Dacheng Li, Eric~P. Xing, Hao Zhang, Joseph~E. Gonzalez, and Ion Stoica. 2023.
\newblock \href {https://arxiv.org/abs/2306.05685} {Judging llm-as-a-judge with mt-bench and chatbot arena}.
\newblock \emph{Preprint}, arXiv:2306.05685.

\bibitem[{Zhong et~al.(2022)Zhong, Liu, Yin, Mao, Jiao, Liu, Zhu, Ji, and Han}]{Zhong2022TowardsAU}
Ming Zhong, Yang Liu, Da~Yin, Yuning Mao, Yizhu Jiao, Peng Liu, Chenguang Zhu, Heng Ji, and Jiawei Han. 2022.
\newblock \href {https://api.semanticscholar.org/CorpusID:252873117} {Towards a unified multi-dimensional evaluator for text generation}.
\newblock In \emph{Conference on Empirical Methods in Natural Language Processing}.

\bibitem[{Zhou et~al.(2024)Zhou, Karidi, Garneau, Cao, Liu, Chen, and Hershcovich}]{Zhou2024DoesMT}
Li~Zhou, Taelin Karidi, Nicolas Garneau, Yong Cao, Wanlong Liu, Wenyu Chen, and Daniel Hershcovich. 2024.
\newblock \href {https://api.semanticscholar.org/CorpusID:269033101} {Does mapo tofu contain coffee? probing llms for food-related cultural knowledge}.
\newblock \emph{ArXiv}, abs/2404.06833.

\end{thebibliography}

\appendix


\section*{Ethical Considerations}

The protocols for data collection and human evaluation described in this paper were reviewed and deemed \textit{exempt} by the Institutional Review Board. All annotators provided informed consent for the use and publication of their annotations and collected questions. They were compensated fairly for their work, with their preferred rates respected for both the question collection and evaluation tasks.
\section{Data Collection}
\label{app_sec:data_collection}

This appendix provides extra details about the data collection process for \name{}. \S\ref{sec:website_survey} describes the identification of websites used for data collection.  \S\ref{sec:high_resource_collection} outlines the data collection methods for high- and mid-resource languages, and \S\ref{sec:low_resource_collection} details the data collection process for low-resource languages. \autoref{tab:examples} contains example entries from the dataset. \autoref{tab:dataset_statistics} and \autoref{app_tab:language_data} provide more details on the number of questions and languages included in the dataset.

\subsection{Website Survey}  
\label{sec:website_survey}

We conducted a survey to find websites with non-English cultural questions. The instructions outlined the survey's goal, defined a good website, and specified what constitutes a culturally specific question. Our criteria for a good website included:

\begin{itemize}
\item At least 500 answered "good" questions (as defined below). Websites could contain other questions as we could filter them out.
\item Most questions and answers should be in a non-English language.
\item Questions should cover a diverse range of topics, not just one or two broad areas (e.g., fashion, technology).
\item The website should contain culturally specific questions not found on English websites or in English QA datasets.
\item The website should have a large community of contributors with many questions answered.
\end{itemize}

The survey evolved through an iterative process of piloting and refining based on the results.

Survey participants were English-proficient crowdworkers on the Prolific platform (\url{https://www.prolific.com}), whose native language was not English. The survey took about 10 minutes to complete, and we paid \$10 for valid responses, totaling \$510. We considered a response valid if it showed a good-faith effort, even if the website was of insufficient quality or duplicated in another response. From 51 responses, we obtained 4 websites used for question collection. Some websites were rejected despite having good questions because the proportion of good to bad questions was too low for feasible collection. Remaining websites were identified by the authors. See \autoref{app_tab:website_data} for the full list of websites employed.

\begin{table*}[tbp]
\centering
{\renewcommand{\arraystretch}{1.5}
\resizebox{1\textwidth}{!}{%
\begin{tabular}{@{}lllll@{}}
\toprule
\addlinespace
\textsc{Website}    & \textsc{Language}   & \textsc{ISO}  & \textsc{\# Q}   & \textsc{URL}  \\ 
\addlinespace
\midrule
Ejaba & Arabic & ar & 29 & \url{https://www.ejaba.com/} \\
Ujeeb & Arabic & ar & 56 & \url{https://ujeeb.com/} \\
Zhihu  & Chinese & zh & 75 & \url{https://www.zhihu.com/} \\
Reddit ELI5 & English & en & 78 & \url{https://www.reddit.com/r/explainlikeimfive/} \\
Gutefrage & German & de & 96 & \url{https://www.gutefrage.net/} \\
Quora & Hebrew & he & 96 & \url{https://he.quora.com} \\
Let's Diskuss & Hindi & hi & 91 & \url{https://hi.letsdiskuss.com/} \\
Gyakori kérdések & Hungarian & hu & 75 & \url{https://www.gyakorikerdesek.hu/} \\
Yahoo Japan & Japanese & ja & 17 & \url{https://chiebukuro.yahoo.co.jp/} \\
OKWave & Japanese & ja & 58 & \url{https://okwave.jp/} \\
Naver & Korean & ko & 75 & \url{https://kin.naver.com/qna/} \\
Yandex & Russian & ru & 75 & \url{https://yandex.ru/q/} \\
Todoexpertos & Spanish & es & 102 & \url{https://www.todoexpertos.com/} \\
\bottomrule
\end{tabular}%
}
}
\caption{Websites from which cultural questions were obtained, with the number of questions retrieved by website. Multiple websites were used for a given language if workers were struggling with a given website.}
\label{app_tab:website_data}
\end{table*}

\begin{table*}[t]
\footnotesize
\centering
{\renewcommand{\arraystretch}{1.1}
\begin{tabular}{@{}lrrrrrr@{}}
\toprule
\addlinespace
\textsc{Language} & \multicolumn{3}{c}{\textsc{Culturally Specific}} & \multicolumn{3}{c}{\textsc{Culturally Agnostic}} \\
\cmidrule(lr){2-4} \cmidrule(lr){5-7}
 & \textsc{\# Q} & \textsc{Q. Bytes} & \textsc{A. Bytes} & \textsc{\# Q} & \textsc{Q. Bytes} & \textsc{A. Bytes} \\
 &  & \textsc{(avg/std)} & \textsc{(avg/std)} &  & \textsc{(avg/std)} & \textsc{(avg/std)} \\
\addlinespace
\midrule
\addlinespace
\textsc{High- \& Mid-Resource Languages} \\
English & 78 & 275.7 / 189.0 & 674.1 / 475.9 & 51 & 67.1 / 31.7 & 632.3 / 636.9 \\
Arabic & 85 & 74.3 / 61.3 & 2105.0 / 2378.6 & 51 & 108.7 / 56.4 & N/A \\
Chinese & 75 & 193.4 / 329.5 & 588.8 / 939.7 & 51 & 68.1 / 31.4 & N/A \\
German & 96 & 304.6 / 227.4 & 1169.0 / 744.7 & 51 & 82.2 / 39.8 & N/A \\
Hebrew & 96 & 142.5 / 84.2 & 2043.6 / 1934.9 & 51 & 93.0 / 42.9 & N/A \\
Hindi & 91 & 122.4 / 52.8 & 3618.8 / 1867.1 & 51 & 184.2 / 90.3 & N/A \\
Hungarian & 75 & 301.1 / 279.8 & 379.3 / 333.2 & 51 & 82.3 / 38.2 & N/A \\
Japanese & 75 & 512.0 / 359.3 & 920.6 / 637.1 & 51 & 104.3 / 50.6 & N/A \\
Korean & 75 & 126.3 / 138.7 & 1008.6 / 936.3 & 51 & 93.0 / 43.3 & N/A \\
Russian & 75 & 310.3 / 438.3 & 4546.7 / 5067.9 & 51 & 134.6 / 70.8 & N/A \\
Spanish & 102 & 429.9 / 271.1 & 852.0 / 817.9 & 51 & 83.6 / 36.1 & N/A \\
\addlinespace
\textsc{Low-resource languages} \\
Afar & 25 & 43.7 / 16.5 & N/A & 51 & 81.1 / 39.8 & N/A \\
Balochi & 65 & 122.7 / 52.4 & N/A & 51 & 96.1 / 48.5 & N/A \\
Faroese & 30 & 47.8 / 16.6 & N/A & 51 & 75.1 / 34.5 & N/A \\
Fijian & 75 & 75.0 / 36.9 & N/A & 51 & 92.5 / 40.6 & N/A \\
Hiligaynon & 65 & 93.4 / 39.1 & N/A & 51 & 83.6 / 39.7 & N/A \\
Kirundi & 53 & 64.6 / 21.2 & 557.2 / 160.9 & 51 & 88.2 / 43.1 & N/A \\
Papiamento & 10 & 66.8 / 28.5 & N/A & 51 & 74.1 / 35.3 & N/A \\
Pashto & 75 & 64.8 / 26.9 & N/A & 51 & 118.1 / 55.6 & N/A \\
Samoan & 25 & 51.2 / 19.3 & N/A & 51 & 80.5 / 37.6 & N/A \\
Tongan & 10 & 81.2 / 19.2 & N/A & 51 & 102.4 / 47.9 & N/A \\
Tswana & 65 & 87.2 / 43.4 & N/A & 51 & 88.8 / 43.4 & N/A \\
Wolof & 50 & 45.3 / 18.9 & N/A & 51 & 78.2 / 44.1 & N/A \\
\bottomrule
\end{tabular}
}
\caption{Combined data statistics for culturally specific and culturally agnostic questions. For each language, we report the number of questions (\# Q), average and standard deviation of question bytes (Q. Bytes) and answer bytes (A. Bytes) in UTF-8 encoding. Answer bytes for culturally agnostic questions are not available, and are marked as N/A.}
\label{tab:dataset_statistics}
\end{table*}

\subsection{High- and Mid-Resource Culturally Specific Questions}   \label{sec:high_resource_collection}

Culturally specific questions in high-resource languages were collected by workers on the Prolific\footnote{\url{https://www.prolific.com/}} platform from the websites in \autoref{app_tab:website_data}. All crowdworkers were English-proficient with their native language matching the language of their allocated websites. Each collector was required to read guidelines, pass a guidelines understanding test and complete a test pilot of 5 questions in order to qualify for the main task. This protocol was reviewed by the Institutional Review Board. Overall, our process yielded 923 questions across 11 languages with answers at a cost of \$1427 USD.

\paragraph{Guidelines} We provided a guidelines slideshow detailing the rules for selecting questions. The main rules for questions where:
\begin{enumerate}
    \item The question should require long answer.
    \item The question should be culture specific.
    \item A native speaker would ask this [question].
    \item The question should be objective.
    \item Questions should not need pictures/links.
\end{enumerate}

\paragraph{Guidelines Understanding Test} Our guidelines understanding test consisted of a form consisting of 11 multiple-selection multiple-choice graded questions. The first question assessed question was ``Which of these are listed as \textbf{important rules for questions} in the guidelines? (you should select all correct answers)'', which required showing understanding of long-form culturally specific information-seeking questions. The remaining 10 questions were curated examples of questions that each may or may not have had issues. Test takers were required to select all the reasons why a question was not suitable according to the guidelines, or select that the question was suitable. We reviewed test results manually, and accordingly chose which workers to pass. We provided passing workers with the test answers, so that they could learn from their mistakes. We paid workers \$3.33 USD for completing the test.

\paragraph{Main Collection Task}
We asked workers to provide examples of culturally specific questions and real users' answers from the identified websites. We manually reviewed all provided examples, using Google Translate to get English translations of website content. In cases where we deemed that an example did not meet our guidelines, we provided feedback and the worker either clarified how their example met the guidelines or replaced the example. For the final dataset, we used \gptturbo{} with the prompt in \autoref{tab:question_translation_prompt} instead of Google Translate to obtain the English translations of questions. We translated answers using \gpto{}, which was released after we had conducted our human evaluation, with the prompt in \autoref{tab:answer_translation_prompt}. We paid the workers \$0.65-1.33 USD per question.

\begin{table*}
\centering
{\renewcommand{\arraystretch}{1.5}
\resizebox{1\textwidth}{!}{%
\begin{tabular}{p{\textwidth}}
\toprule
You are to evaluate whether a given question is *culturally specific* to \textit{[language]}. A question is considered *culturally specific* to a language if it meets both of the following conditions: \par
1. The question is written in \textit{[language]}. \par
2. **At least one** of the following applies: \par
   - The question refers to topics, concepts, objects, entities, or events that are unique to one or a few cultures associated with speakers of \textit{[language]}. \par
   - The answer to the question may vary depending on the cultural or regional context—meaning that the answer would be specific to the \textit{[language]} culture and may differ when considered in the context of other cultures. \\

Culturally specific topics may include, but are not limited to, areas such as heritage, current events, history, religion, literature, society, travel, cuisine, fauna, economy, education, law, and the arts. **Note:** A question being written in \textit{[language]} does not automatically make it culturally specific. \\

**Example 1 (Culturally Specific to English):** \par
- *Question:* "Why does Eminem give so much praise to Dr Dre? He talks about him like a father in his music." \par
- *Explanation:* This question refers to Eminem and Dr Dre, who are prominent figures in the American music industry—a distinct aspect of American (and related Western) culture. The question's cultural context makes it culturally specific to English. \\

**Example 2 (Culturally Specific to English):** \par
- *Question:* "How do you obtain a gun license?" \par
- *Explanation:* The process for obtaining a gun license differs significantly between countries. In the USA, the process is well-established and culturally significant, while it may differ or be non-existent in other countries. Thus, the question is culturally specific to English when referring to American cultural context. \\

**Example 3 (Not Culturally Specific to English):** \par
- *Question:* "Do ships cause the ocean to be higher than it normally would be?" \par
- *Explanation:* This question is general and universal; its answer does not depend on any cultural or regional context. Therefore, it is not culturally specific. \\

--- \\

**Task:** \par
Assess whether the following question is culturally specific to \textit{[language]} or not. Provide your explanation in English, wrapping it in \textasciigrave<explanation></explanation>\textasciigrave tags. Then, output \textasciigrave<result>PASS</result>\textasciigrave if the question is culturally specific, or \textasciigrave<result>FAIL</result>\textasciigrave if it is not. Your response should contain **only** these two tags and nothing else. \\

**Here is the question to assess:** \par
<question> \par
\textit{[question]} \par
</question> \\
\bottomrule
\end{tabular}
}
\caption{Prompt used with \gptomini{} to filter questions collected from community QA websites for culturally specific questions. Strings in the form \textit{[form]} are placeholders that are replaced at runtime.}
\label{tab:cultural_filter_prompt}
}
\end{table*}

\begin{table*}
\centering
{\renewcommand{\arraystretch}{1.5}
\resizebox{1\textwidth}{!}{%
\begin{tabular}{p{\textwidth}}
\toprule
You are evaluating questions in a dataset for quality. Your task is to determine whether a given question meets the following quality criteria: \\

1. **Language:** The question is written in \textit{[language]}. \par
2. **Long-form Answer:** The question cannot be answered with just a short phrase or entity; it requires a few sentences to answer. \par
3. **Not Asking For Answerer's Opinions:** The question does not ask for opinions, personal experiences, perspectives or recommendations of the **answerer**. A question that exbihits bias or implies a particular view point of the **asker** does **not** violate this criterion. \par
4. **Public Information:** Answering the question does not require access to non-public information. \par
5. **Privacy:** The question does not contain any personally identifiable information (e.g., name, username, phone number, or home address). \\

**Example 1 (Satisfies Criteria):** \par
- *Question:* "Why is Norton hated so much? What makes an antivirus/antimalware program good or bad anyway?" \par
- *Explanation:* 1. The question is in English. 2. The question requires a explanation comprising of multiple sentences. 3. The question does not ask for an opinion, even though it indicates a negative viewpoint towards Norton. 4. Answering the question does not require access to non-public information. 5. The question does not contain any personally identifiable information. The question meets all the criteria and so is satisfactory. \\

**Example 2 (Does Not Satisfy Criteria):** \par
- *Question:* "How would you suggest I revise mathematics before my first economics class?" \par
- *Explanation:* The question is explicitly asking for a recommendation and so does not meet the quality critera. \\

--- \\

**Task:**  
Assess whether the following question satisfies all of the quality criteria listed above. Provide a detailed explanation of your assessment in English, wrapped in \textasciigrave<explanation></explanation>\textasciigrave tags. Then, output \textasciigrave<result>PASS</result>\textasciigrave if the question satisfies the quality criteria, or \textasciigrave<result>FAIL</result>\textasciigrave if it does not. Do not output anything outside of the \textasciigrave<explanation></explanation>\textasciigrave and \textasciigrave<result></result>\textasciigrave tags. \\

**Here is the question to assess:** \par
<question> \par
\textit{[question]} \par
</question> \\
\bottomrule
\end{tabular}
}
\caption{Prompt used with \gptomini{} to filter questions collected from community QA websites based on general quality criteria. Strings in the form \textit{[form]} are placeholders that are replaced at runtime.}
\label{tab:general_filter_prompt}
}
\end{table*}

\begin{table*}
\centering
{\renewcommand{\arraystretch}{1.5}
\resizebox{1\textwidth}{!}{%
\begin{tabular}{p{\textwidth}}
\toprule
Your task is to translate a question from \textit{[language]} into English. You will be given the \textit{[language]} answer as the context.
\vspace{\baselineskip}

Here is the \textit{[language]} answer. Use it as the context to make the translation sound natural in the English:
\textit{[answer]}
\vspace{\baselineskip}

Translate the following question from \textit{[language]} into English. Make it sound as natural as possible:
\textit{[question]} \\
\bottomrule
\end{tabular}
}
\caption{Prompt used with \gptturbo{} to translate non-English questions into English. Strings in the form \textit{[form]} are placeholders that are replaced at runtime.}
\label{tab:question_translation_prompt}
}
\end{table*}

\begin{table*}
\centering
{\renewcommand{\arraystretch}{1.5}
\resizebox{1\textwidth}{!}{%
\begin{tabular}{p{\textwidth}}
\toprule
Your task is to translate the answer of a \textit{[language]} question from \textit{[language]} into English. You will be given the \textit{[language]} question as the context.
\vspace{\baselineskip}

Here is the \textit{[language]} question. Use it as the context to make the translation sound natural in the English:
\textit{[question]}
\vspace{\baselineskip}

Translate the following answer from \textit{[language]} into English. Make it sound as natural as possible:
\textit{[answer]} \\
\bottomrule
\end{tabular}
}
\caption{Prompt used with \gpto{} to translate non-English answers into English. Strings in the form \textit{[form]} are placeholders that are replaced at runtime.}
\label{tab:answer_translation_prompt}
}
\end{table*}

\subsection{Low-Resource Culturally Specific Questions}   
\label{sec:low_resource_collection}
Questions for low-resource languages were collected by hiring native speakers proficient in English through Upwork. They were paid \$0.65 to \$1.00 USD per submitted question with its English translation. Annotators were required to read the guidelines and complete a short comprehension task, for which they were paid \$7 USD. Additionally, answers to all Kirundi questions were paid \$2 USD per answer. This protocol was reviewed by the Institutional Review Board.

Annotators were instructed to write up to 25 questions in their native language along with English translations, ensuring the questions met the following criteria:

\begin{itemize}
\item The question requires a long-form answer (at least 3-4 sentences).
\item The question is culturally specific, meaning it is more likely to be asked in the region where the language is spoken.
\item The question is something a native speaker of the language might ask.
\item The question has an objective answer (i.e., not based on opinions).
\end{itemize}

\begin{table*}
\centering
{\renewcommand{\arraystretch}{1.5}
\resizebox{1\textwidth}{!}{%
\begin{tabular}{lp{0.9\textwidth}}
\toprule
\textsc{Field} & \textsc{Content} \\
\midrule
Language & English \\
Question (Original) & Why does the President of the United States need to be born in the United States to be eligible to run? It seems like the country that a person was born in has little to do with their abilities to lead.  \\
Question (English) & Why does the President of the United States need to be born in the United States to be eligible to run? It seems like the country that a person was born in has little to do with their abilities to lead.  \\
Culturally Specific & True \\
Answer & It's not a matter of their leadership ability, it's intended to guarantee loyalty. The idea is that most people are loyal to their birth country, so you'd want someone born in the US to be the one acting as commander in chief of the US armed forces, among other things. May not be perfect logic, but it's tradition at this point. Nobody has made a big enough deal of it accompanied by a strong enough argument to get it changed. \\
\midrule
Language & Hungarian \\
Question (Original) & Hogyan lehet kikeverni a Horthy-kori sisakok színét? \\
Question (English) & How can you mix the color of helmets from the Horthy era? (translated by \gptturbo{}) \\
Culturally Specific & True \\
Answer & M35 tábori sisak. Neten rákeresve találsz róla képeket. Nem kell megijedni a sok árnyalattól, annak idején sem volt tökéletesen egységes.

Ez egy zöldesbarna szín. Talán a RAL6025 áll a legközelebb hozzá. Festékboltban kikeverik géppel. Nem drágább, mint külön megvenni hozzá egy egy dobozzal a festéket és kevergetni.

Ecsettel festették az eredetit. \\
\midrule
Language & Fijian \\
Question (Original) & Na cava na vuna era vinakata kina na Nasi ni veiyanuyanu mera sa lesi i Viti Levu? \\
Question (English) & What is the reason the nurses from the outer islands want to be assigned to Viti Levu? \\
Culturally Specific & True \\
Answer & N/A \\
\midrule
Language & Kirundi \\
Question (Original) & Ni kubera iki twama dukeneye gushira ama aprikasiyo ku gihe? Hoba iki iyo tutabikoze? \\
Question (English) & Why do we need to constantly do software updates? What happens if I don't? \\
Culturally Specific & False \\
Answer & N/A \\
\bottomrule
\end{tabular}
}
\caption{Examples of entries in \name{}. Metadata like questions source (specific website or annotator) are omitted here for simplicity.}
\label{tab:examples}
}
\end{table*}

\begin{table*}[tbp]
\centering
{\renewcommand{\arraystretch}{1.5}
\resizebox{1\textwidth}{!}{%
\begin{tabular}{@{}llp{0.18\textwidth}lllp{0.18\textwidth}p{0.25\textwidth}l@{}}
\toprule
\addlinespace
\textsc{Language}   & \textsc{ISO} & \textsc{Family}          & Branch    & \textsc{Morphology} & \textsc{Order} & \textsc{Script}   & Region  & \textsc{Speakers} \\ 
\addlinespace
\midrule
\addlinespace
\multicolumn{9}{c}{High- \& Mid-Resource} \\
\addlinespace
\midrule
Arabic & ar & Afro-Asiatic & Semitic & fusional & SVO & Arabic alphabet & Arab world & 720M \\
Chinese & zh & Sino-Tibetan & Sinitic & analytic & SVO & Hanzi & Mainland China, Taiwan, Singapore & 1.38B \\
English & en & Indo-European & Germanic & analytic & SVO & Latin & World-wide & 1.5B \\
German & de & Indo-European & Germanic & fusional & SVO & Latin & Germany, Austria, Switzerland, etc. \raggedright & 133M \\
Hebrew & he & Afro-Asiatic & Semitic & fusional & SVO & Hebrew script & Israel & 9.3M \\
Hindi & hi & Indo-European & Indo-Iranian & fusional & SOV & Devanagari & India & 610M \\
Hungarian & hu & Uralic & Finno-Ugric & agglutinative & SVO & Latin & Hungary & 13M \\
Japanese & ja & Japonic & Japanese & agglutinative & SOV & Kanji, Kana & Japan & 123M \\
Korean & ko & Koreanic & Korean & agglutinative & SOV & Hangul & Korea & 82M \\
Russian & ru & Indo-European & Balto-Slavic & fusional & SVO & Cyrillic & Russia, Russian-speaking world \raggedright & 255M \\
Spanish & es & Indo-European & Italic & fusional & SVO & Latin & Spain, Central and South Americas, the US & 559M \\
\midrule
\addlinespace
\multicolumn{9}{c}{Low-Resource}  \\
\addlinespace
\midrule
Afar & aa & Afro-Asiatic & Cushitic & agglutinative & SOV & Latin & Ethiopia, Djibouti, Eritrea \raggedright & 2.6M \\
Balochi & bal & Indo-European & Indo-Iranian & agglutinative & SOV & Balochi Standard Alphabet & Pakistan, Iran, Afghanistan \raggedright & 8.8M \\
Faroese & fo & Indo-European & Germanic & fusional & SVO & Latin & Faroe Islands, Denmark & 69K \\
Fijian & fj & Austronesian & Malayo-Polynesian & agglutinative & VOS & Latin & Fiji & 640K \\
Hiligaynon & hil & Austronesian & Malayo-Polynesian & analytic & VSO & Latin & Philippines & 9.1M \\
Kirundi & rn & Niger-Kongo & Atlantic–Congo & agglutinative & SVO & Latin & Burundi & 12-13M \\
Papiamento & pap & Portuguese-based creole & Afro-Portuguese & analytic & SVO & Latin & Aruba, Curaçao, Bonaire & 300K \\
Pashto & ps & Indo-European & Indo-Iranian & fusional & SOV & Pashto alphabet & Afghanistan, Pakistan and Iran \raggedright & 58.8M \\
Samoan & sm & Austronesian & Malayo-Polynesian & analytic & VSO & Latin & Samoa & 510K \\
Tongan & to & Austronesian & Polynesian & agglutinative & VSO & Latin & Tonga & 187K \\
Tswana & tn & Niger-Kongo & Atlantic–Congo & agglutinative & SVO & Latin & Botswana, South Africa, Zimbabwe & 13.9M \\
Wolof & wo & Niger-Kongo & Atlantic–Congo & agglutinative & SVO & Latin primarily & Senegal & 12.3M \\
\bottomrule
\end{tabular}%
}
}
\caption{Linguistic and usage information of the languages in the \name{} dataset}
\label{app_tab:language_data}
\end{table*}


\section{Question Categorization} \label{sec:categorization}

In this section we describe the process of categorizing all questions into predefined set of categories.

\paragraph{Method} We selected 25 random culturally specific questions from the dataset. We manually created a list of broad categories with descriptions and examples, and then 2 authors independently applied the categorization on the 25 questions. We reviewed disagreements and accordingly refined the categories. Then we used \gpto{} to categorize using the prompts in \autoref{tab:category_prompts}, with temperature set to $0.0$. After minor clarifications to category descriptions, we found that \gpto{} produced adequate categories for all 25 questions. We consequently used the model to categorize all of \name. Our final categories, with descriptions and examples, can be found in \autoref{tab:categories_of_questions}.

\begin{figure*}[tbp]
\centering
\includegraphics[width=1\textwidth]{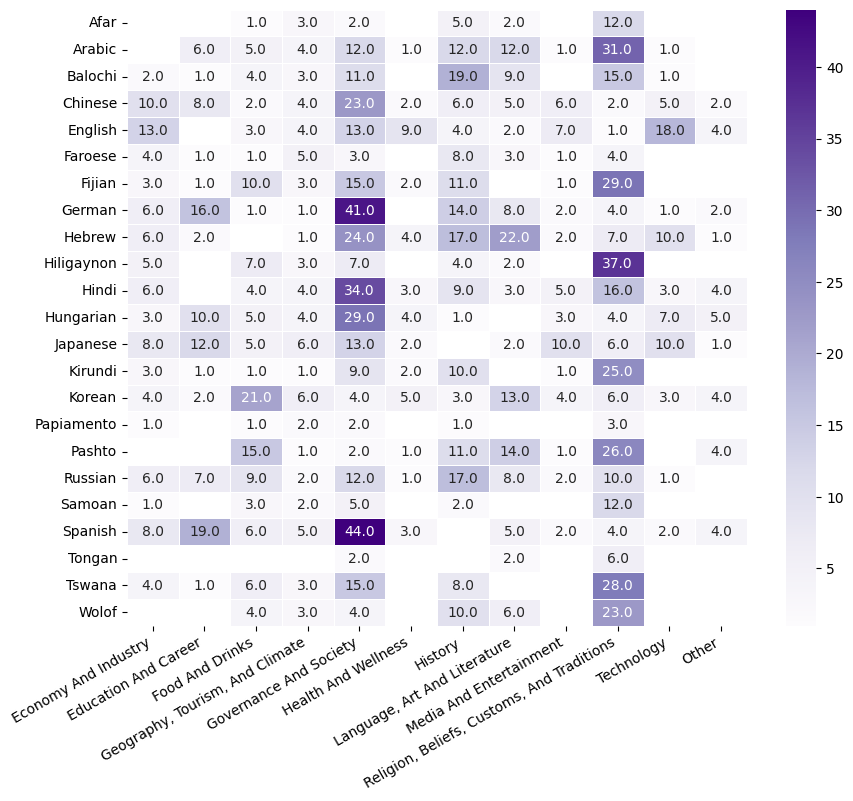} 
\caption{Number of human-collected questions by category and language.}
\label{fig:questions_by_categories_and_lang}
\end{figure*}

\paragraph{Results} Figure \ref{fig:questions_by_categories_and_lang} shows the number of questions by category and language. We observe that one of \textsc{Religion, Beliefs, Customs, and Traditions}, \textsc{Governance and Society} and \textsc{History} is the top category of almost every language (the exceptions being English and Korean). Furthermore, \textsc{Religion, Beliefs, Customs, and Traditions} is the predominantly the top category for low-resource languages (10/12). This difference is likely due to the question collection process for low-resource languages.

\begin{figure*}[tbp]
\centering
\includegraphics[width=1\textwidth]{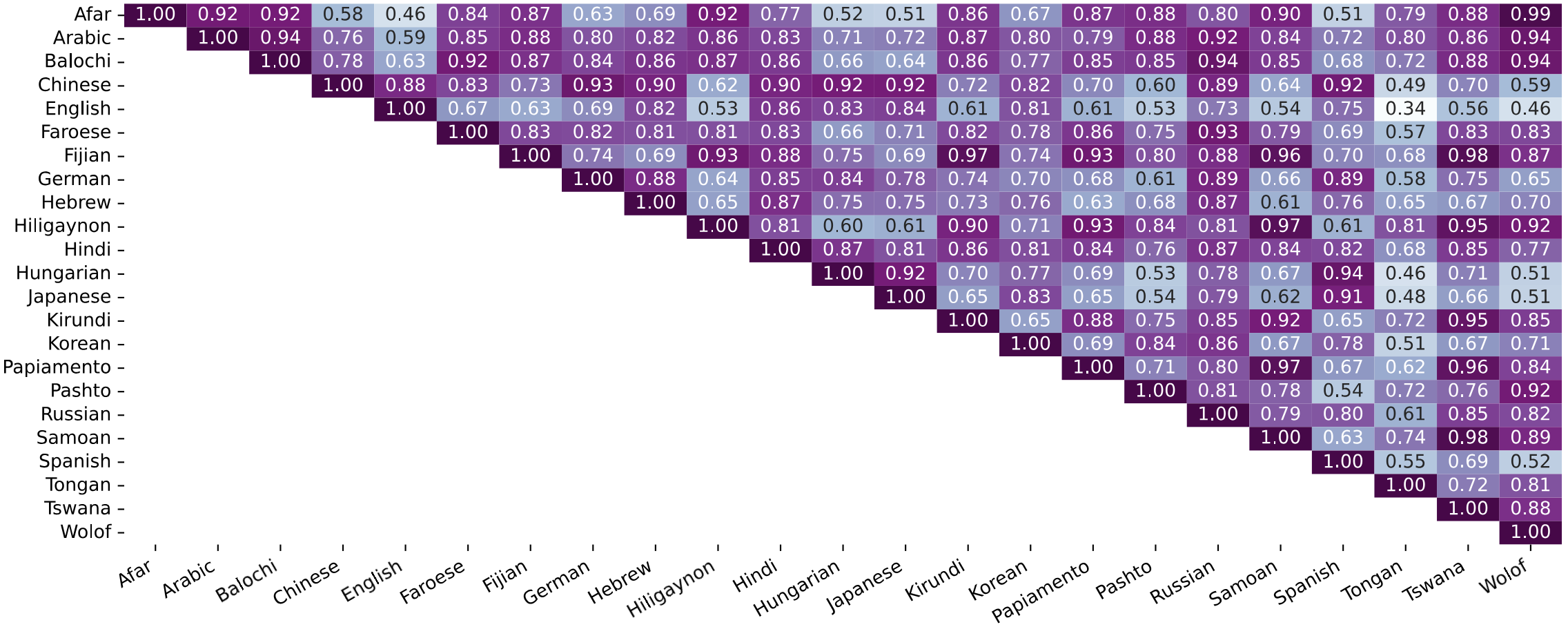}
\caption{Bhattacharyya coefficients of the category distributions, pairwise between languages. The Bhattacharyya coefficient ranges from 0 to 1, with a higher number meaning more similar distributions.}
\label{fig:categories_bhattacharyya}
\end{figure*}

To compare the distribution of categories between languages, we compute pairwise Bhattacharyya coefficients between the data from the languages (Figure~\ref{fig:categories_bhattacharyya}). The Bhattacharyya coefficient ranges from 0 to 1 with a higher number meaning similar distributions. We see generally high coefficients, indicating that the category distributions are similar between languages.

\begin{table*}
\centering
{\renewcommand{\arraystretch}{1.5}
\resizebox{1\textwidth}{!}{%
\begin{tabular}{lp{0.9\textwidth}}
\toprule
\textsc{Language} & \textsc{Prompt} \\
\midrule
English
&
You are categorizing questions about different cultures into specific categories. Your task is to assign one category to each question. Here are the available categories:
\vspace{\baselineskip}

education and career – \textit{[Education and Career Description]}. Example: \textit{[Education and Career Example]}

\textit{[categories]}\dots

health and wellness – \textit{[Health and Wellness Description]}. Example: \textit{[Health and Wellness Example]}

other – \textit{[Other Description]}
\vspace{\baselineskip}

Here is the question to categorize:

<original\_question>\textit{[question]}</original\_question>
\vspace{\baselineskip}

Categorize this question into one of the categories. Output your choice in the following format:
<category>category name</category>
\vspace{\baselineskip}

Your choice: 
\\
All except English
& You are categorizing questions about different cultures into specific categories. Your task is to assign one category to each question. Here are the available categories:
\vspace{\baselineskip}

education and career – \textit{[Education and Career Description]}. Example: \textit{[Education and Career Example]}

\textit{[categories]}\dots

health and wellness – \textit{[Health and Wellness Description]}. Example: \textit{[Health and Wellness Example]}

other – \textit{[Other Description]}
\vspace{\baselineskip}

Here is the question to categorize:

<original\_question>\textit{[question]}</original\_question>

<translation>\textit{[translation]<}/translation>
\vspace{\baselineskip}

Categorize this question into one of the categories. Output your choice in the following format:
<category>category name</category>
\vspace{\baselineskip}

Your choice: \\
\bottomrule
\end{tabular}
}
\caption{Prompts used with \gpto{} to categorize questions. Strings in the form \textit{[form]} are placeholders that are replaced at runtime. The categories used are in \autoref{tab:categories_of_questions}.}
\label{tab:category_prompts}
}
\end{table*}

\begin{table*}
\centering
{\renewcommand{\arraystretch}{1.5}
\resizebox{1\textwidth}{!}{%
\begin{tabular}{p{0.22\textwidth}p{0.5\textwidth}p{0.3\textwidth}}
\toprule
\textsc{Category} & \textsc{Description} & \textsc{Example} \\
\midrule
\textsc{Education and Career} \raggedright
    & Questions related to school, education system, jobs and career paths. Includes developing new skills for new jobs.
        & \textit{Why do young children drop out of school?} \\
\textsc{Governance and Society} \raggedright
    & Questions about laws, governance and policies, as well as politics and social issues.
        & \textit{What are the reasons why Japan cannot have casinos?} \\
\textsc{Geography, Tourism, and Climate} \raggedright
    & Questions concerning the geography, climatic conditions, environmental factors of a region, tourism and travelling.
        & \textit{What is the significance of Gorée Island?} \\
\textsc{Technology} \raggedright
    & Questions about the technology, technological advancements, uses of technology and digital innovation.
        & \textit{Are stores that accept VISA debit cards marked VISA? Or is it a store with a VISA PLUS mark?} \\
\textsc{Economy and Industry} \raggedright
    & Questions regarding modern-day economic practices, key industries, trade, and economic development.
        & \textit{Why is our country not developing like others?} \\
\textsc{Media and Entertainment} \raggedright
    & Questions about the media and entertainment specific to the region.
        & \textit{Why are trademarks obscured on broadcast?} \\
\textsc{Food and Drinks} \raggedright
    & Questions related to culinary traditions, typical foods and beverages, preparation methods, culinary practices, and cultural significance of meals.
        & \textit{Why is bread with sausage called a sandwich in Russia?} \\
\textsc{History} \raggedright
    & Questions about historical events, significant figures, and important periods that have shaped a culture.
        & \textit{How did the Bujumbura market burn?} \\
\textsc{Language, Art and Literature} \raggedright
    & Questions about the language, dialects, as well as art forms and literary works/traditions.
        & \textit{Why is the Balochi Language categorized into its three main dialects?} \\
\textsc{Religion, Beliefs, Customs, and Traditions} \raggedright
    & Questions regarding religious practices, beliefs, rituals, customs, traditions, and holiday.
        & \textit{In the Islamic religious teaching, what’s the meaning of the seven tens?} \\
\textsc{Health and Wellness} \raggedright
    & Questions related to traditional and modern health practices, public health issues, and well-being.
        & \textit{Why methadone? What makes it "better" than other opioids for maintenance therapy or tapering off another drug?} \\
\textsc{Other}
    & Questions that do not fit neatly into the above categories.
        & \\
\bottomrule
\end{tabular}
}
\caption{Categories of questions in \name{}.}
\label{tab:categories_of_questions}
}
\end{table*}

\section{Automatic Evaluation}
\label{app:auto_eval}

In this section of the appendix we present the details of automatic evaluation.  All evaluated models are listed in \autoref{tab:evaluated_models}. Examples of the model tendencies detected by automatic evaluation are in \autoref{tab:faulty_generations_examples}.

\subsection{Method Details}
\label{app:auto_eval_method}

\begin{figure*}[tbp]
  \adjustbox{trim=1.2cm 1cm 2cm 1cm}{%
    \includeinkscape[width=1.2\linewidth]{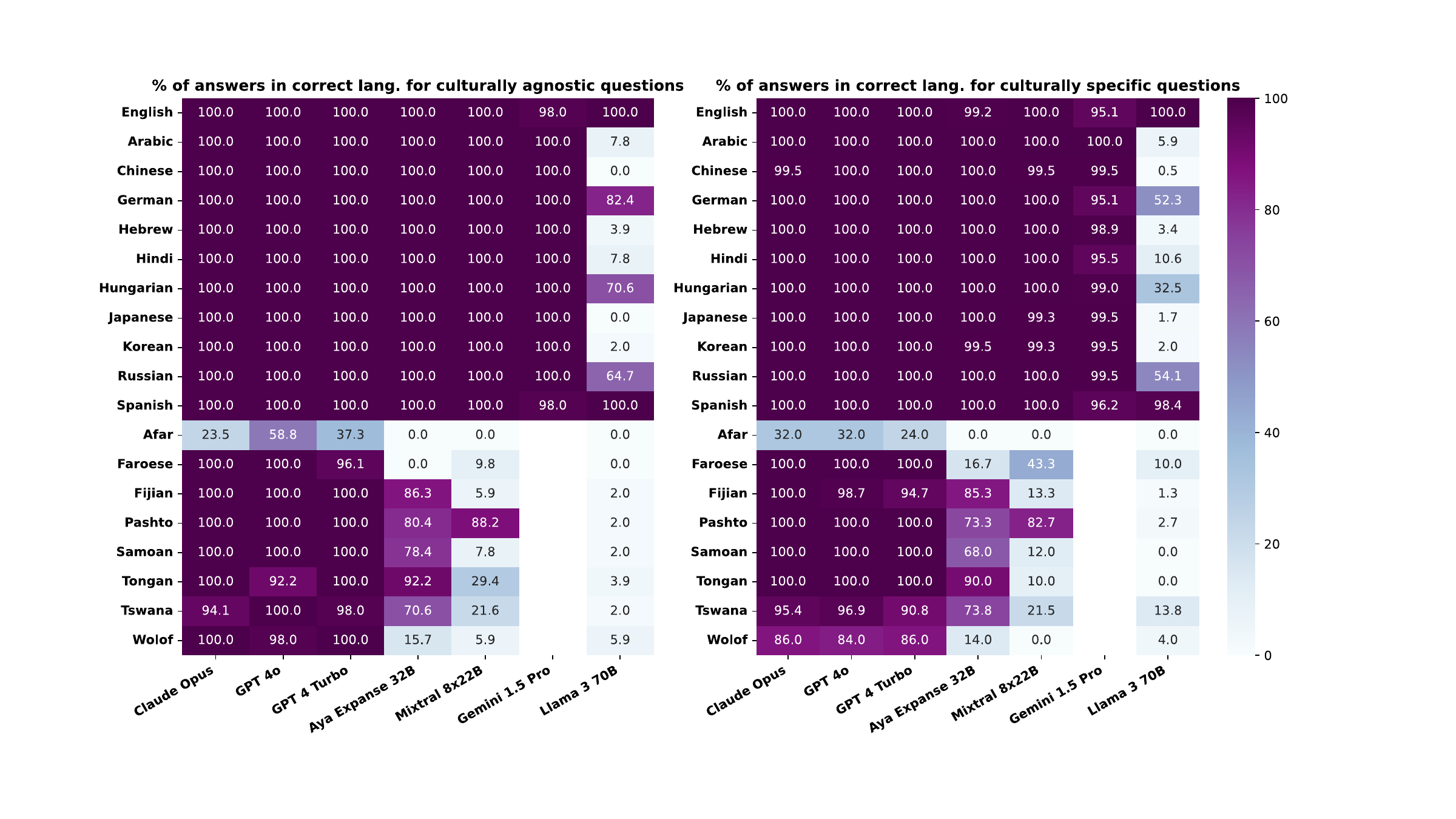}%
  }
 \caption{Percentage of responses generated in the correct language by model for culture specific and culturally agnostic questions. Blank cells for \gemini{} indicate languages where the API returned an error message. Balochi, Hiligaynon, and Papiamento are omitted since language detection libraries performed poorly for these languages. Additionally, detection accuracy for Kirundi was compromised, with instances of Kirundi being incorrectly identified as Kinyarwanda. Please see \autoref{app:acc_lang_detect_on_test_set} for details.}
  \label{appfig:heatmap_lang_accuracy}
\end{figure*}

\paragraph{Language accuracy} \autoref{appfig:heatmap_lang_accuracy} displays the percentage of responses each model generated in the correct language, independent of correctness or fluency of the answer. We used \texttt{polyglot} (\mbox{\url{https://pypi.org/project/polyglot/}}) and \texttt{langid} (\url{https://pypi.org/project/py3langid/}) for language identification, choosing them based on their performance for specific languages. This identification was also applied to the questions to estimate its performance across languages. Our pipeline accurately recognized 100\% of instances in 14 languages. For other languages, accuracy typically remained above 90\%, with Fijian at 98.67\%, Russian at 97.33\%, Tongan at 96.92\%, Samoan at 92.00\%, and Wolof at 90\% (see \autoref{app:acc_lang_detect_on_test_set}). However, identification accuracy for Kirundi was notably lower at 35.85\%, as the libraries frequently misclassified it as the closely related Kinyarwanda. The automatic identification process failed entirely for Balochi, Hiligaynon, and Papiamento, which is reflected in seemingly low performance for these languages across all the models.

\begin{table*}[t]
    \centering
    \begin{tabular}{llr}
        \toprule
        \textsc{Language} & \textsc{Lang ID} & \textsc{Accuracy (\%)} \\
        \midrule
        Afar          & aa  & 100.00 \\
        Arabic        & ar  & 100.00 \\
        Balochi       & bal & {\color{red}0.00}  \\
        Chinese       & zh  & 100.00 \\
        English       & en  & 100.00 \\
        Faroese       & fo  & 100.00 \\
        Fijian        & fj  & 98.67 \\
        German        & de  & 100.00 \\
        Hebrew        & he  & 100.00 \\
        Hiligaynon    & hil & {\color{red}0.00}  \\
        Hindi         & hi  & 100.00 \\
        Hungarian     & hu  & 100.00 \\
        Japanese      & ja  & 100.00 \\
        Kirundi       & rn  & {\color{red}35.85} \\
        Korean        & ko  & 100.00 \\
        Papiamento    & pap & {\color{red}0.00}  \\
        Pashto        & ps  & 100.00 \\
        Russian       & ru  & 97.33 \\
        Samoan        & sm  & 92.00 \\
        Spanish       & es  & 100.00 \\
        Tongan        & to  & 100.00 \\
        Tswana        & tn  & 96.92 \\
        Wolof         & wo  & 90.00 \\
        \bottomrule
    \end{tabular}
    \caption{Accuracy of the language detection pipeline on the test set made from questions in the given language. Note that the language detection libraries are often more accurate on longer texts (i.e., texts longer than the length of a single question).}
    \label{app:acc_lang_detect_on_test_set}
\end{table*}

\begin{figure*}[tbp]
  \adjustbox{trim=1.2cm 1cm 2cm 1cm}{%
    \includeinkscape[width=1.2\linewidth]{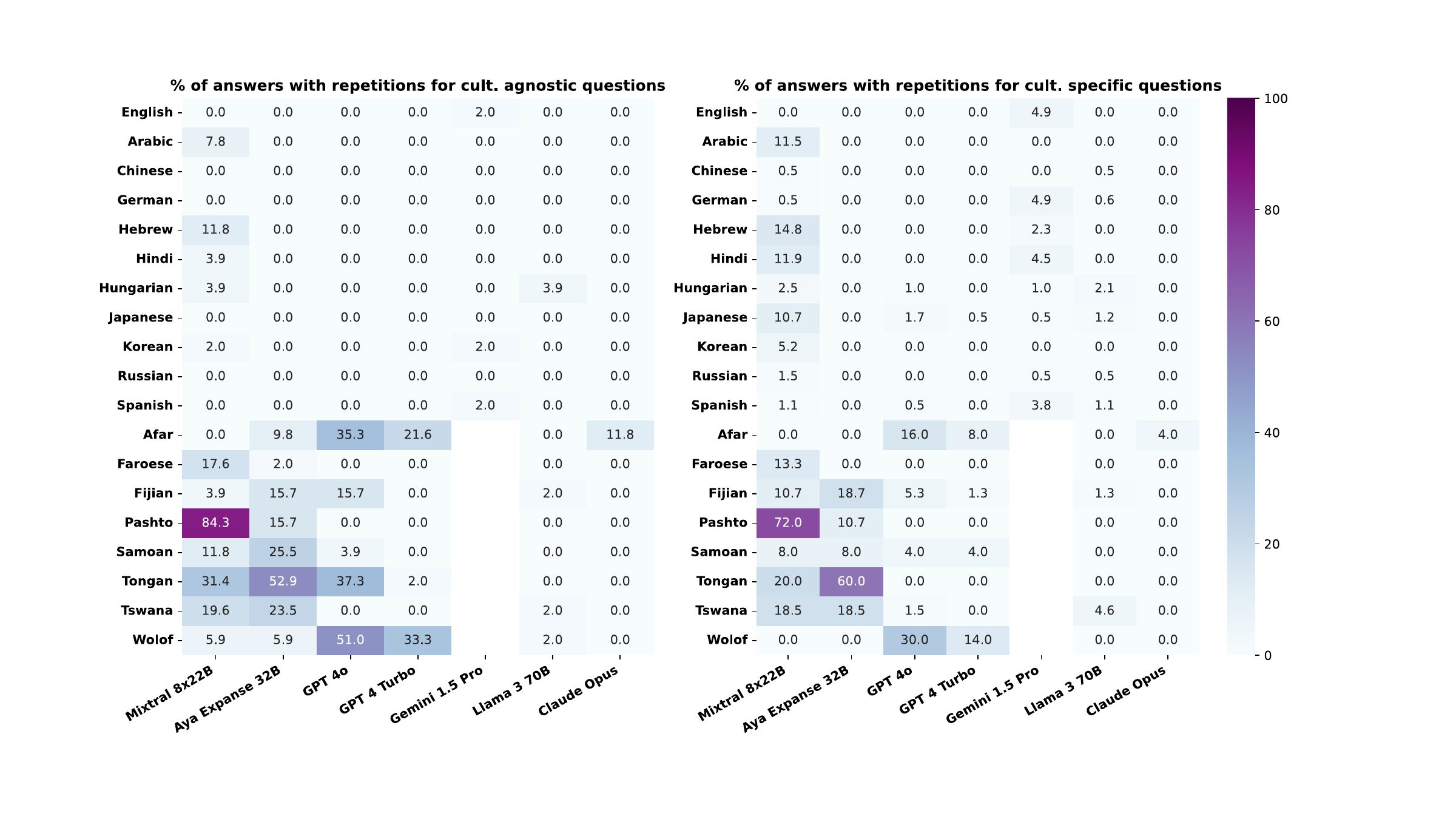}%
  }
  \caption{Percentage of answers with identified repetitions for culturally specific and culturally agnostic questions. Blank cells for \gemini{} indicate languages where the API returned an error message.}
  \label{appfig:heatmap_repetitions}
\end{figure*}

\paragraph{Repetitions} \autoref{appfig:heatmap_repetitions} illustrates the percentage of responses affected by repetitions, analyzed by language across different models. To identify these repetitions, we employed \texttt{tiktoken} (\url{https://github.com/openai/tiktoken}) with the \texttt{o200\_base} encoding. We specifically identified instances where at least 20 consecutive tokens were repeated at least four times within an answer.

\begin{figure*}[tbp]
  \includegraphics[width=\linewidth]{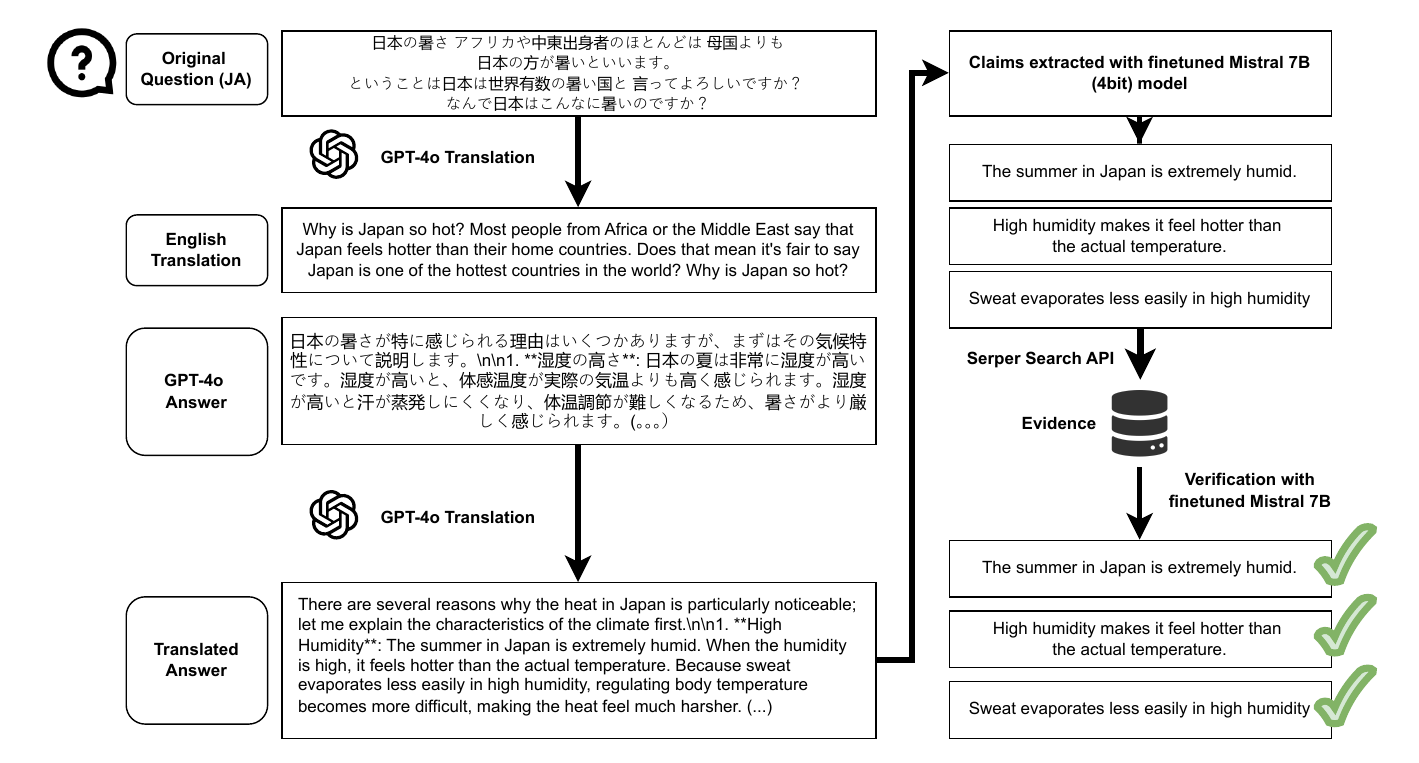}
  \caption{Claim extraction and verification pipeline. Example showing extraction and verification of claims for a question and answer in Japanese. English translations were obtained with \gpto{}. Only part of the answer is provided for readability.}
  \label{fig:claim_extract_pipe}
\end{figure*}

\paragraph{Claim extraction and verification pipeline} We first translated the answers into English with \gpto{}. Then we extract claims using a finetuned Mistral 7B model and use them to query Serper API for evidence. Then we prompt a finetuned Mistral 7B model for verification. Both models were introduced in \citet{song2024veriscore}. The pipeline is visualized in \autoref{fig:claim_extract_pipe}.

\begin{figure*}[tbp]
  \adjustbox{trim=1.2cm 1cm 2cm 1cm}{%
    \includeinkscape[width=1.2\linewidth]{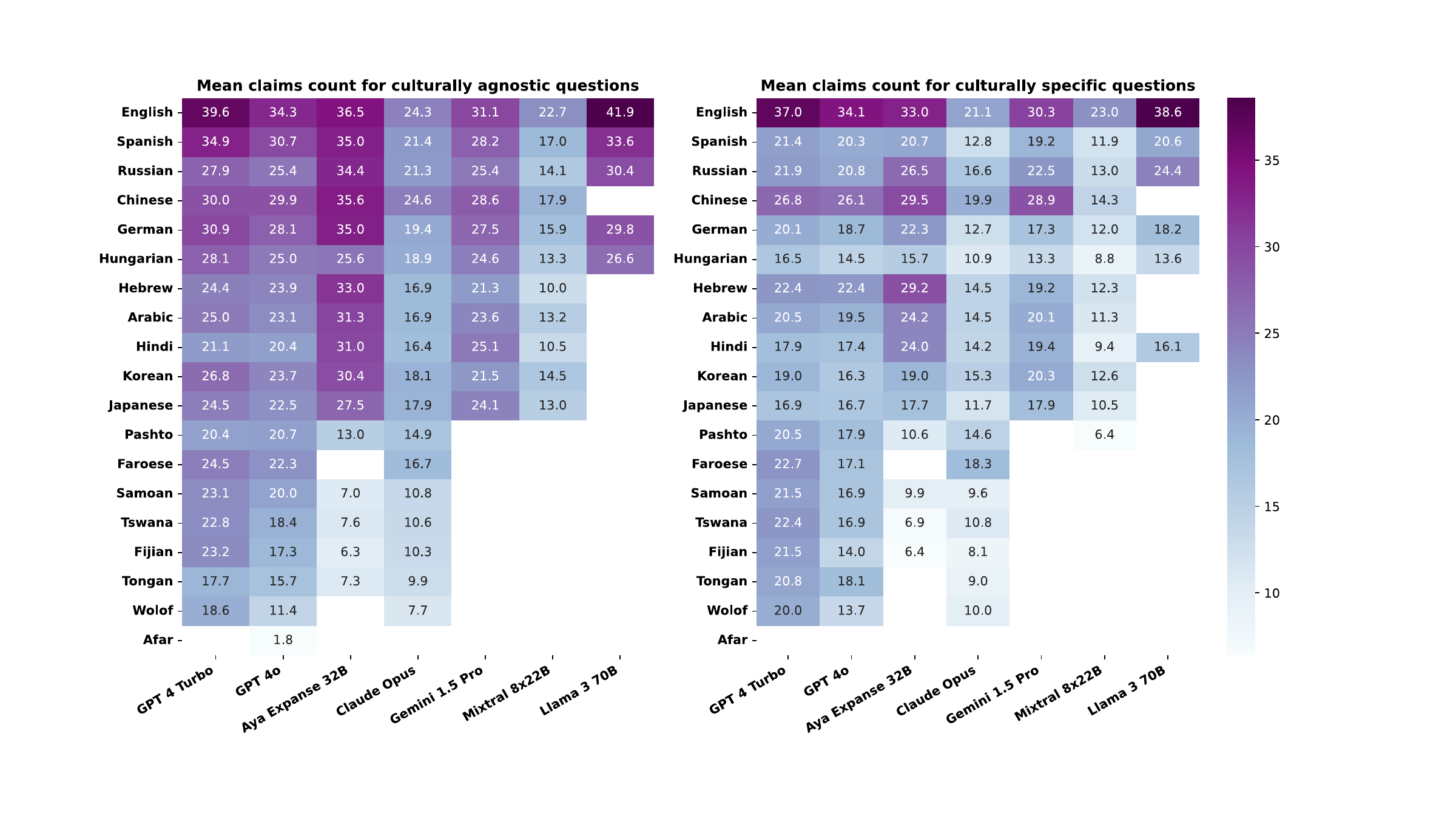}%
  }
  \caption{Mean claim count for answers without surface-level issues. The left heatmap shows the results for culturally agnostic questions while the right heatmap shows the results for culturally specific questions.  Only languages where at least 10 answers were free from surface-level issues are included.}
  \label{fig:mean_claim_count}
\end{figure*}

We report the mean claim count by model, language of the question, and question type in \autoref{fig:mean_claim_count}. We exclude all answers with surface-level issues as well as languages for which the model produced less than 50\% of valid answers (i.e., answers without identified surface level issues).

\subsection{Further Analysis}
\label{app:auto_eval_analysis}

\paragraph{Model surface-level errors}
We further analyzed the responses for specific textual indicators. Detected patterns in model responses are presented with examples in \autoref{tab:faulty_generations_examples}.

\begin{figure*}[tbp]
  \adjustbox{trim=1.2cm 1cm 2cm 1cm}{%
    \includeinkscape[width=1.2\linewidth]{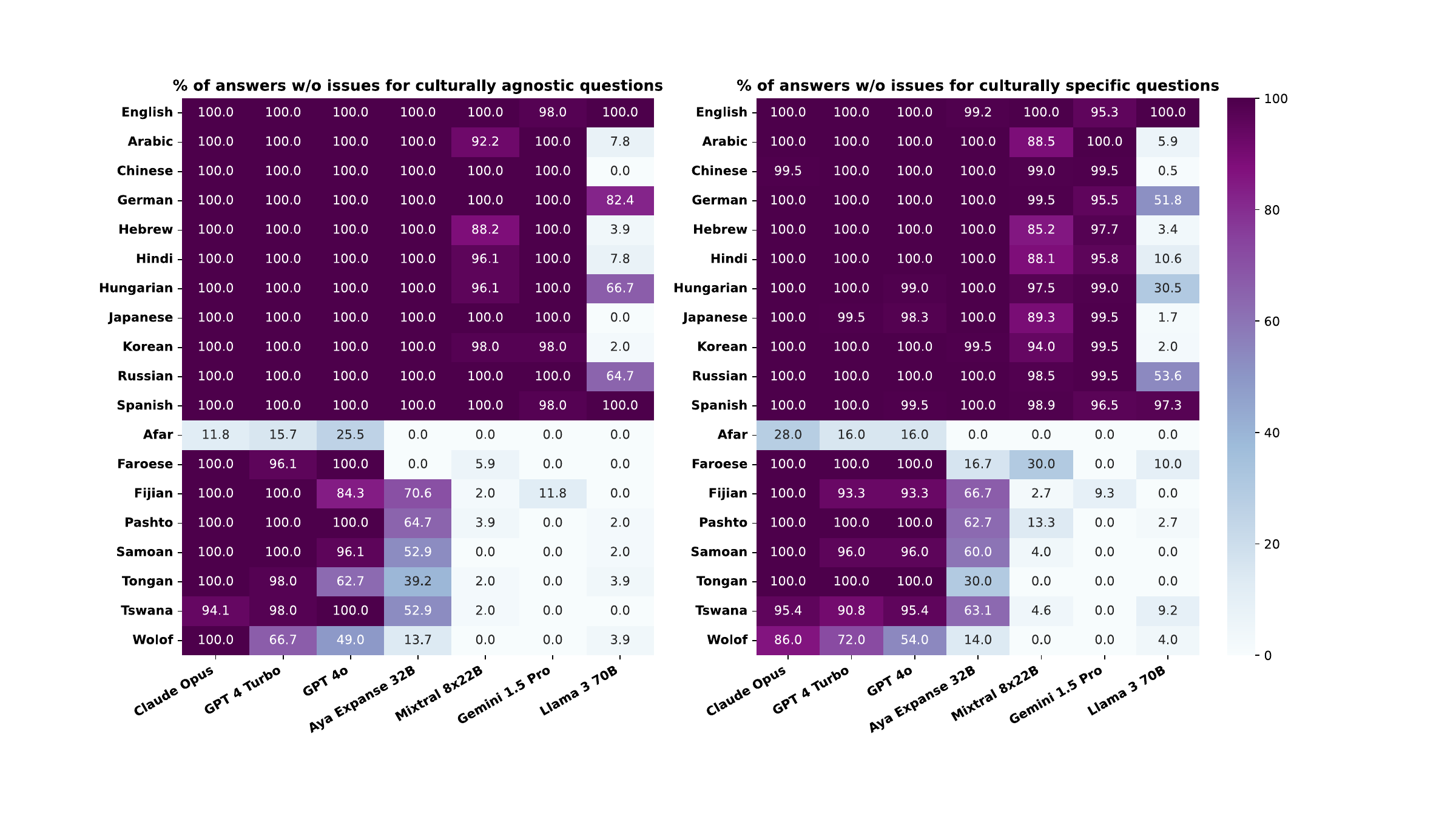}%
  }
  \vspace{-2em}
  \caption{Percentage of model answers without surface issues per language. The left heatmap shows the results for culturally agnostic questions while the right heatmap shows the results for culturally specific questions.}
  \label{fig:automatic_by_lang}
\end{figure*}

\begin{figure*}[tbp]
\centering
  \adjustbox{trim=1.2cm 1cm 2cm 1cm}{%
    \includeinkscape[width=1.2\linewidth]{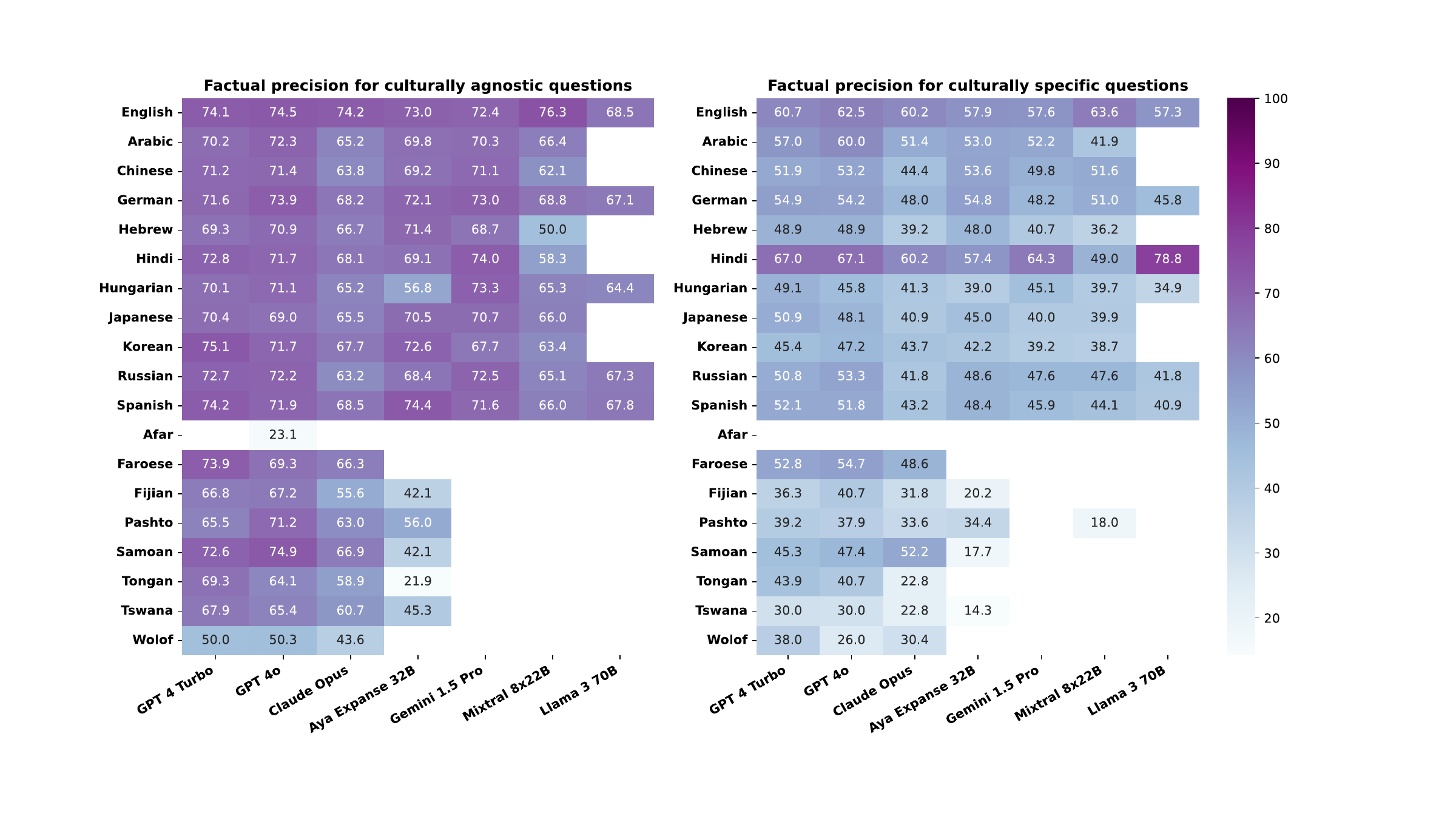}%
  }
  \caption{Factual precision for answers without surface-level issues. The left heatmap shows the results for culturally agnostic questions while the right heatmap shows the results for culturally specific questions. We remove model-language combinations for which there are not at least 10 answers without surface-level issues. Factual precision degrades on culturally specific questions, especially for low-resource languages.}
  \vspace{-0.4em}
  \label{fig:factuality_by_lang}
\end{figure*}

\begin{figure*}[tbp]
\centering
  \adjustbox{trim=1.2cm 1cm 2cm 1cm}{%
    \includeinkscape[width=1.2\linewidth]{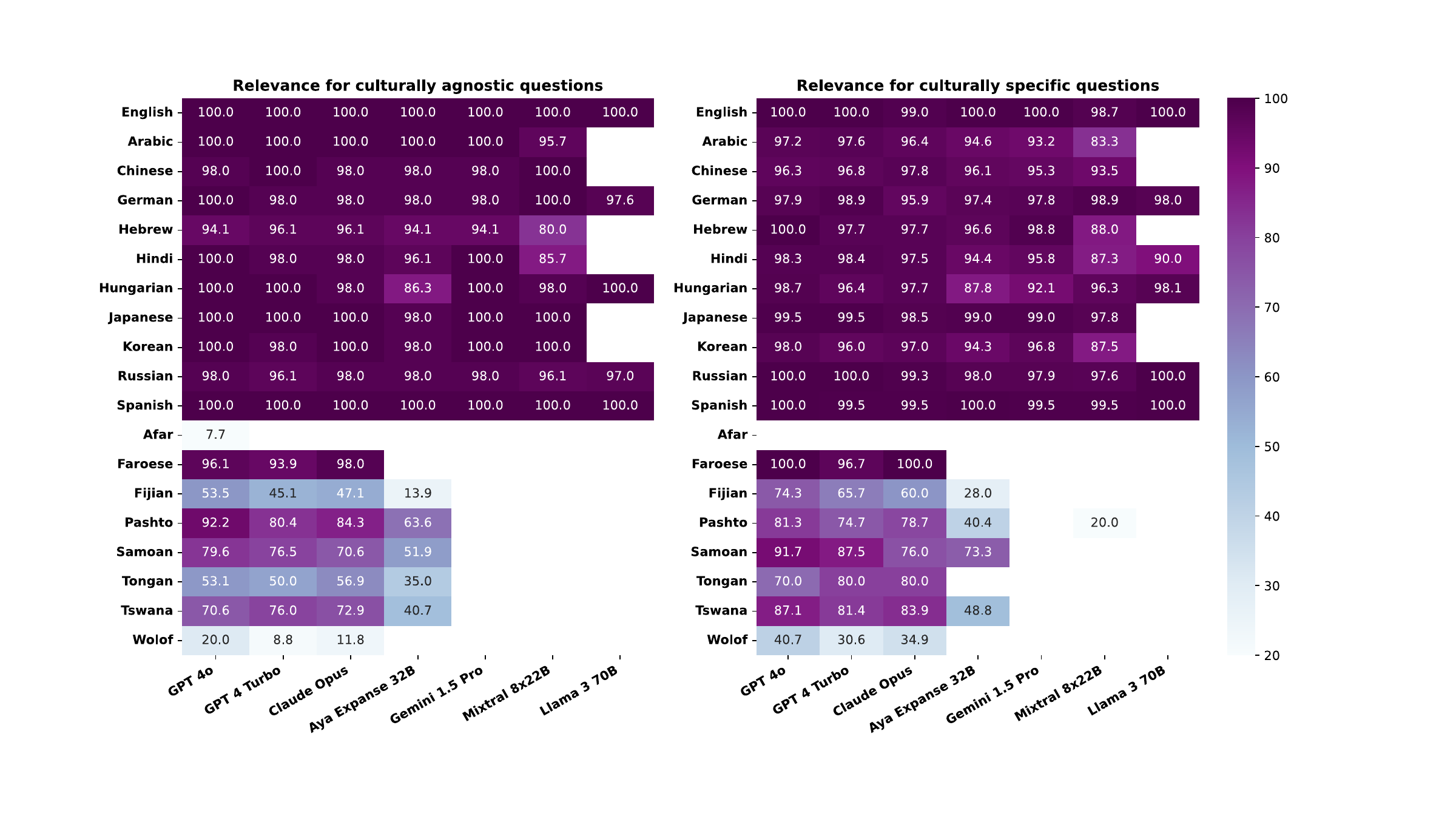}%
  }
  \caption{Relevance for answers without surface-level issues. The left heatmap shows the results for culturally agnostic questions while the right heatmap shows the results for culturally specific questions. We remove model-language combinations for which there are not at least 10 answers without surface-level issues. Answer relevance degrades for low-resource languages but is similar on culturally specific and culturally agnostic questions.}
  \vspace{-0.4em}
  \label{fig:relevance_by_lang}
\end{figure*}

Our textual analysis demonstrates issues in \mixtral{} responses for low-resource languages. 31.47\% of \mixtral{} responses to questions in low-resource languages contain phrases like ``sorry'', ``apologize'' or ``understand'' (e.g., \textit{"I'm sorry for any confusion, but it seems you're using a language that I'm not currently able to understand or translate."}). \mixtral{} responses to questions in high-resource languages do not contain these apology-related keywords, revealing an inability to answer the question specifically in low-resource languages. The apologetic textual markers were seen in less than 1\% of other models' responses except for \llama{}'s, where they were present in 14.74\% of low-resource and 10.48\% of high-resource language answers.

Textual indicators also uncover deficiencies in \llama{} responses. Notably, 37.87\% of responses from \llama{} explicitly mention the English name of the language (e.g., \textit{"I see you're speaking in Balochi!"}), indicating that although the system recognizes the language of the question, it nonetheless responds in English. This is in contrast to \mixtral{}, which does so in 7.21\% of responses, \gptturbo{} at 1.84\%, and less than 1\% for other models. Additionally, approximately 19.71\% of \llama{} responses include terms like ``translate'' or ``translation'' (e.g., \textit{"I apologize, but I'm having trouble understanding your question. Could you please rephrase or translate your question into a language I can understand, such as English?"}), where the system either declines to answer (with or without apology), requests an English translation, or provides a translation itself.
In comparison, 8.43\% of \mixtral{} responses exhibit similar behavior, with less than 1\% for other models. Lastly, we observed an unusually high proportion of emojis in responses generated by \llama{}, with 17.54\% containing at least one emoji.

\paragraph{Human- vs automatically-collected questions}

\autoref{app_tab:overall_by_scraped} shows model performance scores on human collected and automatically collected questions. We see comparable results between the two question sets, though some model rankings change. Specifically, \gptturbo{} and \gemini{} move ahead of \gpto{} and \claude{} respectively in overall performance. Nevertheless, we see important trends like poor factuality scores and high \mixtral{} repetitions on both sets of questions.

\begin{figure*}[tbp]
  \adjustbox{trim=1.2cm 1cm 2cm 1cm}{%
    \includeinkscape[width=1.2\linewidth]{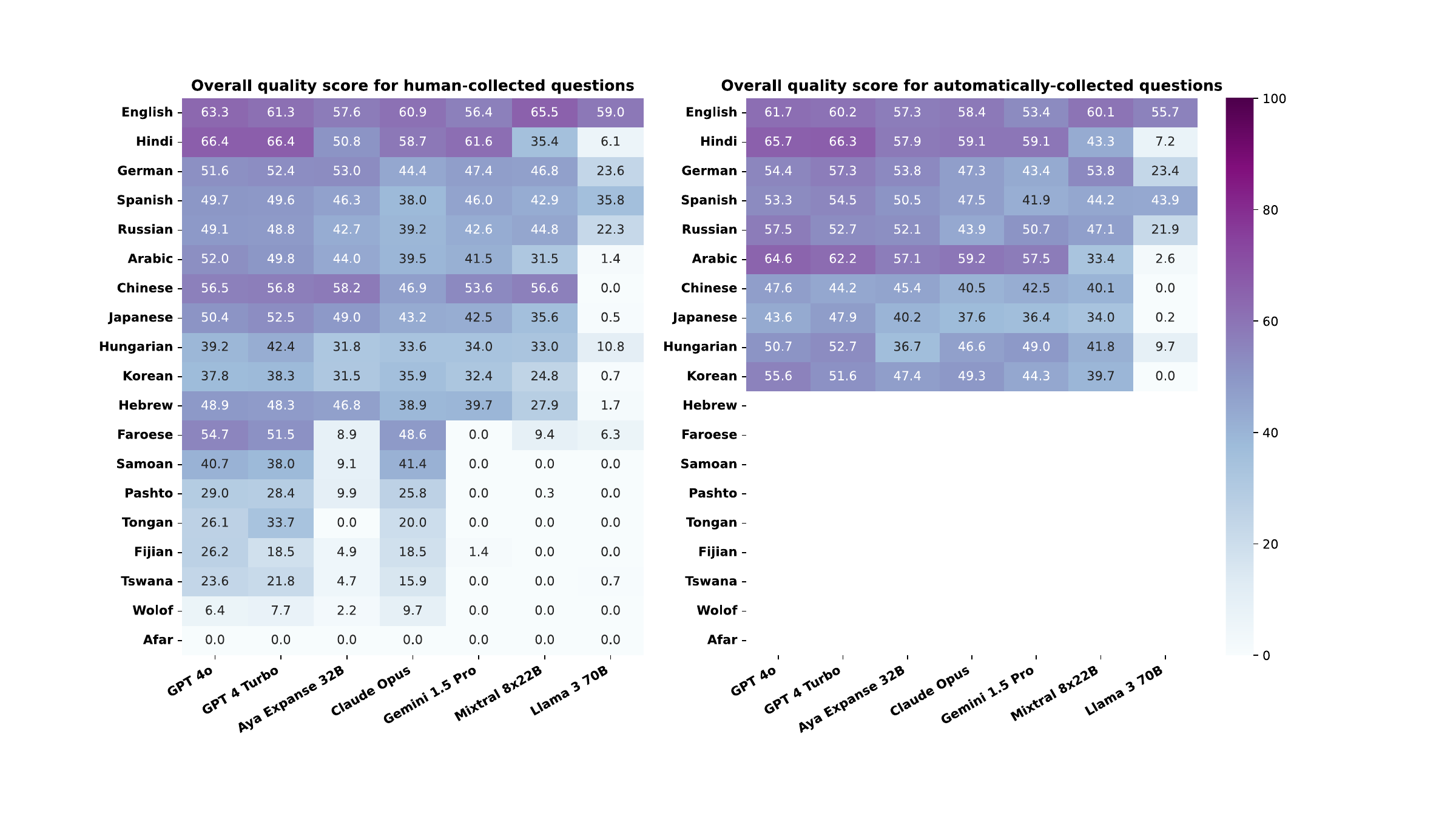}%
  }
 \caption{Answer scores $S$ based on our quality criteria: surface issues, factuality and relevance. The left heatmap shows the results for human collected questions while the right heatmap shows the results for a subset of automatically collected questions (100 per language). Model performance differs notably between human collected and automatically collected questions in non-English languages.}
  \label{app_fig:heatmap_overall_by_scraped}
\end{figure*}

\autoref{app_fig:heatmap_overall_by_scraped} breaks down model overall performance on the human collected and automatically collected questions by language. We observe that model performance is higher on the automatically collected question for most languages. To determine whether the performance difference between the two question sets is significant, for each language we conduct a 2-sample Kolmogorov–Smirnov test on the overall answer scores for those languages, with the null hypothesis that the answer scores are drawn from the same distribution. The tests refute the null hypothesis with p-values below $0.01$ for all languages except English and Hindi, for which p-values are $0.22$ and $0.08$ respectively. Although model performance is not identical on both question sets, analysis like \autoref{app_tab:overall_by_scraped} shows that model performance on automatically collected questions is an effective proxy for performance on human collected questions.

\begin{table*}[t]
\footnotesize
\centering
\begin{tabularx}{\textwidth}{@{}l*{7}{>{\centering\arraybackslash}X}@{}}
\toprule
\multirow{2}{*}{\textsc{}} & \multicolumn{3}{c}{\textbf{Surface Level}} & \multicolumn{2}{c}{\textbf{Fine-Grained}} & \textbf{Overall} \\ 
\cmidrule(lr){2-4} \cmidrule(lr){5-6} \cmidrule(lr){7-7}
& Wrong Lang. (\%) $\downarrow$ & Repetitions (\%) $\downarrow$ & W/o Issues (\%) $\uparrow$ 
& Factual Pr. (\%) $\uparrow$ & Relevance (\%) $\uparrow$ & Overall Score $S$ $\uparrow$ \\
\midrule
\gptturbo{}     & \textbf{0.0} / \textbf{0.0} & \textbf{0.0} / 0.1 & \textbf{100.0} / \textbf{99.9} & \textbf{52.3} / 55.8 & 98.4 / 98.3 & \textbf{52.0} / 55.0 \\
\gpto{}         & \textbf{0.0} / \textbf{0.0} & 0.1 / 0.5 & 99.9 / 99.5 & \textbf{52.3} / \textbf{56.5} & \textbf{98.7} / \textbf{98.5} & 51.8 / \textbf{55.5} \\
\ayanew{}       & 0.1 / 0.1 & \textbf{0.0} / \textbf{0.0} & 99.9 / \textbf{99.9} & 48.3 / 51.2 & 96.6 / 95.8 & 46.8 / 49.8 \\
\gemini{}       & 0.8 / 3.1 & 0.8 / 2.9 & 99.2 / 96.9 & 47.7 / 48.6 & 94.8 / 95.2 & 46.1 / 47.8 \\
\claude{}       & \textbf{0.0} / \textbf{0.0} & \textbf{0.0} / \textbf{0.0} & \textbf{100.0} / \textbf{99.9} & 44.6 / 50.4 & 98.5 / 97.1 & 44.0 / 49.0 \\
\mixtral{}      & 0.3 / 0.1 & 4.4 / 4.7 & 95.4 / 95.2 & 45.5 / 47.7 & 93.8 / 94.3 & 41.6 / 43.8 \\
\llama{}        & 61.2 / 66.0 & 0.7 / 0.5  & 38.2 / 33.7 & 44.9 / 49.8 & 96.4 / 96.3 & 16.2 / 16.5 \\
\bottomrule
\end{tabularx}
\caption{Model performance aggregated across 10 high resource languages on human collected and a subset of automatically collected questions (100 per language). Each cell reports values on human collected / automatically collected culturally specific question.}
\vspace{-1em}
\label{app_tab:overall_by_scraped}
\end{table*}

\begin{table*}
\centering
\footnotesize
\resizebox{\textwidth}{!}{%
\begin{tabular}{l l l l  }
\toprule
\textsc{Model} & \textsc{Checkpoint} & \textsc{Avail.} & \textsc{\# Param}  \\
\midrule
\gptturbo{} & \texttt{gpt-4-0125-preview} &  \textbf{closed} & ?   \\
\gpto{} & \texttt{gpt-4o-2024-05-13} & \textbf{closed} & ?  \\
\claude{} & \texttt{claude-3-opus-20240229} & \textbf{closed} & ?   \\
\gemini{} & \texttt{gemini-1.5-pro-preview-0514} & \textbf{closed} & ?  \\
\midrule
\llama{} & \texttt{Meta-Llama-3-70B-Instruct}  x& \textbf{open-weight} & 70B  \\
\mixtral{} & \texttt{Mixtral-8x22B-Instruct-v0.1}  x & \textbf{open-weight} & 8x22B  \\
\ayanew{} & \texttt{aya-expanse-32b} x &\textbf{open-weight}  & 32B  \\
\bottomrule
\end{tabular}
}
\caption{General information about models we evaluate using \name.}
\label{tab:evaluated_models}
\end{table*}

\begin{table*}
\centering
\footnotesize
{\renewcommand{\arraystretch}{1.5}
\resizebox{\textwidth}{!}{%
\begin{tabular}{p{0.2\textwidth} p{0.24\textwidth} p{0.56\textwidth}}
\toprule
\textsc{Issue} & \textsc{Question} & \textsc{Generation} \\
\midrule
The model answers a non-English question in English
& A BKV járatain egy (külföldi) nyugdíjas miként igazolja, hogy nyugdíjas a sofőrnél? \dots

\textit{How can a (foreign) senior citizen prove their eligibility for free travel on Budapest's public transport (BKV) to the driver? \dots}
& I understand your concern! As a foreign pensioner, your mother is entitled to free travel on BKV (Budapest public transportation) services, but she needs to prove her pensioner status to the driver or inspector. According to the BKV website, foreign pensioners can use their national pensioner ID or a document that proves their pensioner status, along with a valid ID (e.g., passport). The document should be in English or have an English translation. Here are some options your mother can consider: \dots {\color{purple}[Hungarian/\llama]} \\
A sequence of at least 20 tokens is repeated at least 4 times
& Ke eng se se neng se le pharologano ka lenyalo la ga Sir Seretse Khama le Ruth Khama mme Ruth Khama o ne a itshwara jang lefa go nale pharologanyo eo? \textit{What was different about Sir Seretse Khama and Ruth Khama's marriage and how did Ruth Khama behave despite the difference?}
&  Ke a leboga gore o ne a reng ditlhong tse di ntle. Ka kakaretso, ga Sir Seretse Khama le Ruth Khama o ne a itshwara jang lefa go nale pharologano ya gagwe. Ruth Khama o ne a neng a reng mabaka a a latelang a a nang le bona, a a nang le lefatsheng la Botswana, a a nang le baagi ba bangwe ba bangata ba bangwe ba bangwe ba bangwe ba bangwe ba bangwe ba bangwe ba bangwe ba bangwe ba bangwe ba bangwe ba bangwe ba bangwe ba bangwe ba bangwe ba bangwe ba bangwe ba bangwe ba bangwe ba bangwe \dots
{\color{purple}[Tswana/\ayanew]} \\
The model apologizes and is unable to answer the question
& Duni daddarba leh yan iyyaan duma duni core kee akah yan iyya? \textit{How do they know how hot the earth's core is?}
& I'm sorry for any confusion, but it seems like the text you've provided is in a language that I don't currently support. I can understand and generate text in several languages, but I need to know which one you're using to provide an accurate response. If you're able to translate your question into English, I'd be more than happy to help! {\color{purple}[Afar/\mixtral]} \\
The model recognizes the language of the text
&
Umugirwa wo gufuba umwami wagenda gute mu Burundi?
\textit{How was the Kings' funeral done in Burundi?}
& A question in Kirundi! Umugirwa wo gufuba umwami wagenda gute mu Burundi? Translation: ``What is the significance of the drum in Burundian culture?'' In Burundian culture, drums (umugirwa) play a significant role in various aspects of life, including: \dots {\color{purple}[Kirundi/\llama]} \\
The model translates the question into English
& Mahimo bala kontrolon ukon himuon sang tawo ang tiempo? Kon mahimo, hasta diin kag paano? \textit{Can weather be controlled or man made? If so, to what extent and how?}
&
What a fascinating question! In English, your question translates to: ``Can we control time like we control a person? If so, up to what extent and how?'' Unfortunately, the answer is no, we cannot control time like we control a person. Time is a fundamental concept that governs the universe, and it is not a physical entity that can be manipulated or controlled. \dots {\color{purple}[Hiligaynon/\llama]} \\
\bottomrule
\end{tabular}
}
}
\caption{Examples of issues in model responses that are detected as surface-level errors.}
\label{tab:faulty_generations_examples}
\end{table*}

\begin{figure*}
\centering
{\renewcommand{\arraystretch}{1.5}
\resizebox{1\textwidth}{!}{%
\begin{tabular}{p{\textwidth}}
\toprule
You will be provided with a question and an answer, both translated from \textit{[language]} into English. The answer was produced by an LLM. \\

Your task is to evaluate whether the answer is relevant to the question. **Relevance means that the answer makes a genuine attempt to address the question as stated.** The factual accuracy of the answer is not important. An answer should be judged as "irrelevant" if it: \par
- Does not attempt to answer the question, \par
- Focuses on a completely different topic, \par
- Is merely a refusal to answer, or \par
- Simply translates the content of the question without addressing it. \\

**Instructions:** \\

1. Read the question and the answer carefully. \par
2. Provide an explanation of your reasoning enclosed in \textasciigrave<explanation></explanation>\textasciigrave tags.
3. Then, give your final judgment enclosed in \textasciigrave<judgment></judgment>\textasciigrave tags with either the word \textasciigrave{}relevant\textasciigrave or \textasciigrave{}irrelevant\textasciigrave. \\

**Content:** \\

**Question:** \par
<question> \par
\textit{[question]} \par
</question> \\

**Answer:** \par
<answer> \par
\textit{[answer]} \par
</answer> \\
\bottomrule
\end{tabular}
}
\caption{Prompt used with \gpto{} to determine the relevance of an LLM-generated answer to its question. Strings in the form \textit{[form]} are placeholders that are replaced at runtime. The question and answer are provided in their English translations.}
\label{tab:relevance_prompt}
}
\end{figure*}

\begin{figure*}
    \centering
    \adjustbox{trim=1.2cm 1cm 2cm 0.3cm}{%
      \includeinkscape[width=1.13\linewidth]{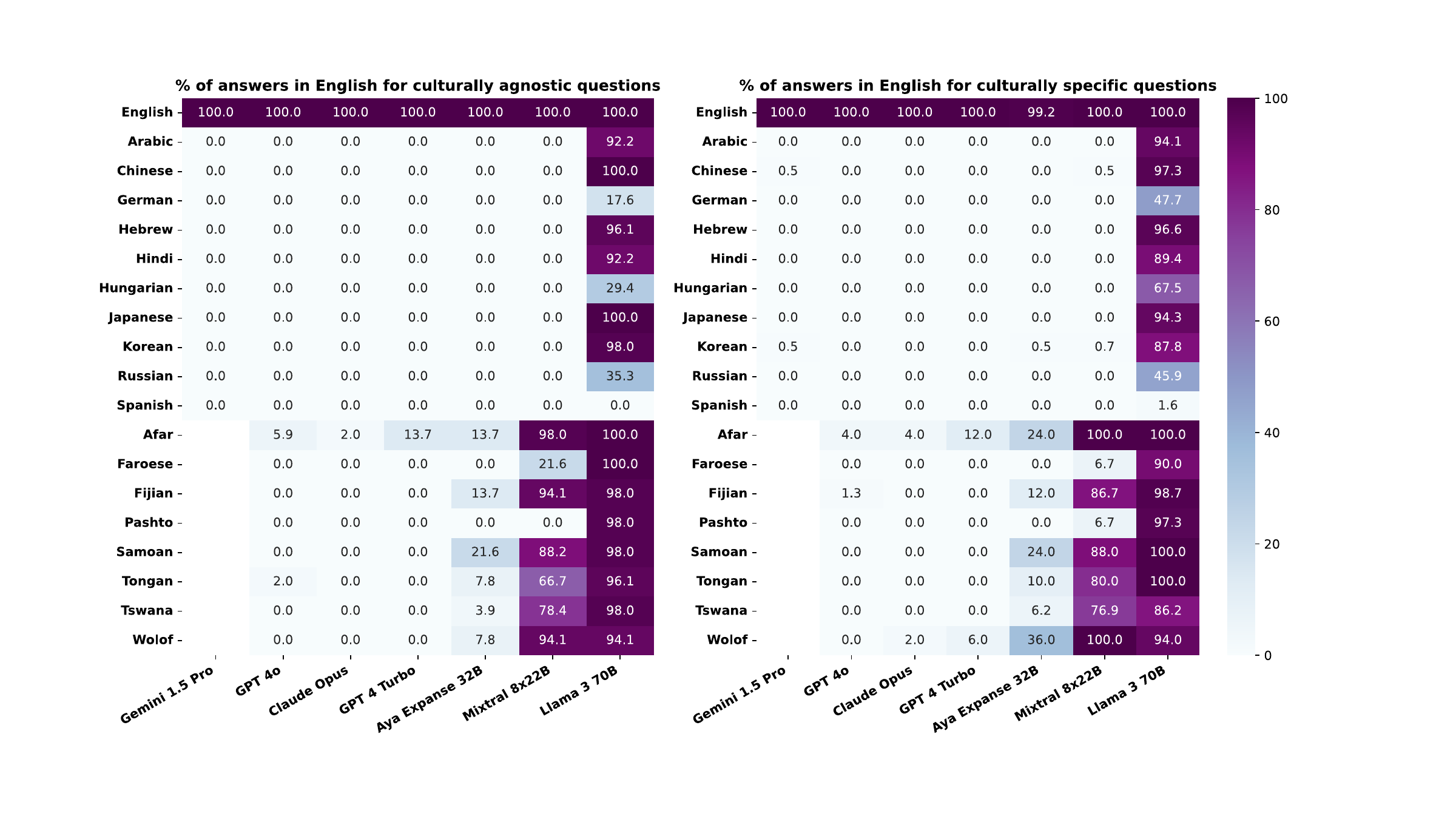}%
    }
    \caption{Percentage of answers produced in English by model which produced the answer for culturally specific and culturally agnostic questions. Blank cells for \gemini{} indicate languages where the API returned an error message.}
    \label{app_fig:english_answer}
\end{figure*}

\paragraph{Answer statistics:} We compute the lengths of generated answers using \texttt{tiktoken} with the \texttt{o200k\_base} encoding. \autoref{app_tab:answer_stats} presents statistics for the length of answers generated by each model. To account for variations in token count due to the language of generation and the presence of repetitions, we provide separate statistics for all answers and for those produced in the correct language without repetitions. Finally, we provide the percentage of answers produced in English for a non-English question in \autoref{app_fig:english_answer}.

\begin{table*}[ht]
    \centering
    \begin{tabular}{lcccccc}
        \toprule
        & \multicolumn{3}{c}{All Data} & \multicolumn{3}{c}{Correct Lang / No Repetitions} \\
        \cmidrule(lr){2-4} \cmidrule(lr){5-7}
        \textsc{Model} & \textsc{Mean} & \textsc{Median} & \textsc{Std} & \textsc{Mean} & \textsc{Median} & \textsc{Std} \\
        \midrule
        \claude{} & 296.4 & 293 & 88.9 & 302.2 & 297 & 79.2 \\
        \gptturbo{} & 472.6 & 482 & 155.2 & 468.9 & 477 & 147.2 \\
        \gpto{} & 446.6 & 425 & 268 & 434.9 & 430 & 184.8 \\
        \gemini{} & 265.6 & 270 & 247.1 & 421.6 & 421 & 177.7 \\
        \midrule
        \ayanew{} & 449.4 & 437 & 187.3 & 476.3 & 460 & 289.7 \\
        \llama{} & 395.9 & 410 & 171.4 & 478.7 & 484 & 138.8 \\
        \mixtral{} & 305.3 & 237 & 281.9 & 255.4 & 252 & 114 \\
        \bottomrule
    \end{tabular}
    \caption{Mean, median, and standard deviation of token counts in answers generated by different models. To account for variations in token count due to the language of generation and the presence of repetitions, we provide separate statistics for all answers and for answers produced in the correct language without repetitions. Token counts were computed using \texttt{tiktoken} with the \texttt{o200k\_base} encoding.}
    \label{app_tab:answer_stats}
\end{table*}

\section{Human Evaluation}
\label{app:hum_eval_details}

In this section, we present the details of human evaluation.

\begin{figure*}[tbp]
  \includegraphics[width=1\linewidth]{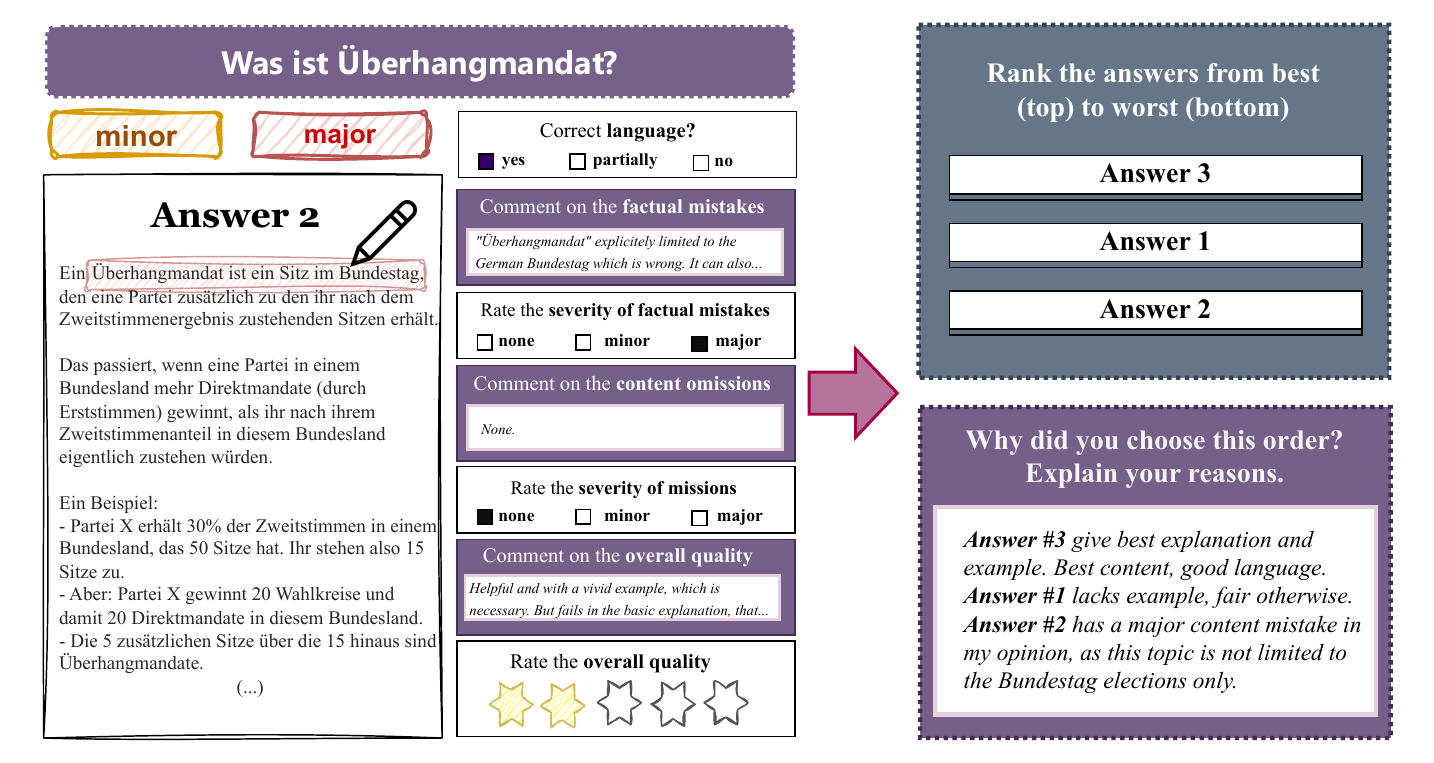} 
  \caption{Our human evaluation pipeline. The annotator has to first read the answer, mark and classify all the mistake, and then comment and rate different properties of the answer. Once they have completed evaluating all three answers they are asked to rank them with respect to each other and provide a justification for the ranking. The example shows a culturally specific questions and one answer in German. The answer was produced by \claude.}
  \label{fig:humeval}
\end{figure*}

\paragraph{Evaluation Task} The evaluation was conducted using \texttt{LabelStudio} \citep{LabelStudio}. On the UI, annotators were presented with a question, a gold answer (if applicable), and three competitive answers in random order. The annotation process for each answer involved: (1) marking any mistakes,\footnote{This step was included to help the annotators visualize any issues with the answer.} (2) stating whether the answer is in the correct language, (3) evaluating factual accuracy, (4) noting any content omissions, (5) commenting on the overall quality of each answer, (6) rating each answer on a 5-point scale (excellent, good, average, poor, unusable). Upon completing the ratings, annotators ranked the three answers from best to worst and provided a free-form explanation for their ranking. \autoref{fig:humeval} illustrates the overall flow of the evaluation task. The study was submitted for the review to Institutional Review Board and received a \textit{non-human subject} determination. 

\paragraph{Guidelines and Consent} We provided human evaluation guidelines, describing how to use the interface (including a tutorial video) and explaining each of the steps in the annotation process. The guidelines link to the consent form.

\paragraph{Data}
Human evaluation was done for answers generated by \claude, \gptturbo, and \mixtral{} for questions in English, German, Hindi, Fijian and Kirundi. For culturally specific questions, annotators chose 10 questions in their language that they felt confident they knew the answer to. For culturally agnostic questions, we sampled 10 English culturally agnostic questions, and used the original English and the translations into the 4 other languages. We provided annotators with bullet-point answers in English for the culturally agnostic questions.

\paragraph{Workers and Cost} German and Hindi annotators were recruited via Prolific, while Fijian and Kirundi annotators were recruited via Upwork. English annotations were performed by one of the authors. All annotators were native speakers of their respective languages and had participated in the question collection. Each question took approximately 20--40 minutes to evaluate, with annotators receiving compensation of \$7.50 USD per question and an additional \$8.00 USD for reviewing the guidelines, totaling \$158 USD per language. The overall cost of the evaluation amounted to approximately \$720 USD.\footnote{We also covered Upwork charges which the platform impose on the freelancers.}

\begin{figure*}[tbp]
    \centering
    \begin{minipage}{0.49\textwidth}
        \centering
        \includegraphics[width=\textwidth]{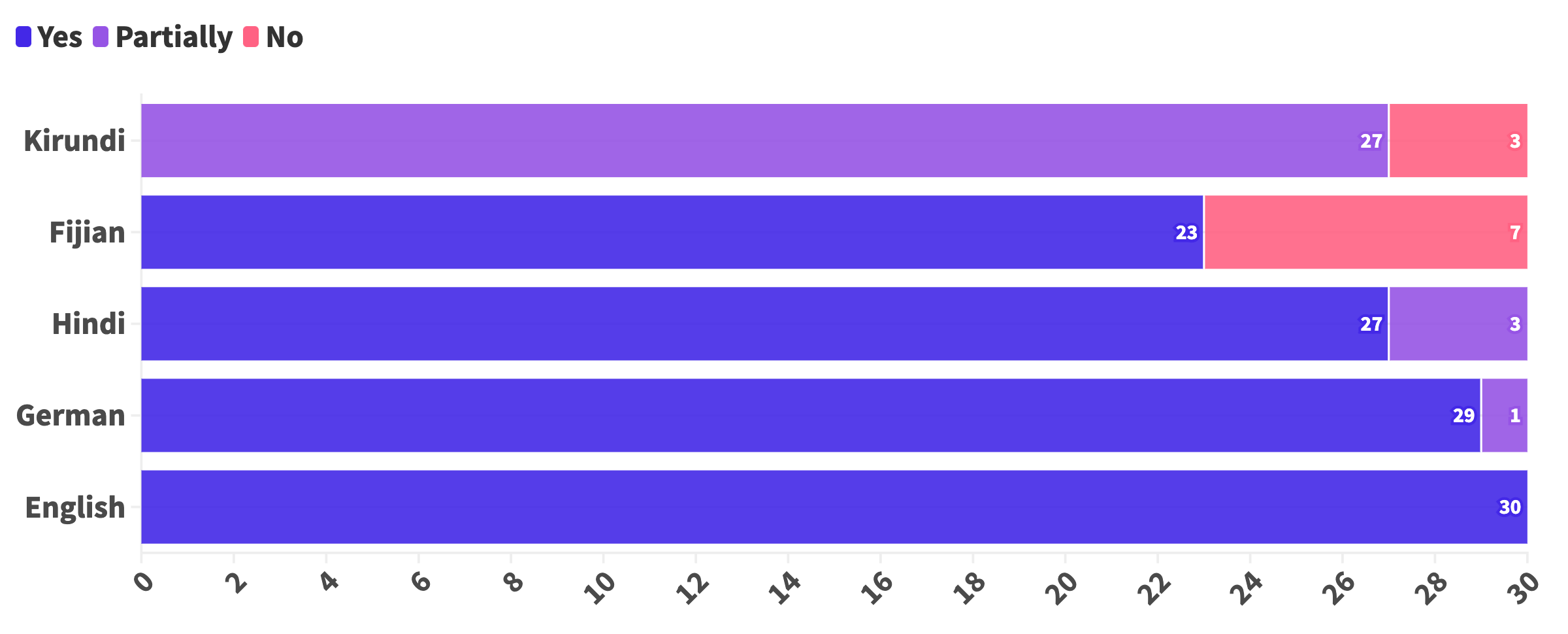}
        \caption{Annotations on Language Correctness for Culturally Specific Questions}
        \label{fig:cult_lang_by_language}
    \end{minipage}%
    \hspace{0.01\textwidth}
    \begin{minipage}{0.49\textwidth}
        \centering
        \includegraphics[width=\textwidth]{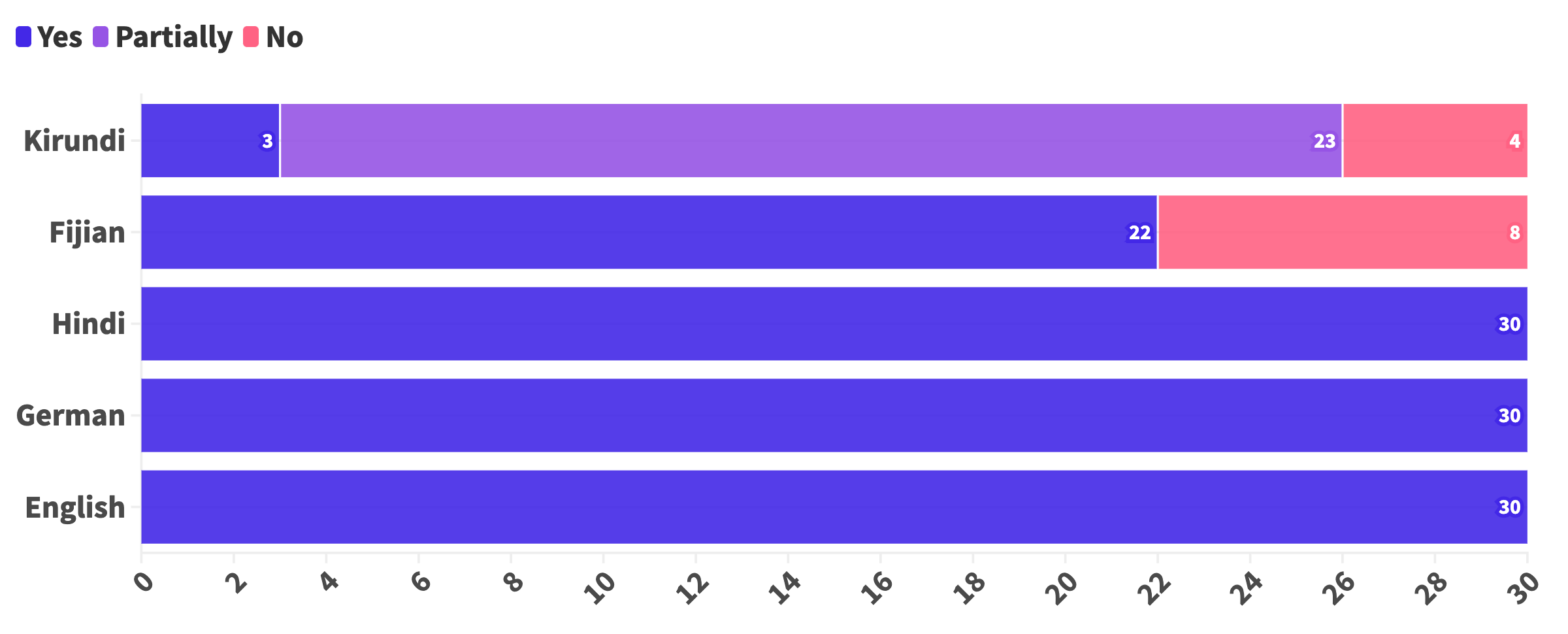}
        \caption{Annotations on Language Correctness for Agnostic-Specific Questions}
        \label{fig:non_cult_lang_by_language}
    \end{minipage}
\end{figure*}

\begin{figure*}[tbp]
    \centering
    \begin{minipage}{0.49\textwidth}
        \centering
        \includegraphics[width=\textwidth]{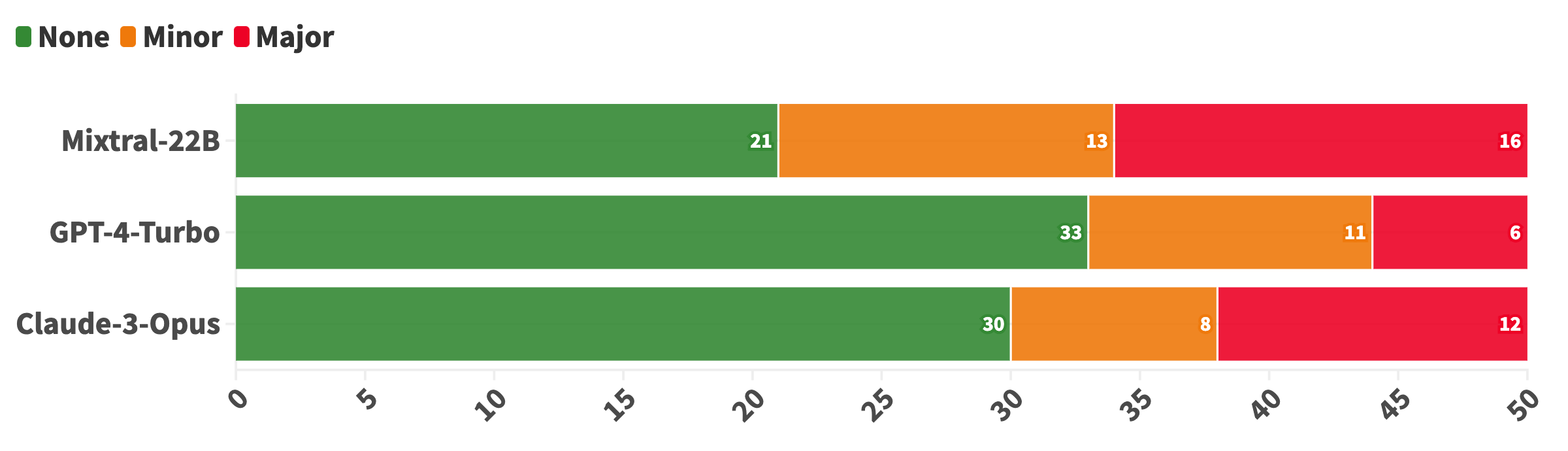}
        \caption{Factuality issues as assessed by the annotators by model for \textbf{culturally specific} questions}
        \label{fig:cult_fact_by_model}
    \end{minipage}%
    \hspace{0.01\textwidth}
    \begin{minipage}{0.49\textwidth}
        \centering
        \includegraphics[width=\textwidth]{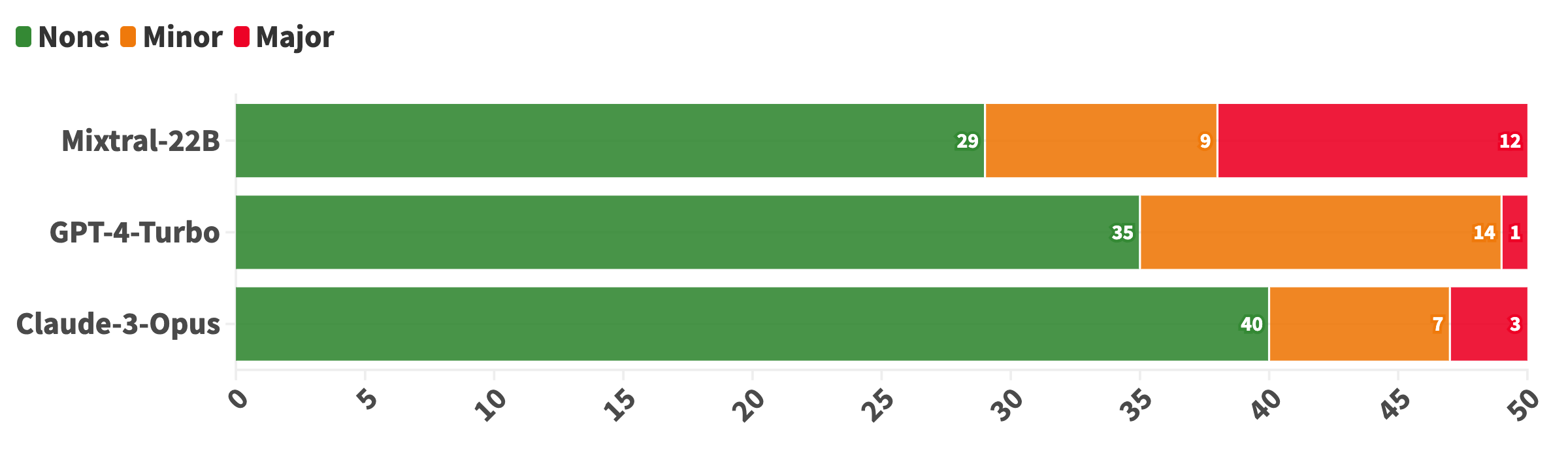}
        \caption{Factuality issues as assessed by the annotators by model for \textbf{culturally agnostic} questions}
        \label{fig:non_cult_fact_by_model}
    \end{minipage}
\end{figure*}

\begin{figure*}[tbp]
    \centering
    \begin{minipage}{0.49\textwidth}
        \centering
        \includegraphics[width=\textwidth]{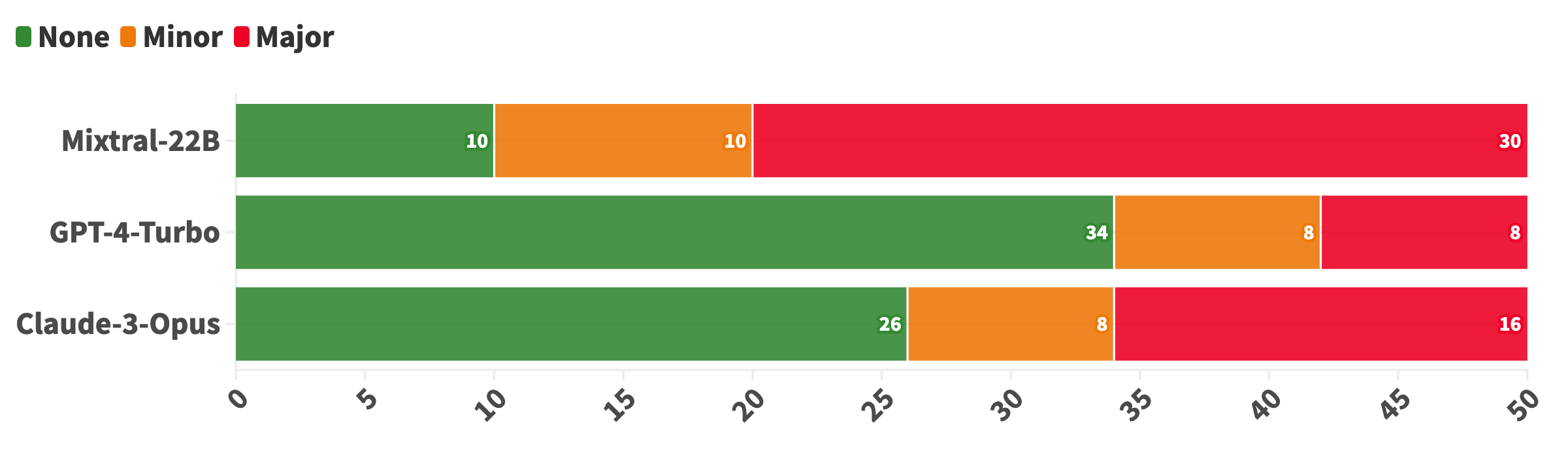}
        \caption{Omissions as assessed by the annotators by model for \textbf{culturally specific} questions}
        \label{fig:cult_omit_by_model}
    \end{minipage}%
    \hspace{0.01\textwidth}
    \begin{minipage}{0.49\textwidth}
        \centering
        \includegraphics[width=\textwidth]{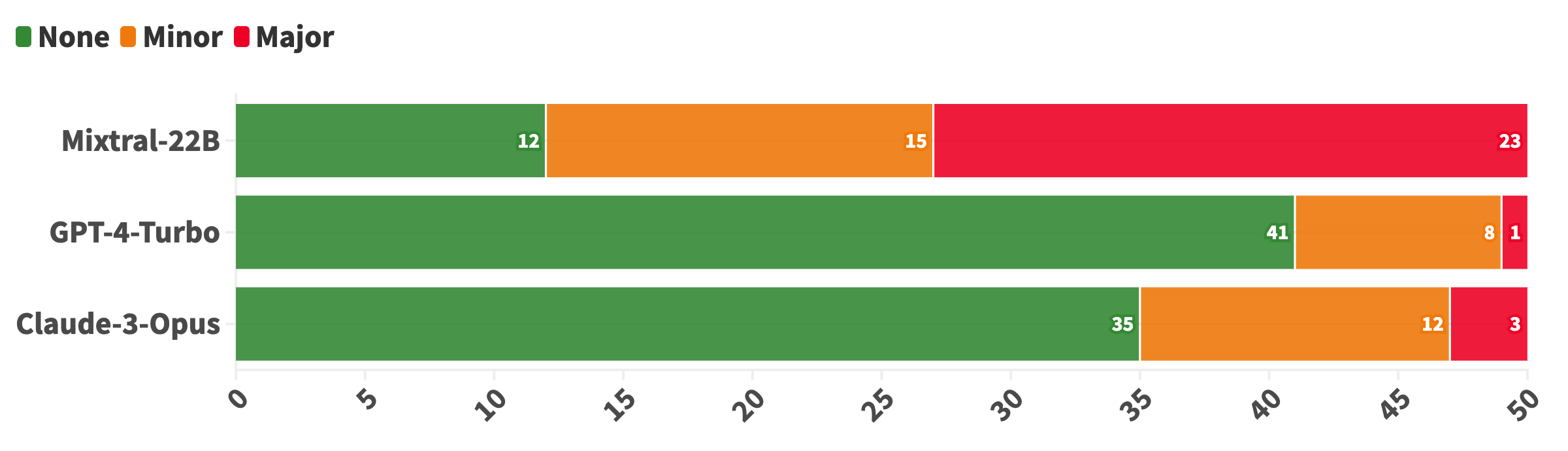}
        \caption{Omissions as assessed by the annotators by model for \textbf{culturally agnostic} questions}
        \label{fig:non_cult_omit_by_model}
    \end{minipage}
\end{figure*}

\begin{figure*}[tbp]
    \centering
    \begin{minipage}{0.49\textwidth}
        \centering
        \includegraphics[width=\textwidth]{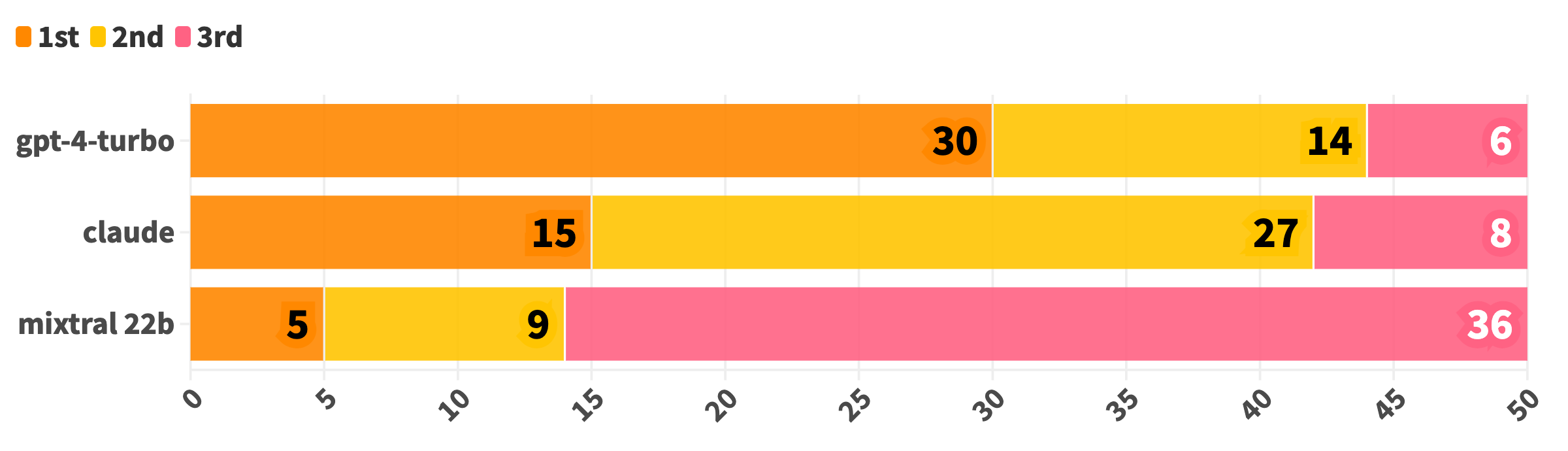}
        \caption{Number of times each model was ranked as \textit{first}, \textit{second}, and \textit{last} for \textbf{culturally specific} questions.}
        \label{fig:cult_ranking_by_model}
    \end{minipage}%
    \hspace{0.01\textwidth}
    \begin{minipage}{0.49\textwidth}
        \centering
        \includegraphics[width=\textwidth]{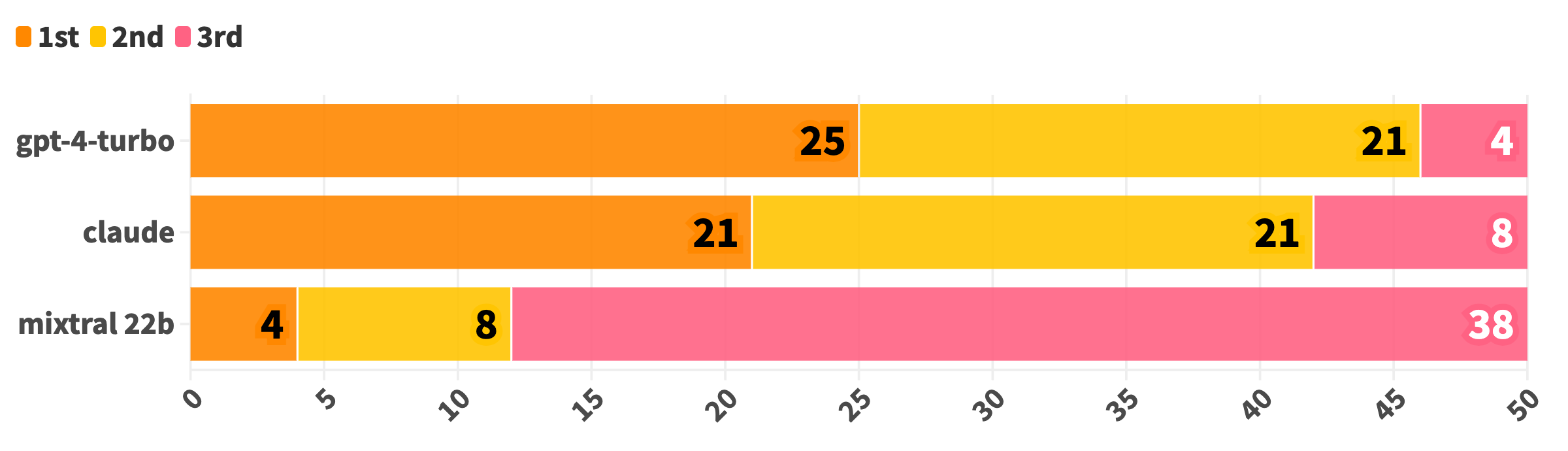}
        \caption{Number of times each model was ranked as \textit{first}, \textit{second}, and \textit{last} for \textbf{culturally agnostic} questions.}
        \label{fig:non_cult_ranking_by_model}
    \end{minipage}
\end{figure*}

\begin{figure*}[tbp]
\centering
\includegraphics[width=1\textwidth]{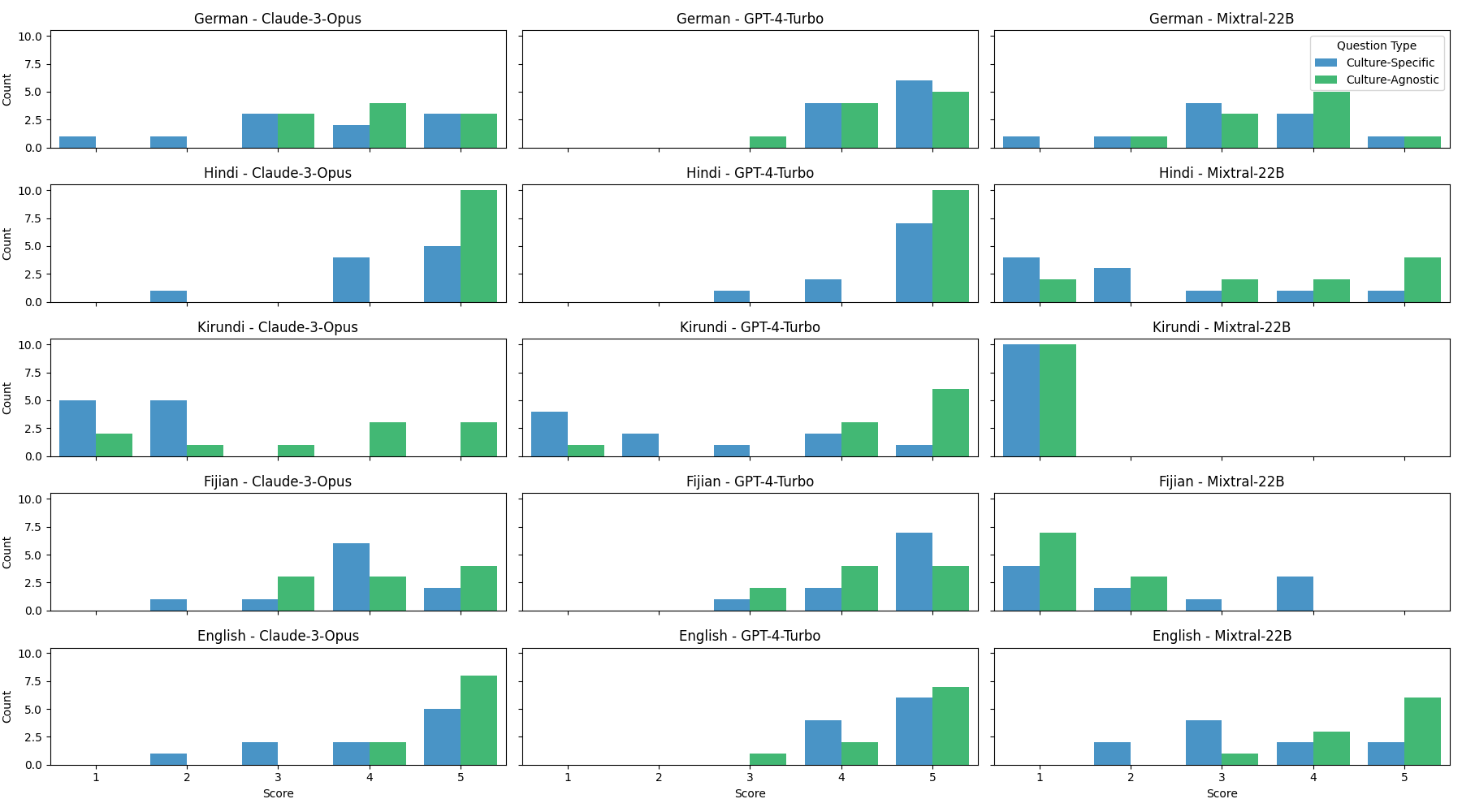} 
\caption{Scores distribution by language and model for Culturally Specific and Culturally Agnostic questions}
\label{fig:ratings_by_model_by_qtype_by_lang}
\end{figure*}

\paragraph{Results} \autoref{fig:cult_lang_by_language} and \autoref{fig:non_cult_lang_by_language} show the results of annotation for whether the answer was generated in the same language as the question (see \autoref{tab:language_correct_all_data} for detailed counts). \autoref{fig:cult_fact_by_model} and \autoref{fig:non_cult_fact_by_model} display the annotations of the severity of factual issues in each answer (see \autoref{tab:factuality_issues_by_model_by_lang} for detailed counts). \autoref{fig:cult_omit_by_model} and \autoref{fig:non_cult_omit_by_model} present the annotations of the severity of omissions in each answer (see \autoref{tab:omission_issues_by_model_by_lang} for detailed counts). \autoref{fig:cult_ranking_by_model} and \autoref{fig:non_cult_ranking_by_model} show the rankings of the models for both culturally specific and culturally agnostic questions. \autoref{fig:ratings_by_model_by_qtype} showsh ratings by model by question type. Finally, \autoref{fig:ratings_by_model_by_qtype_by_lang} shows the distributions of scores assigned for each model by the question type and language of generation.

\paragraph{Statistical analysis} We conducted a statistical analysis using the \texttt{clmm()} function from the \texttt{ordinal} package in R. Each model was fitted with the ordinal ratings (1--5) as the response variable and different predictors, allowing for random intercepts for annotators. \autoref{app:stats_agnos_sign_better_score} shows the results of a model with question type (either culturally specific or culturally agnostic) as the predictor. \autoref{app:clmm_main} presents the results of an analysis with model type, question type, and their interaction as predictors, complemented by \autoref{app:post_hoc_main}, which shows the results of a post-hoc analysis. Finally, \autoref{app:r_squared_diff_predictors} displays the R\textsuperscript{2} values for models with different predictors, namely model type, question type, omission ratings, factuality ratings, and language accuracy ratings.

\begin{table*}[t]
\centering
\begin{tabular}{lcccc}
\toprule
Formula & \multicolumn{4}{c}{\texttt{rating} $\sim$ \texttt{model} + (1 | \texttt{language/annotator})} \\
\midrule
 & \multicolumn{4}{c}{Random Effects} \\
\cmidrule(r){2-5}
Group & Name & Variance & Std. Dev. & \\
\midrule
language & (Intercept) & 0.7175 & 0.847 \\
\midrule
Number of groups: & \multicolumn{4}{c}{5} \\
\midrule
 & \multicolumn{4}{c}{Fixed Effects} \\
\cmidrule(r){2-5}
Coefficient & Estimate & Std. Error & z value & Pr($>|z|$) \\
\midrule
\gptturbo & 0.8635 & 0.2885 & 2.993 & 0.00276 ** \\
\mixtral & -1.9493 & 0.2844 & -6.854 & 7.18e-12 *** \\
\midrule
Significance codes: & \multicolumn{4}{c}{0 ‘***’ 0.001 ‘**’ 0.01 ‘*’ 0.05 ‘.’ 0.1 ‘ ’ 1} \\
\bottomrule
    \end{tabular}
    \caption{Results of cumulative link mixed model with ordinal ratings as the response variable and model as the predictor.}
    \label{tab:clmm_model}
\end{table*}

\begin{table*}[t]
    \centering
    \begin{tabular}{lcccccc}
        \toprule
        \textbf{Contrast} & \textbf{Estimate} & \textbf{SE} & \textbf{df} & \textbf{\textit{z}-ratio} & \textbf{\textit{p}-value} \\
        \midrule
        \claude{} -- \gptturbo{} & -0.863 & 0.288 & Inf & -2.993 & \textbf{0.0078} \\
        \claude{} -- \mixtral{} & 1.949 & 0.284 & Inf & 6.854 & \textbf{<.0001} \\
        \gptturbo{} -- \mixtral{} & 2.813 & 0.315 & Inf & 8.936 & \textbf{<.0001} \\
        \midrule
        \multicolumn{5}{l}{\textbf{P value adjustment:} Tukey method for comparing a family of 3 estimates} \\
        \bottomrule
    \end{tabular}
    \caption{Post-hoc analysis for the model in \autoref{tab:clmm_model}. Tests performed using the \texttt{emmeans} library in R.}
    \label{tab:posthoc_clmm_model}
\end{table*}

\begin{table*}[t]
\centering
\begin{tabular}{lcccc}
\toprule
Formula & \multicolumn{4}{c}{\texttt{rating} $\sim$ \texttt{type} + (1 | \texttt{language/annotator})} \\
\midrule
 & \multicolumn{4}{c}{Random Effects} \\
\cmidrule(r){2-5}
Group & Name & Variance & Std. Dev. & \\
\midrule
annotator & (Intercept) & 0.9418 & 0.9705 & \\
Number of Groups: annotator & \multicolumn{4}{c}{5} \\
\midrule
 & \multicolumn{4}{c}{Fixed Effects} \\
\cmidrule(r){2-5}
Coefficient & Estimate & Std. Error & z value & Pr($>|z|$) \\
\midrule
Culturally Agnostic & 0.7259 & 0.2192 & 3.312 & 0.000926*** \\
\midrule
Significance codes: & \multicolumn{4}{c}{0 ‘***’ 0.001 ‘**’ 0.01 ‘*’ 0.05 ‘.’ 0.1 ‘ ’ 1} \\
\bottomrule
\end{tabular}
\caption{Results of cumulative link mixed model with ordinal ratings as the response variable and question type (culturally specific vs culturally agnostic) as the predictor.}
\label{app:stats_agnos_sign_better_score}
\end{table*}

\begin{table*}[t]
\centering
{\renewcommand{\arraystretch}{1.5}
\resizebox{0.9\textwidth}{!}{%
\begin{tabular}{lcccccc}
\toprule
\textbf{Formula} & \multicolumn{6}{l}{\texttt{rating} $\sim$ \texttt{model} * \texttt{type} + (1 | \texttt{language}/\texttt{annotator})} \\
\midrule
\multicolumn{7}{l}{\textbf{Random effects:}} \\
\midrule
\textbf{Groups} & \textbf{Name} & \textbf{Variance} & \textbf{Std.Dev.} & \multicolumn{3}{c}{} \\
Language & (Intercept) & 0.7566 & 0.8698 & \multicolumn{3}{c}{} \\
Annotator:Language & (Intercept) & 0.7566 & 0.8698 & \multicolumn{3}{c}{} \\
\midrule
\multicolumn{7}{l}{\textbf{Fixed effects:}} \\
\midrule
\textbf{Coefficient} & \textbf{Estimate} & \textbf{Std. Error} & \textbf{\textit{z}-value} & \textbf{Pr(>|z|)} & \multicolumn{2}{c}{} \\
\midrule
\gptturbo & 1.1679 & 0.3868 & 3.020 & 0.002531 & ** & \\
\mixtral & -1.7013 & 0.3853 & -4.415 & 1.01e-05 & *** & \\
Culturally Agnostic & 1.3561 & 0.3956 & 3.428 & 0.000607 & *** & \\
\gptturbo:Culturally Agnostic & -0.6182 & 0.5849 & -1.057 & 0.290489 & & \\
\mixtral:Culturally Agnostic  & -0.7163 & 0.5467 & -1.310 & 0.190109 & & \\
\bottomrule
\end{tabular}
}
}
\caption{Cumulative link mixed model fitted with the Laplace approximation fitted with \texttt{clmm()} in R. The response variable is the ratings (an ordinal variable on a 5-point scale), with predictors being model (\claude, \gptturbo, or \mixtral) and question type (culturally specific and culturally agnostic). Annotator nested within language is included as a random effect. The baseline model is \claude\ and the baseline question type is culturally specific. Model's conditional $R^2$ is 0.497 (including random effects) and marginal $R^2$ is 0.266 (only fixed effects). Please refer to \autoref{app:post_hoc_main} for post-hoc analysis.}
\label{app:clmm_main}
\end{table*}

\begin{table*}[t]
\centering
\begin{tabular}{lcccccc}
\toprule
\textbf{Contrast} & \textbf{Estimate} & \textbf{SE} & \textbf{df} & \textbf{\textit{z}-ratio} & \textbf{\textit{p}-value} \\
\midrule
Spec. \claude - Agn. \claude               & -1.356 & 0.396 & Inf & -3.428 & \textbf{0.0091} \\
Spec. \claude  - Spec. \gptturbo      & -1.168 & 0.387 & Inf & -3.020 & \textbf{0.0380} \\
Spec. \claude  - Agn. \gptturbo            & -1.906 & 0.424 & Inf & -4.492 & \textbf{0.0001} \\
Spec. \claude  - Spec. \mixtral               &  1.701 & 0.385 & Inf &  4.415 & \textbf{0.0002} \\
Spec. \claude  - Agn. \mixtral           &  1.061 & 0.374 & Inf &  2.835 & 0.0687 \\
Agn. \claude - Spec. \gptturbo           &  0.188 & 0.412 & Inf &  0.457 & 1.0000 \\
Agn. \claude - Agn. \gptturbo      & -0.550 & 0.442 & Inf & -1.242 & 1.0000 \\
Agn. \claude - Spec. \mixtral           &  3.057 & 0.429 & Inf &  7.123 & \textbf{<.0001} \\
Agn. \claude - Agn. \mixtral     &  2.418 & 0.414 & Inf &  5.842 & \textbf{<.0001} \\
Spec. \gptturbo - Agn. \gptturbo         & -0.738 & 0.436 & Inf & -1.694 & 1.0000 \\
Spec. \gptturbo - Spec. \mixtral              &  2.869 & 0.420 & Inf &  6.836 & \textbf{<.0001} \\
Spec. \gptturbo- Agn. \mixtral        &  2.229 & 0.404 & Inf &  5.514 & \textbf{<.0001} \\
Agn. \gptturbo - Spec. \mixtral        &  3.607 & 0.462 & Inf &  7.800 & \textbf{<.0001} \\
Agn. \gptturbo - Agn. \mixtral  &  2.967 & 0.445 & Inf &  6.669 & \textbf{<.0001} \\
Spec. \mixtral - Agn. \mixtral       & -0.640 & 0.383 & Inf & -1.669 & 1.0000 \\
\bottomrule
\end{tabular}
\caption{Post-hoc analysis for the model in \autoref{app:clmm_main} with Bonferroni adjustment. Spec. refers to culturally specific questions while Agn. refers to culturally agnostic questions. Tests performed using the \texttt{emmeans} library in R.}
\label{app:post_hoc_main}
\end{table*}

\begin{table*}[h]
\centering
\begin{tabular}{lcc}
\toprule
\textbf{Predictor} & \textbf{Conditional R\textsuperscript{2}} & \textbf{Marginal R\textsuperscript{2}} \\
\midrule
Model               & 0.214 & 0.189 \\
Omission            & 0.752 & 0.740 \\
Factuality          & 0.614 & 0.560 \\
Language Acc.       & 0.339 & 0.327 \\
Q-Type              & 0.093 & 0.061 \\
Model * Q-Type      & 0.497 & 0.266 \\
\bottomrule
\end{tabular}
\caption{Conditional and Marginal R\textsuperscript{2} values for different predictors. We fit cumulative link mixed models (\texttt{clmm()} in R) with \textit{ratings} as the response variable and different predictors. All models included random intercepts for annotators. Omission, Factuality, and Language Accuracy were treated as ordinal variables (no issues > minor issues > major issues), whereas Q-Type and Model are categorical variables with two and three levels respectively. The last model was fitted with the interaction between the Model and the Q-Type. The Conditional R\textsuperscript{2} refers to the variance explained by both fixed effects (predictors) and random effects (annotators), while Marginal R\textsuperscript{2} refers to the variance explained by fixed effects only.}
\label{app:r_squared_diff_predictors}
\end{table*}

\begin{table*}[t]
\centering
{\renewcommand{\arraystretch}{1.5}
\resizebox{1\textwidth}{!}{%
\begin{tabular}{p{0.8cm}cp{2cm}cp{9cm}}
\hline
\textsc{Model} & \textsc{Win Rate} & \textsc{Reason} & \textsc{\# (\textit{spec.}/\textit{agn.})} & \textsc{Comment} \\
\hline
\multirow{5}{*}{\rotatebox[origin=c]{90}{\gptturbo}} & \multirow{5}{*}{55\%} & \textsc{Content} & 27 / 24 & \textit{Answer 1 (\gptturbo) is the perfect answer and and explains all the points needed to understand how to play the game 'Teen Patti'.} \\
 &  & \textsc{Language/ Format} \raggedright & 11 / 7 & \textit{Answer 3 (\gptturbo) is very well structured and easy to follow. It covers all the information as well. }  \\
&  & \textsc{Factuality/ Relevance}  \raggedright & 10 / 3 & \textit{A3 (\gptturbo) is more factual than A1 (\claude) and A2 (\mixtral).} \\
\midrule
\multirow{5}{*}{\rotatebox[origin=c]{90}{\claude}}  & \multirow{5}{*}{36\%} & \textsc{Content} & 6 / 16 & \textit{Answer 3 (\claude) covers the topic in its entirety and hence is ranked 1st.} \\
 &  & \textsc{Language/ Format} \raggedright & 5 / 13 & \textit{Answer 2 (\claude) is more readable because the information is listed as points.} \\
 & & \textsc{Factuality/ Relevance}  \raggedright & 6 / 6 & \textit{A3 (\claude) is more detailed and factual than A1 (\gptturbo) and A2 (\mixtral).} \\
 \midrule
\multirow{6}{*}{\rotatebox[origin=c]{90}{\mixtral}}  & \multirow{5}{*}{9\%}& \textsc{Content} & 2 / 1 & \textit{A1 (\mixtral) is better explained than A2 (\claude) and A3 (\gptturbo).} \\
 &  & \textsc{Language/ Format} \raggedright & 3 / 2 & \textit{All answers have equal quality content, so they are distinguished by their structure/verbosity. Answer 3 (\mixtral) has a very natural structure. Answer 2 (\gptturbo) and answer 1 (\claude) have redundancies and answer 1's are slightly worse.} \\
 &  &\textsc{Factuality/ Relevance}  \raggedright & 1 / 0 & \textit{Answer 2 (\mixtral) was slightly more specific to Western Europe than Answer 1 (\gptturbo), but both were roughly equal in quality.} \\
\hline
\end{tabular}
}
}
\caption{Win rates of the three models in human-evaluated 3-way comparisons of answers for 100 questions. Reasons behind the annotators' decisions are provided, with separate reason counts for \textit{culturally specific} and \textit{culturally agnostic} questions. A breakdown of reasons into finer-grained categories is provided in Table \ref{tab:categories_of_rank_annotations}.}
\label{tab:model_winrate_reason}
\end{table*}

\begin{table*}[tbp]
\centering
\begin{tabular}{@{}llccc@{\hspace{1cm}}lccc@{}}
\toprule
\textsc{Language} & \textsc{Model} & \multicolumn{3}{c}{\textsc{Culturally Specific}} & \multicolumn{4}{c}{\textsc{Culturally Agnostic}} \\
\cmidrule(lr){3-5} \cmidrule(lr){6-9}
 & & \textsc{Yes} & \textsc{Partially} & \textsc{No} &  & \textsc{Yes} & \textsc{Partially} & \textsc{No} \\
\midrule
\addlinespace
\multirow{3}{*}{German} & \claude & 10 & 0 & 0 &  & 10 & 0 & 0 \\
 & \gptturbo & 10 & 0 & 0 &  & 10 & 0 & 0 \\
 & \mixtral & 9 & 1 & 0 &  & 10 & 0 & 0 \\
\midrule
\addlinespace
\multirow{3}{*}{Hindi} & \claude & 9 & 1 & 0 & & 10 & 0 & 0 \\
 & \gptturbo & 10 & 0 & 0 &  & 10 & 0 & 0 \\
 & \mixtral & 8 & 2 & 0 &  & 10 & 0 & 0 \\
\midrule
\addlinespace
\multirow{3}{*}{Kirundi} & \claude & 0 & 10 & 0 &  & 1 & 9 & 0 \\
 & \gptturbo & 0 & 10 & 0 &  & 1 & 9 & 0 \\
 & \mixtral & 0 & 7 & 3 &  & 1 & 5 & 4 \\
\midrule
\addlinespace
\multirow{3}{*}{Fijian} & \claude & 10 & 0 & 0 &  & 10 & 0 & 0 \\
 & \gptturbo  & 10 & 0 & 0 & & 10 & 0 & 0 \\
 & \mixtral & 3 & 0 & 7 & & 2 & 0 & 8 \\
\midrule
\addlinespace
\multirow{3}{*}{English} & \claude & 10 & 0 & 0 &  & 10 & 0 & 0 \\
 & \gptturbo & 10 & 0 & 0 &  & 10 & 0 & 0 \\
 & \mixtral & 10 & 0 & 0 & & 10 & 0 & 0 \\
\bottomrule
\end{tabular}
\caption{Count of instances generated in the language of the question by model and question-type, and the language being evaluated}
\label{tab:language_correct_all_data}
\end{table*}

\begin{table*}[tbp]
\centering
\begin{tabular}{@{}llcccccc@{}}
\toprule
\textsc{Language} & \textsc{Model} & \multicolumn{3}{c}{\textsc{Culturally Specific}} & \multicolumn{3}{c}{\textsc{Culturally Agnostic}} \\
\cmidrule(lr){3-5} \cmidrule(lr){6-8}
 & & \textsc{None} & \textsc{Minor} & \textsc{Major} & \textsc{None} & \textsc{Minor} & \textsc{Major} \\
\midrule
\multirow{3}{*}{German} & \claude & 8 & 1 & 1 & 10 & 0 & 0 \\
 & \gptturbo & 8 & 2 & 0 & 10 & 0 & 0 \\
 & \mixtral & 8 & 2 & 0 & 9 & 1 & 0 \\
\midrule
\multirow{3}{*}{Hindi} & \claude & 7 & 2 & 1 & 10 & 0 & 0 \\
 & \gptturbo & 7 & 3 & 0 & 10 & 0 & 0 \\
 & \mixtral & 1 & 4 & 5 & 6 & 2 & 2 \\
\midrule
\multirow{3}{*}{Kirundi} & \claude & 0 & 1 & 9 & 3 & 4 & 3 \\
 & \gptturbo & 0 & 4 & 6 & 3 & 6 & 1 \\
 & \mixtral & 0 & 0 & 10 & 0 & 0 & 10 \\
\midrule
\multirow{3}{*}{Fijian} & \claude & 7 & 3 & 0 & 8 & 2 & 0 \\
 & \gptturbo & 8 & 2 & 0 & 3 & 7 & 0 \\
 & \mixtral & 5 & 5 & 0 & 5 & 5 & 0 \\
\midrule
\multirow{3}{*}{English} & \claude & 8 & 1 & 1 & 9 & 1 & 0 \\
 & \gptturbo & 10 & 0 & 0 & 9 & 1 & 0 \\
 & \mixtral & 7 & 2 & 1 & 9 & 1 & 0 \\
\bottomrule
\end{tabular}
\caption{Factuality issues in model generation by model, question type and language of the question}
\label{tab:factuality_issues_by_model_by_lang}
\end{table*}

\begin{table*}[tbp]
\centering
\begin{tabular}{@{}llccccccccc@{}}
\toprule
\multirow{2}{*}{\textsc{Language}} & \multirow{2}{*}{\textsc{Model}} & \multicolumn{3}{c}{\textsc{Culturally Specific}} & & \multicolumn{3}{c}{\textsc{Culturally Agnostic}} \\ \cmidrule{3-5} \cmidrule{7-9}
& & None & Minor & Major & & None & Minor & Major \\ \midrule
\multirow{3}{*}{German} 
& \claude{}       & 6 & 1 & 3 & & 7 & 3 & 0 \\
& \gptturbo{}  & 6 & 4 & 0 & & 8 & 2 & 0 \\
& \mixtral{}  & 1 & 7 & 2 & & 3 & 6 & 1 \\ \midrule

\multirow{3}{*}{Hindi} 
& \claude{}       & 8 & 1 & 1 & & 10 & 0 & 0 \\
& \gptturbo{}  & 9 & 0 & 1 & & 10 & 0 & 0 \\
& \mixtral{}   & 3 & 1 & 6 & & 4 & 4 & 2 \\ \midrule

\multirow{3}{*}{Kirundi} 
& \claude{}      & 0 & 0 & 10 & & 4 & 3 & 3 \\
& \gptturbo{}  & 2 & 2 & 6 & & 6 & 3 & 1 \\
& \mixtral{}   & 0 & 0 & 10 & & 0 & 0 & 10 \\ \midrule

\multirow{3}{*}{Fijian} 
& \claude{}       & 6 & 3 & 1 & & 6 & 4 & 0 \\
& \gptturbo{}  & 8 & 1 & 1 & & 7 & 3 & 0 \\
& \mixtral{}  & 3 & 0 & 7 & & 0 & 0 & 10 \\ \midrule

\multirow{3}{*}{English} 
& \claude{}       & 6 & 3 & 1 & & 8 & 2 & 0 \\
& \gptturbo{}   & 9 & 1 & 0 & & 10 & 0 & 0 \\
& \mixtral{}  & 3 & 2 & 5 & & 5 & 5 & 0 \\ \bottomrule

\end{tabular}
\caption{Count of omission issues by severity type, model, and language for culturally specific and culturally agnostic questions}
\label{tab:omission_issues_by_model_by_lang}
\end{table*}

\paragraph{Analysis of the annotations} We conducted manual analyses of the comments provided by the annotators. For each analysis, we iteratively designed an annotation schema to analyze the submitted comments. \autoref{tab:categories_of_rank_annotations} describes the annotation schema used for analyzing the comments on model ranking (i.e., the annotator's reason for ranking a model 1st, 2nd, or 3rd). The results of this analysis are presented in \autoref{tab:reason_for_ranking}. \autoref{tab:categories_factuality_annotations} describes the categories used to analyze the comments on factuality. The results of this analysis are presented in \autoref{app_tab:factuality_comments_annot_counts}. Finally, \autoref{tab:categories_of_general_comment_annotations} describes the categories used to analyze the general comments left by the annotators for each answer. The results of this analysis are presented in \autoref{tab:issue_count_of_general_comment_annotations} and \autoref{tab:merit_count_of_general_comment_annotations}.

\subsection{Additional Insights}   \label{sec:human_eval_insights}

We capture here insights gained from analyzing human evaluation that we could not fit in the main body of text.

\paragraph{Enumerating facts makes responses seem less human-like.} German and Hindi annotators remarked about the presence fact enumerations (often in the form of dot points) for some model answers. For German, the enumeration structure made responses seem artificial (e.g. \textit{`Again very AI made structure. ``here are common methods'' and a following enumeration plus the asterisk titles...'}). For Hindi, listing facts makes the responses not seem human-like, though not necessarily like an AI either (e.g. \textit{`The answer is just stating points on why is smoking harmful, so it neither sounds human-like nor artificial.'}). More broadly, the fact enumeration structure was described negatively in 5 responses, neutrally in 18 responses and positively in 2.

\paragraph{\gptturbo{} made the most grammar/spelling errors.} 9 out of 12 spelling and grammar issues were noted for \gptturbo{} responses. 8 of these issues occurred for Fijian (e.g. `There is a minor error, and the system might have spelled ``nodra"" incorrectly. However, the language content is relevant so the rating is 4 out of 5, and it sounds like a human.') and the last was in German (`Defninetly helpful, complete and clear. Also fluent. One spelling mistake found: Zusammengefasend is no German word should be "zusammengefasst" or similar. But that could be a human-alike typo.'). This mistakes were present in otherwise mostly positive responses, suggesting that the issues were not due to lack of language understanding. We suspect that this phenomenon may be the result of a tokenizer issue.

\begin{table*}
\centering
{\renewcommand{\arraystretch}{1.5}
\resizebox{0.9\textwidth}{!}{%
\begin{tabular}{lp{0.30\textwidth}p{0.50\textwidth}}
\toprule
\textsc{Type} & \textsc{Description} & \textsc{Example}\\
\midrule
\multicolumn{3}{c}{\textit{Content: Completeness/Explanation}} \\
\midrule
\textsc{Completeness} &   The answer was perceived as complete.  & \textit{Answer 3 (Claude-3-Opus) covers the topic in its entirety and hence is ranked 1st.} {\color{purple}[Hindi]} \\
\textsc{Explanation/Examples} &  The answer included useful explanation and/or examples. & \textit{A3 (Gpt-4-Turbo) is better explained than A1 and A2. } {\color{purple}[Fijian]} \\
\textsc{Details/Background} &  The answer included details and/or necessary background. & \textit{Answer 1 (GPT-4-Turbo) and 2 (Claude-3-Opus) are similar but answer 1 has detailed information about the methods to measure body mass compared to 2.} {\color{purple}[Hindi]} \\
\textsc{General} &   The answer was general, which was appropriate for the given question. & \textit{Answer 3 (GPT-4-Turbo) is perfect. The writing style of the answer is the best compared to the other answers. For instance, it mentions the timeline in general rather than pointing out exact years of the event (In my opinion, different sources and online transcripts have a little variation in years in terms of history so it is the best to keep it general).} {\color{purple}[Hindi]} \\
\midrule
\multicolumn{3}{c}{\textit{Language/Presentation}} \\
\midrule
\textsc{Language} &   The answer was fluent/used better language or was less AI-like. & \textit{Both answer 1 (GPT-4-Turbo) and 3 (Claude-3-Opus) are good. Answer 1 sounds more human-like which is why it is ranked 1st. } {\color{purple}[Hindi]}\\
\textsc{Structure} &  The structure of the answer was better. & \textit{All answers have equal quality content, so they are distinguished by their structure/verbosity. Answer 3 (Mixtral-22B) has a very natural structure. } {\color{purple}[English]} \\
\textsc{Simple/Clear/Specific} &  The answer was clear and/or simple, to the point. & \textit{Answer 3 (GPT-4-Turbo) is slightly clearer than answer 1 (Claude-3-Opus).} {\color{purple}[English]} \\
\textsc{Succinct} &   The answer was succinct. &  \textit{All three answers are complete by content in my view (good answer requires more, but that is more than question covers). Answer two (Mixtral-22B) I regard the best, as the density of content in a few lines is awesome - in most of the other questions, "death by long text and details" is valid, here I opt for short and good.}  {\color{purple}[German]} \\
\midrule
\multicolumn{3}{c}{\textit{Factuality/Correctness}} \\
\midrule
\textsc{Factual} &  The answer is better in terms of factuality. & \textit{A3 (GPT-4-Turbo) is more factual than A1 (Claude-3-Opus) and A2 (Mixtral-22B).} {\color{purple}[Fijian]} \\
\textsc{Relevant} & The answer is the most relevant to the question. Often mentioned when other answers were irrelevant. &  \textit{Answer 3 (GPT-4-Turbo) is placed in the first position because it is relevant though not specific to Burundi. It could be used if it were specific. It has some important information (...).} {\color{purple}[Kirundi]} \\
\textsc{No issues} &  There were no apparent issues in the answer. & \textit{Answer 3 (Claude-3-Opus) had no notable issues.} {\color{purple}[English]} \\
\textsc{Partial Answer} &  The answer at least partially addresses the question (while other answers may be refusals, repetitions, or simply irrelevant/wrong). &   \textit{Answer 3 (GPT-4-Turbo) is the only one that tries to answer the question. The other 2 just point out differences between the medications.} {\color{purple}[English]}  \\
\bottomrule
\end{tabular}
}
\caption{Categories used for analysis of reasons for specific ranking of the answers}
\label{tab:categories_of_rank_annotations}
}
\end{table*}

\begin{table*}
\centering
\resizebox{\textwidth}{!}{%
\begin{tabular}{llcccccccccccc}
\toprule
& & \multicolumn{4}{c}{Completeness - Explanation} & \multicolumn{4}{c}{Language - Presentation} & \multicolumn{4}{c}{Factuality - Correctness} \\
\cmidrule{3-14}
Model & Type & Complete & Explanation & Details & General & Language & Structure & Simple & Succinct & Factual & Relevant & No issues & Partial ans \\
\midrule
\claude & \textit{Spec.} & 4 & 0 & 2 & 0 & 1 & 2 & 1 & 1 & 1 & 1 & 2 & 2 \\
\gptturbo & \textit{Spec.} & 11 & 5 & 9 & 2 & 3 & 2 & 4 & 2 & 4 & 1 & 2 & 3 \\
\mixtral & \textit{Spec.} & 1 & 1 & 0 & 0 & 1 & 0 & 1 & 1 & 1 & 0 & 0 & 0 \\
\midrule
\claude & \textit{Agn.} & 11 & 0 & 4 & 1 & 2 & 1 & 7 & 3 & 1 & 1 & 4 & 0 \\
\gptturbo & \textit{Agn.} & 14 & 4 & 6 & 0 & 1 & 4 & 1 & 1 & 0 & 0 & 2 & 1 \\
\mixtral & \textit{Agn.} & 0 & 0 & 1 & 0 & 0 & 1 & 0 & 1 & 0 & 0 & 0 & 0 \\
\bottomrule
\end{tabular}%
}
\caption{Count of different reasons mentioned by the annotator for ranking each model's answer as the best out of three. Note that in some cases more than one reason might have been give by the annotator. \textit{Spec.} refers to Culturally Specific questions, while \textit{Agn.} refers to Culturally Agnostic questions.}
\label{tab:reason_for_ranking}
\end{table*}

\begin{table*}
\centering
{\renewcommand{\arraystretch}{1.5}
\resizebox{0.9\textwidth}{!}{%
\begin{tabular}{lp{0.30\textwidth}p{0.50\textwidth}}
\toprule
\textsc{Type} & \textsc{Description} & \textsc{Comment Example ({\color{purple}Language/Model})} \\
\midrule
\multicolumn{3}{c}{\textit{Direct Factual Errors}} \\
\midrule
\textsc{Date} &   Issues involving incorrect temporal references.  & \textit{It is mentioned that Nifty was launched in 1995 but it was actually launched in 1996.} {\color{purple}[Hindi/\gptturbo]} \\
\textsc{Entity} &  Incorrect entity such as a person, place, or organization. & \textit{Almost everything is incorrect because the answer states that Ntare Rugamba is the person who accepted to die in the place of the king, while Ntare Rugamba is the king who ruled before the King Mwezi Gisabo.} {\color{purple}[Kirundi/\gptturbo]} \\
\textsc{Event} &  Errors in the details or occurrence of events. & \textit{ It says that Aurangzeb got the mosque built at the place, however this claim is very strong. He got the temple destroyed but it is not sure if he got the mosque built, as it was a decade after the demolition of the temple.} {\color{purple}[Hindi/\mixtral]} \\
\textsc{Reason} & Incorrect reasons or causative explanations for events or situations.  & \textit{"Ni o lobika na ligamu, o sa vakalevutaka na kena yawa mai yalomu" means folding your elbow increase the distance from your spirit"} {\color{purple}[Fijian/\gptturbo]} \\
\midrule
\multicolumn{3}{c}{\textit{Contextual and Logical Errors}} \\
\midrule
\textsc{Scope} &  Errors involving the incorrect extent or range of a fact.  & \textit{"Überhangmandat" explicitely limited to the German Bundestag which is wrong. It can also apply for regional votes for a single state for instance.} {\color{purple}[German/\claude]} \\
\textsc{Illogical} &   Statements that are logically inconsistent or defy common sense.  & \textit{Yes, there is an error in the phrases "e rawa ni vakavuna na turu ni wai ni mata ena vuku ni kena vakacacani na mata ni mata" means the cutting of onion can damage the eye of the eye sounds grammatically incorrect.} {\color{purple}[Fijian/\gptturbo]}  \\
\textsc{Negation} &   Incorrect use of negation or statements that present the opposite of what is true.  & \textit{Answer implies that internists are not PCPs} {\color{purple}[English/\claude]} \\
\textsc{Placement} &   Inaccuracies resulting from the improper contextual placement of information.  & \textit{Putting "7. secondhand smoke" as a dot point incorrectly suggests that it is a reason why smoking is bad for the heart} {\color{purple}[English/\claude]} \\
\midrule
\multicolumn{3}{c}{\textit{Interpretative and Relevance Errors}} \\
\midrule
\textsc{Irrelevant or Hallucinated} &  All information in the answer is irrelevant or is fabricated.  & \textit{The entire answer is wrong. The question is to state the difference between General and OBC category of people (based on caste and society status), but the answer is about general and gynecology practitioners (doctors). It is irrelevant to the question.} {\color{purple}[Hindi/\mixtral]} \\
\textsc{Different Culture} &  The answer includes references to the wrong culture.  & \textit{The answer given is incorrect because it is not specific to Burundi. It is in Kinyarwanda and talks about Rwanda.} {\color{purple}[Kirundi/\claude]}  \\
\textsc{Unclear or Undocumented} &  Facts that are not clearly expressed or lack sufficient documentation or sourcing.  & \textit{The phrase "leweka" is not clear in the content. This issue might have a typo error.} {\color{purple}[Fijian/\claude]}  \\
\textsc{Not an answer} &  The model refuses to answer usually producing the response in English sometimes accompanied with an attempt to translate the question.  &  \textit{Incorrect language. (refusal to answer produced in English)} {\color{purple}[Fijian/\mixtral]}\\
\bottomrule
\end{tabular}
}
\caption{Categories used for analysis of comments on the factuality of the answers}
\label{tab:categories_factuality_annotations}
}
\end{table*}

\begin{table*}
\centering
\resizebox{1\textwidth}{!}{%
\begin{tabular}{l l c c c c c c}
\toprule
\textsc{Model} & \textsc{Type} & \textsc{Unclear/ Undoc.} & \textsc{Date} & \textsc{Entity} & \textsc{Event} & \textsc{Reason} & \textsc{Scope} \\
\midrule
\claude     & \textit{Spec.}      & 1 & 1 & 2 & 1 & 0 & 2 \\
\gptturbo & \textit{Spec.}      & 0 & 1 & 1 & 0 & 0 & 1 \\
\mixtral & \textit{Spec.}       & 1 & 1 & 2 & 2 & 1 & 1 \\
\claude      & \textit{Agn.}   & 2 & 0 & 0 & 0 & 0 & 0 \\
\gptturbo & \textit{Agn.}  & 5 & 0 & 0 & 0 & 1 & 0 \\
\mixtral & \textit{Agn.}  & 2 & 0 & 0 & 0 & 1 & 0 \\
\midrule
\textsc{Model} & \textsc{Type} & \textsc{Illogical} & \textsc{Placement} & \textsc{Negation} & \textsc{Irrelevant or Hallucinated} & \textsc{Different Culture} & \textsc{Not an Answer} \\
\midrule
\claude      & \textit{Spec.}       & 4 & 0 & 1 & 6 & 4 & 0 \\
\gptturbo & \textit{Spec.}      & 2 & 1 & 0 & 0 & 7 & 0 \\
\mixtral & \textit{Spec.} & 3 & 0 & 2 & 7 & 0 & 8 \\
\claude      &\textit{Agn.} & 1 & 1 & 0 & 6 & 0 & 0 \\
\gptturbo & \textit{Agn.}   & 1 & 1 & 1 & 4 & 0 & 0 \\
\mixtral & \textit{Agn.}  & 0 & 0 & 1 & 5 & 0 & 10 \\
\bottomrule
\end{tabular}
}
\caption{Count of different types of factuality issues mentioned by annotators in their comments. The issues are presented by question type (\textit{culturally specific} or \textit{culturally agnostic}) and by model which generated the answer. The taxonomy used for this annotation can be found in \autoref{tab:categories_factuality_annotations}.}
\label{app_tab:factuality_comments_annot_counts}
\end{table*}

\begin{table*}
\centering
{\renewcommand{\arraystretch}{1.5}
\resizebox{0.8\textwidth}{!}{%
\begin{tabular}{lp{0.25\textwidth}p{0.55\textwidth}}
\toprule
\textsc{Category} & \textsc{Description} & \textsc{Comment Example ({\color{purple}Language/Model})} \\
\midrule
\multicolumn{3}{c}{\textit{Content Issues}} \\
\midrule
\textsc{Unhelpfulness}
    & Annotator cannot discern the question's answer from the provided answer text
        & \textit{This answer is not very helpful because it not specific to Burundi. Of course it contains some relevant information, but it lacks specificity.} {\color{purple}[Kirundi/\gptturbo]} \\
\textsc{Irrelevance}
    & Contains information not needed to answer the question
        & \textit{The answer is irrelevant to the question and it isn't helpful. It explains a completely different thing (explains the importance of blood sample rather than explaining why the blood sample is taken from the elbow area).} {\color{purple}[Hindi/\mixtral]} \\
\textsc{Underspecificity}
    & Ideas are not explained in sufficient detail
        & \textit{Some instances of generic reasons without much detail about how it applies in Western Europe in particular.} {\color{purple}[English/\gptturbo]} \\
\textsc{Incompleteness}
    & Extra information is needed to understand the answer
        & \textit{The points mentioned in the answer are true but they do not cover the entirety of the topic. ... So it could be said that the answer is subjective rather than complete and clear.} {\color{purple}[Hindi/\gptturbo]} \\
\textsc{Repetitiveness}
    & Some phrase or idea is repeated
        & \textit{The phrases "Sa vakamuria na noda vakarau ni dautama o ira na iTaukei vulagi.....," is repeated in almost every line...} {\color{purple}[Fijian/\mixtral]} \\
\textsc{Grammar/spelling}
    & Grammar or spelling mistakes are present
        & \textit{There is a minor error, and the system might have spelled "nodra" incorrectly.} {\color{purple}[Fijian/\claude]} \\
\midrule
\multicolumn{3}{c}{\textit{Content Merits}} \\
\midrule
\textsc{Helpfulness}
    & Information needed to answer the question is present and understandable
        & \textit{It is helpful because it tells us the reasons for updating apps: Security, Good performance and New features.} {\color{purple}[Kirundi/\gptturbo]} \\
\textsc{Completeness}
    & Answer does not need extra information to be understood
        & \textit{It is complete, even makes an arc from outer core to inner core and a camparison to the sun surface temperature. So it covers more for what has been asked for but I like this additional information.} {\color{purple}[German/\gptturbo]} \\
\midrule
\multicolumn{3}{c}{\textit{Presentation Issues}} \\
\midrule
\textsc{Artificiality}
    & Perceived by the annotator to be written by AI
        & \textit{Structure with asterix and enumeration very artificial / non human like.} {\color{purple}[German/\gptturbo]} \\
\textsc{Non-fluency}
    & Reading the text is awkward/difficult for the annotator
        & \textit{...Also Argument 5 is "estimation" and this is again mentioned in the summary -> a bit stange and hence not fluent.} {\color{purple}[German/\claude]} \\
\textsc{Inclarity}
    & Answer ideas or terms are difficult to understand
        & \textit{Fluent and complete, but not clear, as complex word i.e. Zytokinen (Cytokines) are not explained.} {\color{purple}[German/\claude]} \\
\midrule
\multicolumn{3}{c}{\textit{Presentation Merits}} \\
\midrule
\textsc{Naturality}
    & Perceived by the annotator to be human-written/natural
        & \textit{Complete, but so different to the other answers. most human-like answer as being relatively general to a more or less vague question.} {\color{purple}[German/\claude]} \\
\textsc{Fluency}
    & Written in a smooth manner and easy to read
        & \textit{There are no inaccuracies in the answer which makes it helpful. It makes use of easier vocabulary which sounds fluent.} {\color{purple}[Hindi/\claude]} \\
\textsc{Clarity}
    & Ideas are expressed in an easy-to-understand manner
        & \textit{The answer is complete and helpful. It is very clear because the information is subdivided into general and OBC sections and it is easy to follow.} {\color{purple}[Hindi/\gptturbo]} \\
\bottomrule
\end{tabular}
}
\caption{Categories used for the analysis of annotators' general comments on the quality of answers}
\label{tab:categories_of_general_comment_annotations}
}
\end{table*}

\begin{table*}
\centering
\resizebox{1\textwidth}{!}{%
\begin{tabular}{l l c c c c c}
\toprule
\textsc{Model} & \textsc{Type} & \textsc{Unhelpfulness*} & \textsc{Incompleteness*} & \textsc{Artificiality*} & \textsc{Non-fluency*} & \textsc{Inclarity*}  \\
\midrule
\claude     & \textit{Spec.}    & 9 & 13 & 4 & 4 & 10 \\
\gptturbo & \textit{Spec.}      & 6 & 7 & 4 & 2 & 5 \\
\mixtral & \textit{Spec.}       & 20 & 17 & 19 & 10 & 15 \\
\midrule
\claude      & \textit{Agn.}    & 3 & 5 & 5 & 1 & 1 \\
\gptturbo & \textit{Agn.}       & 1 & 4 & 11 & 1 & 5 \\
\mixtral & \textit{Agn.}        & 16 & 14 & 17 & 8 & 9 \\
\midrule
\textsc{Model} & \textsc{Type} & \textsc{Irrelevance} & \textsc{Underspecifity} & \textsc{Repetitiveness} & \textsc{Grammar/Spelling} \\
\midrule
\claude      & \textit{Spec.}       & 1 & 1 & 0 & 1 \\
\gptturbo & \textit{Spec.}          & 4 & 1 & 0 & 2 \\
\mixtral & \textit{Spec.}           & 4 & 7 & 4 & 1 \\
\midrule
\claude &\textit{Agn.}              & 1 & 1 & 2 & 1 \\
\gptturbo & \textit{Agn.}           & 0 & 0 & 0 & 7 \\
\mixtral & \textit{Agn.}            & 3 & 1 & 4 & 0 \\
\bottomrule
\end{tabular}
}
\caption{Counts of different types of issues noted in annotators' comments about general answer quality. The issues are presented by question type (\textit{culturally specific} or \textit{culturally agnostic}) and by model which generated the answer. The taxonomy used for this annotation can be found in Table \ref{tab:categories_of_general_comment_annotations}. Our UI suggested to  annotators to make comments (positive or negative) about categories marked with *.}
\label{tab:issue_count_of_general_comment_annotations}
\end{table*}

\begin{table*}
\centering
\resizebox{1\textwidth}{!}{%
\begin{tabular}{l l c c c c c}
\toprule
\textsc{Model} & \textsc{Type} & \textsc{Helpfulness} & \textsc{Completeness} & \textsc{Naturality} & \textsc{Fluency} & \textsc{Clarity}  \\
\midrule
\claude     & \textit{Spec.}    & 24 & 19 & 24 & 22 & 22 \\
\gptturbo & \textit{Spec.}      & 24 & 20 & 27 & 22 & 19 \\
\mixtral & \textit{Spec.}       & 13 & 7  & 18 & 15 & 12 \\
\midrule
\claude      & \textit{Agn.}    & 26 & 20 & 30 & 22 & 30 \\
\gptturbo & \textit{Agn.}       & 29 & 24 & 26 & 17 & 25 \\
\mixtral & \textit{Agn.}        & 16 & 12 & 22 & 13 & 16 \\
\bottomrule
\end{tabular}
}
\caption{Counts of different types of merits noted in annotators' comments about general answer quality. The merits are presented by question type (\textit{culturally specific} or \textit{culturally agnostic}) and by model which generated the answer. The taxonomy used for this annotation can be found in Table \ref{tab:categories_of_general_comment_annotations}. Our UI suggested to  annotators to make comments (positive or negative) about all these categories.}
\label{tab:merit_count_of_general_comment_annotations}
\end{table*}

\end{document}